\documentclass[12pt,onecolumn]{article}

\usepackage[margin=1in]{geometry}
\usepackage{graphicx}
\usepackage{subcaption}
\usepackage{caption}
\usepackage{amsmath}
\usepackage{amssymb}
\usepackage{times}
\usepackage{placeins} 
\usepackage{float}
\usepackage{booktabs}
\usepackage{multirow}   
\usepackage[hyphens]{url}  
\usepackage[bookmarks=false]{hyperref}
\usepackage{cite}
\usepackage{times}  
\usepackage{helvet}  
\usepackage{courier}  
\usepackage[hyphens]{url}  
\usepackage{graphicx} 
\usepackage{amsmath}   
\usepackage{amssymb}   
\usepackage{amsfonts}  
\usepackage{booktabs}   
\usepackage{multirow}   
\usepackage{xcolor}
\usepackage{subcaption}   
\renewcommand{\thetable}{S\arabic{table}}
\title{\bf Advancing Cache-Based Few-Shot Classification via Patch-Driven Relational Gated Graph Attention \\[0.25cm] 
}

\date{}

\begin{document}
\maketitle

\vspace{-2.5cm}

\begin{center}
\author{
Tasweer Ahmad\footnotemark[1],
Arindam Sikdar\footnotemark[1],
Sandip Pradhan,
Ardhendu Behera\footnotemark[2]
}
\end{center}

\footnotetext[1]{Equal contribution.}
\footnotetext[2]{Corresponding author: beheraa@edgehill.ac.uk.}


\begingroup
\renewcommand\thefootnote{}
\footnotetext{Tasweer Ahmad, Arindam Sikdar, Sandip Pradhan, and Ardhendu Behera are with the Department of Computer Science, Edge Hill University, Ormskirk, Lancashire, L39 4QP, UK.}
\addtocounter{footnote}{-1}
\endgroup





\begin{abstract}
\noindent
Few-shot image classification remains difficult under limited supervision and visual domain shift. Recent cache-based adaptation approaches (e.g., Tip-Adapter) address this challenge to some extent by learning lightweight residual adapters over frozen features, yet they still inherit CLIP’s tendency to encode global, general-purpose representations that are not optimally discriminative to adapt the generalist to the specialist's domain in low-data regimes. We address this limitation with a novel patch-driven relational refinement that learns cache adapter weights from intra-image patch dependencies rather than treating an image embedding as a monolithic vector. Specifically, we introduce a relational gated graph attention network that constructs a patch graph and performs edge-aware attention to emphasize informative inter-patch interactions, producing context-enriched patch embeddings. A learnable multi-aggregation pooling then composes these into compact, task-discriminative representations that better align cache keys with the target few-shot classes. Crucially, the proposed graph refinement is used only during training to distill relational structure into the cache, incurring no additional inference cost beyond standard cache lookup. Final predictions are obtained by a residual fusion of cache similarity scores with CLIP zero-shot logits. Extensive evaluations on 11 benchmarks show consistent gains over state-of-the-art CLIP adapter and cache-based baselines while preserving zero-shot efficiency. We further validate battlefield relevance by introducing an \emph{Injured vs. Uninjured Soldier} dataset for casualty recognition. It is motivated by the operational need to support triage decisions within the ``platinum minutes'' and the broader ``golden hour'' window in time-critical UAV-driven search-and-rescue and combat casualty care. Our source code and dataset will be available at \textit{\url{https://github.com/tasveerahmad/Patch-Relational-Graph-Attention}}.
\end{abstract}



\section{Introduction}
Large-scale vision-language models (VLMs), most notably CLIP~\cite{radford2021learning}, have revolutionized the cross-modal representation learning by jointly reasoning over images and texts~\cite{jia2021scaling,li2021supervision,yao2022filip,li2022grounded,wang2022medclip,zhang2024vision}. Trained using contrastive learning on web-scale image–text corpora, these models learn transferable embeddings that enable strong zero-shot generalization across a wide range of downstream tasks, often without task-specific retraining. This transferability is particularly appealing for low-data regimes, where collecting and curating labels is expensive, and it has opened new opportunities for few-shot learning under realistic distribution shifts. Zero-shot and few-shot classification with VLMs is commonly studied under two evaluation paradigms. Inductive methods~\cite{gao2024clip, khattak2023maple, zhou2022learning} treat every test example independently, without exploiting dependencies within the query set. This aligns well with practical deployment constraints such as streaming inputs, privacy limits, and low-latency requirements. Transductive approaches~\cite{boudiaf2020information, guneet2020baseline, lazarou2021iterative}, in contrast, exploit the global structure of the unlabeled query set to refine predictions collaboratively, often yielding accuracy gains when full-batch access is feasible. CLIP provides a strong foundation in both settings, but the inductive regime is frequently preferred in real-world systems because it avoids test-time coupling and additional computation.

Despite its effectiveness, CLIP zero-shot recognition is typically implemented by matching an image embedding to text embeddings produced from prompts such as ``a photo of a [CLASS]'', which can degrade under fine-grained recognition and domain shift, and is often sensitive to prompt design. However, its performance declines in fine-grained or domain-shifted settings and is highly sensitive to prompt design. These limitations have motivated few-shot adaptation techniques that retain CLIP’s generalization while exploiting a small labeled support set. Within the inductive paradigm, two prominent strategies have emerged: i) prompt-learning and ii) adapter-based tuning. Prompt-learning~\cite{zhou2022learning, zhu2023prompt} replaces hand-crafted prompts with a small number of learnable continuous tokens, improving alignment with target classes under limited supervision. Adapter-based methods~\cite{zhang2022tip, gao2024clip} instead introduce lightweight trainable modules into the image and/or text branch to steer frozen CLIP representations toward the target task, offering a practical trade-off between accuracy, parameter efficiency, and scalability.

\begin{figure*}[t!]
	\centering
	\includegraphics[width=1\textwidth]{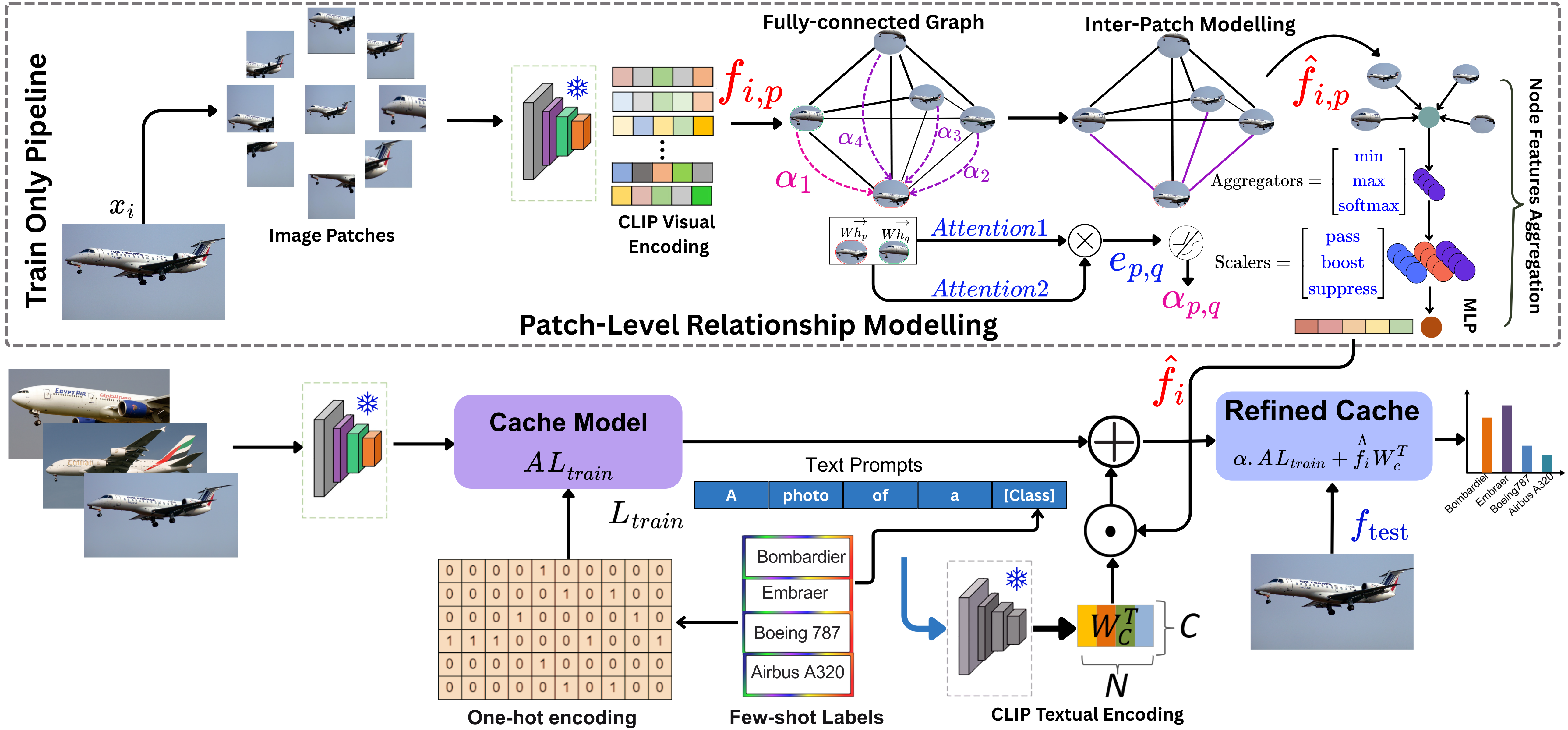} 
	\caption{Overview of our proposed framework. During training, CLIP encoded image patches form nodes ($f_{i,p}$) of a fully-connected graph, where dependencies between patches are modeled via our relational attention mechanism ($\hat{f}_{i,p}$ in Eqn~(\ref{eq:aggregation})). Aggregated node features ($\hat{f}_i$ in Eqn~(\ref{eq:logits_train})) update a cache representation. At inference, predictions are made by combining the refined cache with CLIP’s zero-shot logits, enabling efficient and robust adaptation.
    }
	\label{block_diagram_v2}
\end{figure*}

\textbf{Motivation:} Although recent prompt-learning and adapter-based methods have advanced few-shot adaptation for VLMs, they largely treat an image as a single global embedding. This ``monolithic vector'' view discards the rich spatial structure that often determines class membership, particularly when the decision hinges on subtle part-level cues, localized attributes, or background foreground interactions. The issue is amplified under distribution shift, where global semantics may remain similar, but discriminative evidence moves to fine-grained regions (e.g., texture, markings, small objects, or pose-specific parts). At the same time, semantic alignment in CLIP is not just determined by individual regions in isolation; relationships among regions (part–whole composition, co-occurrence, and contextual support) can be crucial for resolving ambiguous classes and for aligning visual evidence with text prompts. Yet most existing few-shot adaptations do not explicitly model these intra-image dependencies and thus cannot selectively amplify the most informative relational cues when training data are scarce.

To address this gap, we introduce a cache-based few-shot framework that injects lightweight \textit{relational reasoning} into CLIP adaptation by modeling \textit{inter-patch dependencies} with a graph formulation. By learning how patches support or suppress each other through edge-aware attention, the cache is enriched with representations that are both more discriminative and more context-aware, improving robustness in fine-grained and shifted domains. Importantly, we confine graph computation to training and distill its benefits into the cache/adapters, preserving the key practical advantage of cache-based methods such as \emph{zero–shot–like inference efficiency with no additional test-time overhead}. This yields a principled and deployable path to reconcile structural modeling with scalability in a few-shot classification.

\section{Related Work}
\paragraph{Vision-Language Pre-trained Models} 
Adapting vision–language models (VLM) for few-shot scenarios focuses on leveraging pre-trained cross-modal representations with minimal labeled data.
Pre-trained VLMs have proven effective in enhancing downstream tasks such as few-shot classification~\cite{gao2024clip},~\cite{yu2023task},~\cite{zhou2022learning}, cross-modal generation \cite{nichol2022glide},~\cite{patashnik2021styleclip},~\cite{ramesh2022hierarchical}, and visual recognition.
Adapting the VLMs for few-shot learning tunes the partial parameters and is broadly classified into prompt-based~\cite{lu2022prompt},~\cite{zhou2022conditional},~\cite{zhou2022learning} and adapter-based tuning~\cite{yu2023task},~\cite{zhang2022tip}, ~\cite{li2024graphadapter}. CoOp \cite{zhou2022learning} pioneered the idea of learnable prompts to adapt task-specific knowledge for few-shot learning. Building on this foundation, numerous subsequent works \cite{chen2023plot}, \cite{yao2023visual}, \cite{khattak2023maple}, \cite{zhu2023prompt} have refined prompt tuning from various angles, including enhancing generalization~\cite{zhou2022conditional}, leveraging knowledge prototypes \cite{zhang2022prompting}, introducing augmentation \cite{zhang2023prompt}, and improving diversity \cite{lu2022prompt}. Meanwhile, adapter-based approaches adapt VLMs by inserting lightweight trainable layers into visual branches. CLIP-Adapter \cite{gao2024clip} employs a single bottleneck layer for efficient few-shot tuning, while TaskRes \cite{yu2023task} separates task-specific and prior knowledge through modular adapters. Tip-Adapter with fine-tuning \cite{zhang2022tip} introduces a learnable cache-based retrieval mechanism for few-shot learning. Unlike these methods, our approach is unique in that it adds patch-level relational modeling through graph reasoning to capture fine-grained context. Recent developments continue to push CLIP beyond lightweight tuning by introducing deeper task-aware or causally grounded adapter modules. For instance, Ta-Adapter~\cite{zhang2024ta} incorporates task-aware prompts within CLIP’s encoders, enabling stronger task specialization but requiring additional trainable components that remain active during inference. Similarly, CCA~\cite{jiang2025causal} disentangles CLIP’s visual features through Independent Component Analysis (ICA) and introduces cross-modal alignment layers, which also increase inference-time complexity. These approaches demonstrate the growing trend toward richer adaptation mechanisms but come at the cost of computational overhead. In contrast, our framework retains frozen CLIP encoders and performs all relational reasoning only during training. By distilling graph-enhanced representations directly into the cache keys, our method maintains the same test-time complexity as standard cache-based models while benefiting from stronger patch-level contextual refinement.
\paragraph{Cache Model} 
Cache-based models~\cite{khandelwal2020knnlm}, \cite{zhang2022tip}, \cite{orhan2018simple} offer a lightweight alternative to fine-tuning by storing training features as key–value pairs of embeddings and labels. During inference, predictions are made via similarity-based retrieval, eliminating the need for gradient updates. However, traditional methods often handle only one modality and face challenges like high storage demands or retrieval noise. Tip-Adapter-F \cite{zhang2022tip} refines cached visual embeddings through a lightweight learnable layer, increasing few-shot accuracy with minimal training while preserving CLIP's multimodal generalization. Building on this foundation, our approach advances cache-based adaptation by introducing patch-level relational reasoning during cache construction, enriching intra-image spatial context to enhance few-shot accuracy while preserving the inference efficiency.
\paragraph{Graph Learning for Few-shot} 
Graph learning has shown strong potential in few-shot learning by enabling message passing that propagates contextual and label information across limited samples. Unlike early metric-based methods~\cite{vinyals2016matching, snell2017protonet, sung2018relationnet},  with fixed similarity graphs, GNNs explicitly model relationships and allow multi-hop reasoning for richer feature refinement. 
Recent work explores the integration of GNNs with few-shot learning~\cite{garcia2018fewshot, kim2019metagnn, ziko2020laplacianshot}, and more recent advances such as GraphAdapter ~\cite{li2024graphadapter} further demonstrate the potential of graph-based reasoning for adapting vision–language models. However, existing approaches typically operate on global image features, whereas our method performs patch-level relational modeling to capture fine-grained spatial dependencies without any inference-time overhead.

\section{Methodology}
\label{sec:methodology}
\paragraph{Problem Formulation} 
We consider the task of few-shot image classification under a semi-parametric adaptation framework. Let the support set $\mathcal{S} = \{ (\mathbf{x}_i, c_i) \mid i = 1,\dots, N \cdot K \}$, where $c_i$ is the class label of $i^{th}$ image $\mathbf{x}_i$ and $N$ denote the number of classes and $K$ the number of shots per class. We denote $M = N \cdot K$ as the total number of support instances. Similarly, the query set is denoted by $\mathcal{Q} = \{ \mathbf{x}_j \}_{j=1}^{T}$, which comprises unlabeled examples from the same label space. Our goal is to adapt a frozen vision language model (e.g., CLIP) to efficiently classify an image, using only a limited number of labeled samples without full fine-tuning. Inspired by Tip-Adapter \cite{zhang2022tip}, we implement this using a lightweight key-value cache that stores visual representations and class labels as $\mathcal{C} = (\mathbf{F}_{\text{train}}, L_\text{train})$, where $\mathbf{F}_{\text{train}} \in \mathbb{R}^{(N\cdot K) \times d}$ contains normalized CLIP embeddings (keys) of dimension $d$ of support samples $\mathbf{f}_i = f_{\text{CLIP}}(\mathbf{x}_i)$, and $L_\text{train} \in \mathbb{R}^{(N\cdot K)\times N}$ represents one hot class label (values). This cache serves as the foundation for retrieval-based prediction. To enable a more precise steering of CLIP’s visual representations $\mathbf{f}_i$ toward the target task, our learnable cache model $\mathbf{F}_{\text{train}}$ is learned using our novel graph-enhanced representations, while the value matrix $L_\text{train}$ remains fixed.

The visual representation of our proposed approach is illustrated in Fig. \ref{block_diagram_v2}. Our graph-enhanced representations focus on the relationships between different parts (patches) of an image by modeling an image as a fully connected graph, where each patch is a node. Specifically, each image $\mathbf{x}_i$ is represented using $P$ patches $\{ \mathbf{x}_{i,p} \}_{p=1}^{P}$. These patches are encoded using the frozen CLIP as $\mathbf{f}_{i,p} = f_{\text{CLIP}}(\mathbf{x}_{i,p}) \in \mathbb{R}^d$ and are represented as nodes of a fully connected undirect graph $\mathcal{G} = (P, E)$ with $P$ nodes and $E$ edges between the nodes. Patches $\{ \mathbf{x}_{i,p} \}$ representing each image $\mathbf{x}_i$ pass through this graph to learn the relational dependencies between them through our novel attention-driven message passing mechanism.  

These graph-refined features ($\mathbf{f}_{i,p} \xrightarrow{\mathcal{G}} \hat{\mathbf{f}}_{i,p}$) representing each node of the graph $\mathcal{G}$ are then combined using a learnable multi-aggregation pooling strategy to form a final feature vector $\hat{\mathbf{f}}_i$ for the input image $\mathbf{x}_i$ during the updating of cache keys $\mathbf{F}_{\text{train}}$. The primary goal is to enhance the cache to improve the discriminative ability required for a few-shot classification. Consequently, during inference, our method employs the standard frozen clip embedding $\mathbf{f}_j$ of the $j^{th}$ query image $\mathbf{x}_j$, preserving the efficiency of zero-shot inference. This whole process is demonstrated in Fig. \ref{block_diagram_v2}.

\begin{figure}[t!]
	\centering
	\includegraphics[width=0.6\columnwidth]{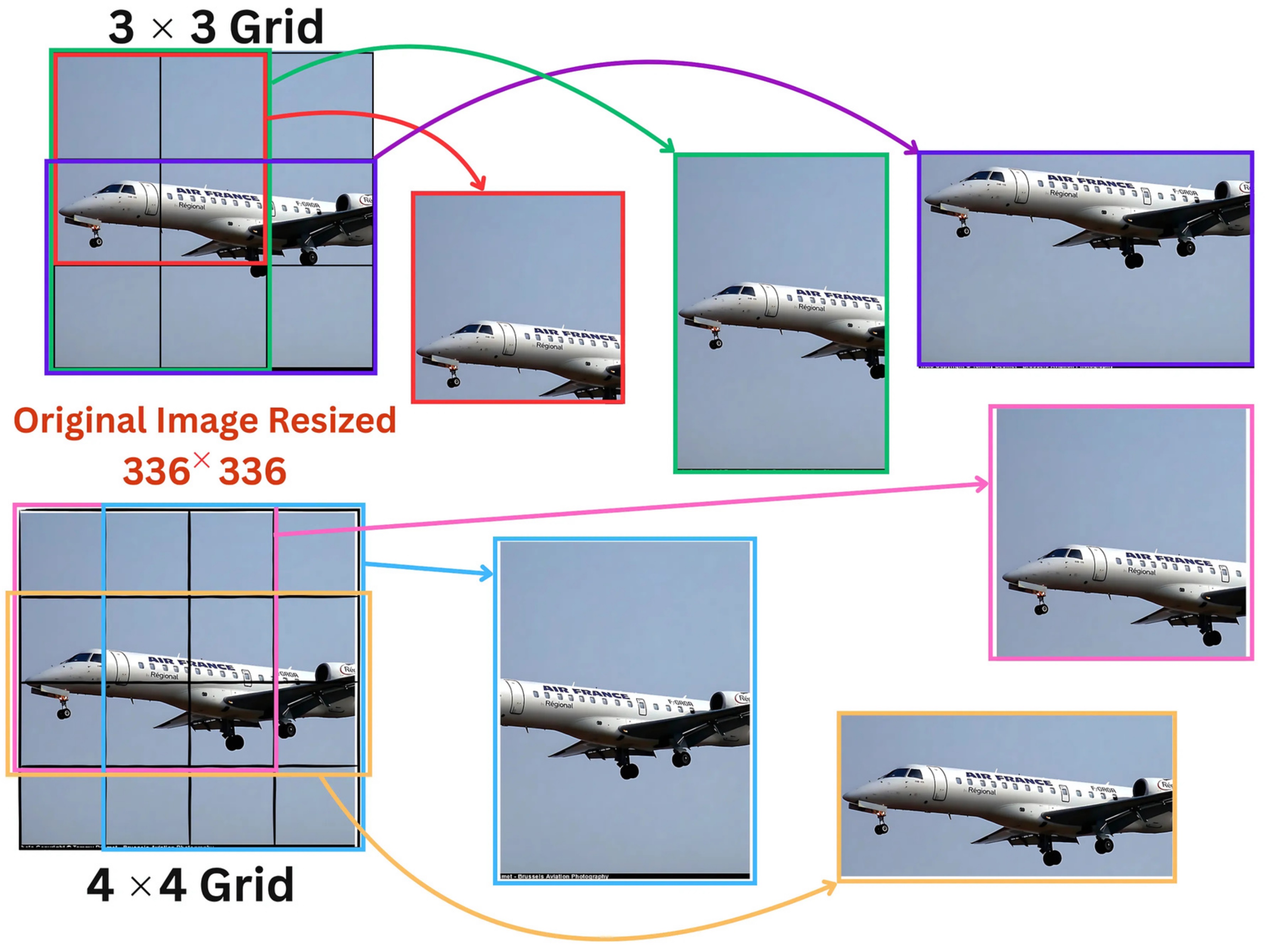}
    \caption{Patch generation using $(3{\times}3)$ and $(4{\times}4)$ grid tiling, where adjacent tiles form multi-scale patches encoded by a frozen CLIP to produce graph nodes. Only 6 regions are shown for illustration.}
	\label{fig:patch_generation}
\end{figure}

\paragraph{Graph-driven Image Patch Embedding} We extract P patches from a training image $\mathbf{x}_i$ as shown in Fig. \ref{fig:patch_generation}. Each patch $p=1,\dots, P$ in $\mathbf{x}_i$ is processed by the frozen CLIP visual encoder to give the feature $\mathbf{f}_{i,p} = f_{\text{CLIP}}(\mathbf{x}_{i,p})$. The aim is to update the visual feature $\mathbf{f}_{i,p}$ by modeling relationships between patches by propagating information between them. Thus, GNN is used to achieve this by exploring the above-mentioned fully connected undirected graph $\mathcal{G} = (P, E)$ described by an adjacency matrix $A \in \mathbb{R}^{P\times P}$. The graph $\mathcal{G}$ takes a set of features $H^{(l)}=\{h_1^l,\dots ,h_P^l\}$ as input at layer $l$ to calculate the output features $H^{(l+1)}=\{h^{l+1}_1,\dots ,h_P^{l+1}\}$ using a well-known algorithm $H^{(l+1)} = \rho{(\hat{A}H^{(l)}W^{(l)})} $ \cite{kipf2017gcn}, with $H^{(0)} = \{\mathbf{f}_{i,1},\dots ,\mathbf{f}_{i,P}\}$ representing patch features for the $i^{th}$ image $\mathbf{x}_i$, $W^{(l)}$ is the weight matrix in the $l^{th}$ layer and $\rho{(\cdot)}$ is a non-linear activation function (e.g., ReLU). $\hat{A} = D^{-1/2}AD^{-1/2}$ is the symmetrically normalized adjacency matrix, and $D$ is the diagonal node degree matrix of $A$. 

\paragraph{Attention-driven Patch Feature Transformation} The goal is to compute the visual representations of each patch $p=1,\dots, P$ in the graph $\mathcal{G}$ by attending to its neighbors, following our novel attention strategy. Let us consider a patch $p\in P$. The calculation of the visual feature for patch $p$ linking other patches $q\in P$ in layer $l+1$ is calculated as: 
\begin{equation}
h_p^{(l+1)} = \rho \Bigg( \sum_{q \in P} \alpha_{p,q}^{(l+1)} W^{l+1} h_q^{(l)} \Bigg)
\label{eq:message_passing}
\end{equation}
It linearly combines the features of the neighboring patches $q\in P$ by the attention coefficients $ \alpha_{p,q}^{(l+1)}$. 
We compute $ \alpha_{p,q}^{(l+1)}=\texttt{softmax}_q(\texttt{LeakyReLU}(e_{p,q}^{l+1}))$. The novelty is the computation of the attention coefficient $e_{p,q}$ that indicates the importance of the features of the patch $p$ to the patch $q$. In our graph model, each patch is allowed to attend to every other patch, since we begin with a noisy graph in which not all patches exhibit the same level of relationships with each other. Thus, we compute and reinforce the relationships between them through our edge-attention mechanism to compute $e_{p,q}$, which is inspired by recent work on self-supervised attention between graph nodes \cite{kim2022find}. Our edge-attention is computed as:  
\begin{equation}
\begin{split}
e_{p,q}^{l+1} = \overbrace{(\mathbf{a}^{l+1})^\top \big[ \mathbf{W}^{l+1} h_p^{l} \, \| \, \mathbf{W}^{l+1} h_q^{l} \big]}^\text{Attention 1} \cdot \\
\sigma \underbrace{\Big( (\mathbf{W}^{l+1} h_p^{l})^\top \cdot \mathbf{W}^{l+1} h_q^{l} \Big)}_\text{Attention 2} 
\end{split}
\label{eq:mx_attention}
\end{equation}
where $\|$ denotes concatenation and $\sigma(\cdot)$ is the sigmoid activation function. Eqn~(\ref{eq:mx_attention}) multiplies two attention mechanisms (\texttt{Attention 1} and \texttt{Attention 2}) with a sigmoid $\sigma(\cdot)$ activation for \texttt{Attention 2}. \texttt{Attention 1} is nothing more than a widely used graph attention GAT \cite{velivckovic2018graph} to compute the coefficients by a single-layer feedforward network parameterized by $\mathbf{a}^{l+1}$. Similarly, \texttt{Attention 2} is the dot-product attention \cite{vaswani2017attention} motivated by node representation learning. The multiplication of two attention mechanisms with a sigmoid is motivated by the gating mechanism of Gated Recurrent Units (GRU). Our goal is to combine \textbf{relationship-based filtering} (\texttt{Attention 1:} \textit{based on a learned function, which of my neighbors are most structurally or contextually relevant to me?}) and \textbf{content-based filtering} (\texttt{Attention 2:} \textit{which of my neighbors are most semantically similar to me?}) that can lead to better patch representation. We hypothesize that this combination of two complementary types of attention creates a richer, more robust, and more expressive model capable of capturing a wider variety of relational dependencies between patches in a graph. It is worth re-emphasizing that our focus is on few-shot classification, and thus, the aim is to aggregate patch features from fewer images based on both content similarity (\texttt{Attention 2}) and learned structural context (\texttt{Attention 1}).

\paragraph{Learnable Multi-Aggregation} After refining the patch embeddings of the image $\mathbf{x}_i$ through our relational graph $\mathcal{G}$, these embeddings ($\mathbf{f}_{i,p} \xrightarrow{\mathcal{G}} \hat{\mathbf{f}}_{i,p}$, where $p\in P$, $\mathbf{f}_{i,p} \in \mathbf{x}_i$, $\hat{\mathbf{f}}_{i,p} = h_p^{L}$, and $L$ is the last layer of our graph $\mathcal{G}$) are aggregated to a single feature vector to enrich cache keys $\mathbf{F}_{\text{train}}$. A standard approach would be a fixed pooling operation such as \textit{mean} or \textit{max}. Instead, we adopt a learnable multi-aggregation pooling mechanism inspired by \cite{corso2020principal} that efficiently combines multiple statistics (e.g., min, max, mean and standard deviation) within our graph context. 
Through multi-aggregation, the patch-refined features $\hat{\mathbf{f}}_{i,p}$ of $P$ patches (nodes in graph $\mathcal{G}$) are aggregated to form the final embedding: 
\begin{equation}
\mathbf{\hat{f}}_i= \sum_{m \in \psi} \gamma_m \cdot \big \{\mathbf{W}_m \cdot \phi_m\big( \{ \hat{\mathbf{f}}_{i,p} \}_{p\in P} \big) \big \}
\label{eq:aggregation}
\end{equation}
In Eqn~(3), $\psi$ denotes the set of aggregation functions (e.g., mean, max, std). For each $m \in \psi$, $\phi_m(\cdot)$ is the corresponding pooling operator applied to the refined patch features $\{\hat{f}_{i,p}\}$. The coefficient $\gamma_m$ is a learnable weight for each aggregation branch, and $W_m$ is the associated linear projection.


\paragraph{Cache Refinement} The goal is to improve the alignment between visual and textual features while maintaining inference efficiency. During training, the final visual feature $\hat{\mathbf{f}}_i$ is L2-normalized, which propagates as a query to retrieve class information from the cache.
Following the strategy in Tip-Adapter~\cite{zhang2022tip}, we initialize a learnable cache key matrix $\Theta_{\text{train}} \in \mathbb{R}^{M \times d}$ using support features $\mathbf{F}_{\text{train}}$, as $\Theta_0 = \mathbf{F}_{\text{train}}$. This enables the cache keys to be refined through gradient-based optimization during training. For learnable cache keys, the affinity matrix for query features $\mathbf{f}_q$ and cache keys is computed as:
\begin{equation}
\mathbf{A} = \exp\big(-\beta \big( \mathbf{1} - \mathbf{f}_q \cdot \Theta_{\text{train}}^\top \big)\big),
\label{eq:affinity}
\end{equation}

\noindent
where $\beta$ controls the sharpness of similarity scaling—larger values amplify closer matches more strongly. In Eqn~(\ref{eq:affinity}), the affinity matrix $\mathbf{A}$ leverages the inner product $\mathbf{f}_q \cdot \Theta_{\text{train}}^\top$ to compute cosine similarity, ensuring robust semantic matching effectively. The resulting $\mathbf{A}$ is used to perform a weighted summation on the fixed one-hot label matrix $\mathbf{L}_{\text{train}}$, which produces the probability of the class as $\mathbf{A}\times \mathbf{L}_{\text{train}} \in \mathbb{R}^{1 \times N}$. 


The graph-refined embedding $\hat{\mathbf{f}}_i$, derived from the same frozen CLIP feature as $\mathbf{f}_i$, is projected onto the pre-trained class weight matrix $\mathbf{W}_c^\top$ to produce an auxiliary logit vector $\mathbf{\hat{f}}_i \mathbf{W}_c^\top \in \mathbb{R}^{1 \times N}$. This stream enriches cache-based prediction with CLIP's prior knowledge and relational cues; and during training, provides gradient signals that help refine the cache keys $\Theta_{\text{train}}$ toward graph-driven semantics. The final prediction is computed as a residual combination of cache-driven and auxiliary logits:


\begin{equation}
\text{logits} = \alpha \cdot \mathbf{A} L_\text{train}+ \mathbf{\hat{f}}_i \mathbf{W}_c^\top,
\label{eq:logits_train}
\end{equation}
where $\alpha$ balances zero-shot priors and cache adaptation, with higher values favoring adaptation under domain shifts and lower values preserving CLIP’s generalization. This residual design in Eqn~(\ref{eq:logits_train}) unifies support-set cache refinement with graph-enhanced semantic cues.

\paragraph{Final Prediction and Inference Pipeline}
At test time, the predictions are made without invoking any graph-based computations, ensuring efficient inference. Given a test image, the features $\mathbf{f}_{\text{test}} \in \mathbb{R}^{1 \times d}$ are extracted using the frozen CLIP image encoder, the cache key similarity score is computed as $\mathbf{A} = \exp\big(-\beta ( \mathbf{1} - \mathbf{f}_{\text{test}} \mathbf{F}_\text{train}^\top )\big)$ and the cache logits are $\mathbf{A}\times L_\text{train}$. Finally, the image and text streams are fused to calculate
$\text{final logits} = \alpha \cdot \mathbf{A} L_\text{train} + \mathbf{f}_{\text{test}} \mathbf{W}_c^\top $. 

\begin{figure*}[htbp]
  \centering
    \begin{subfigure}[b]{0.8\textwidth}
    \includegraphics[width=\textwidth]{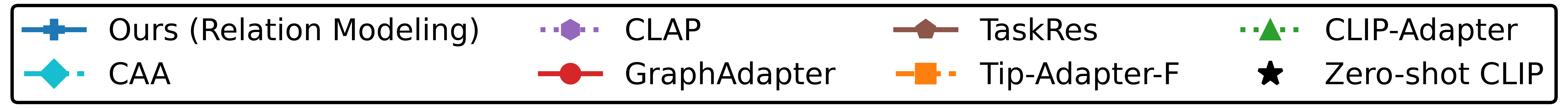}
    \label{fig:legends_Only}
  \end{subfigure}
    \begin{subfigure}[b]{0.3\textwidth}
    \includegraphics[width=\textwidth]{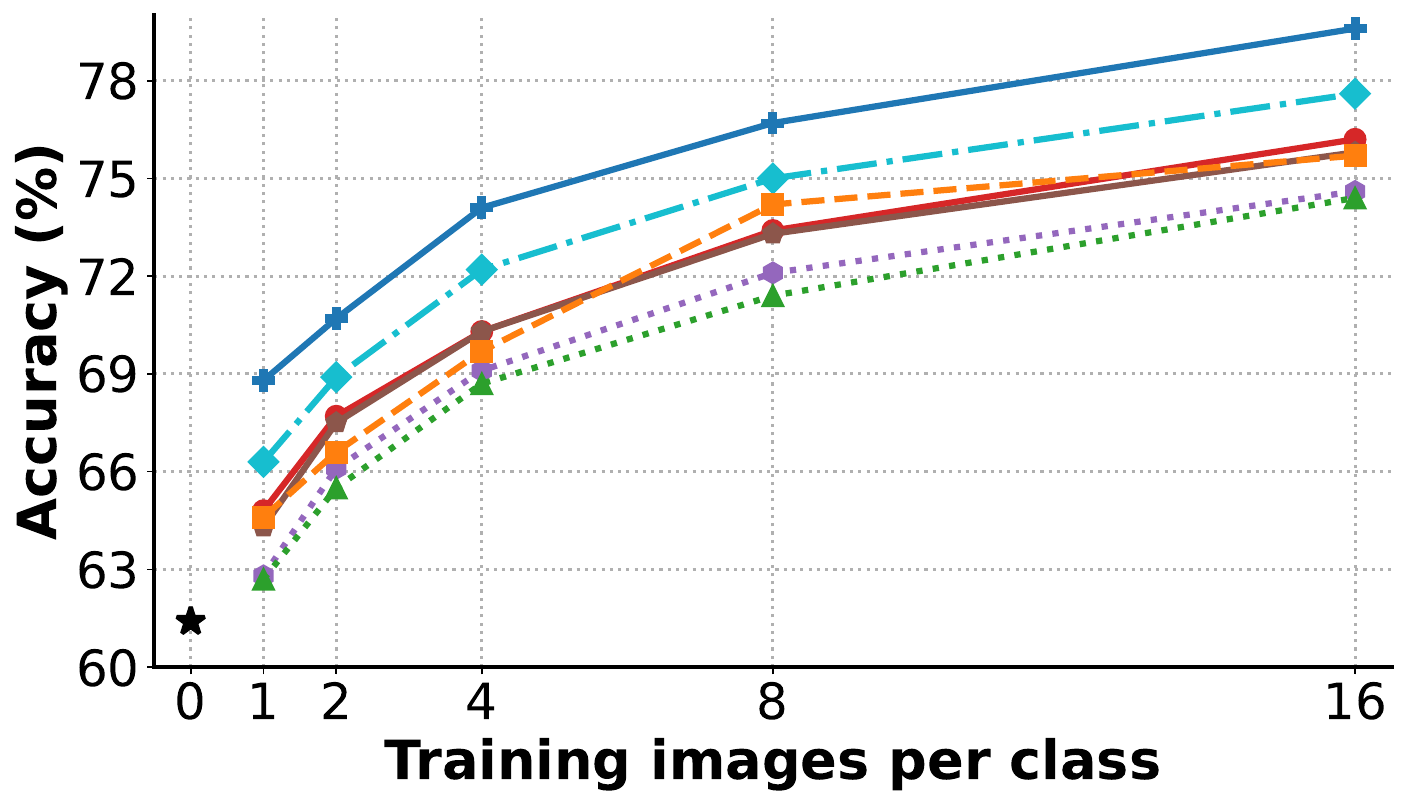}
    \caption{\textbf{Average}}
    \label{fig:average}
  \end{subfigure}
  \hfill
  \begin{subfigure}[b]{0.3\textwidth}
    \includegraphics[width=\textwidth]{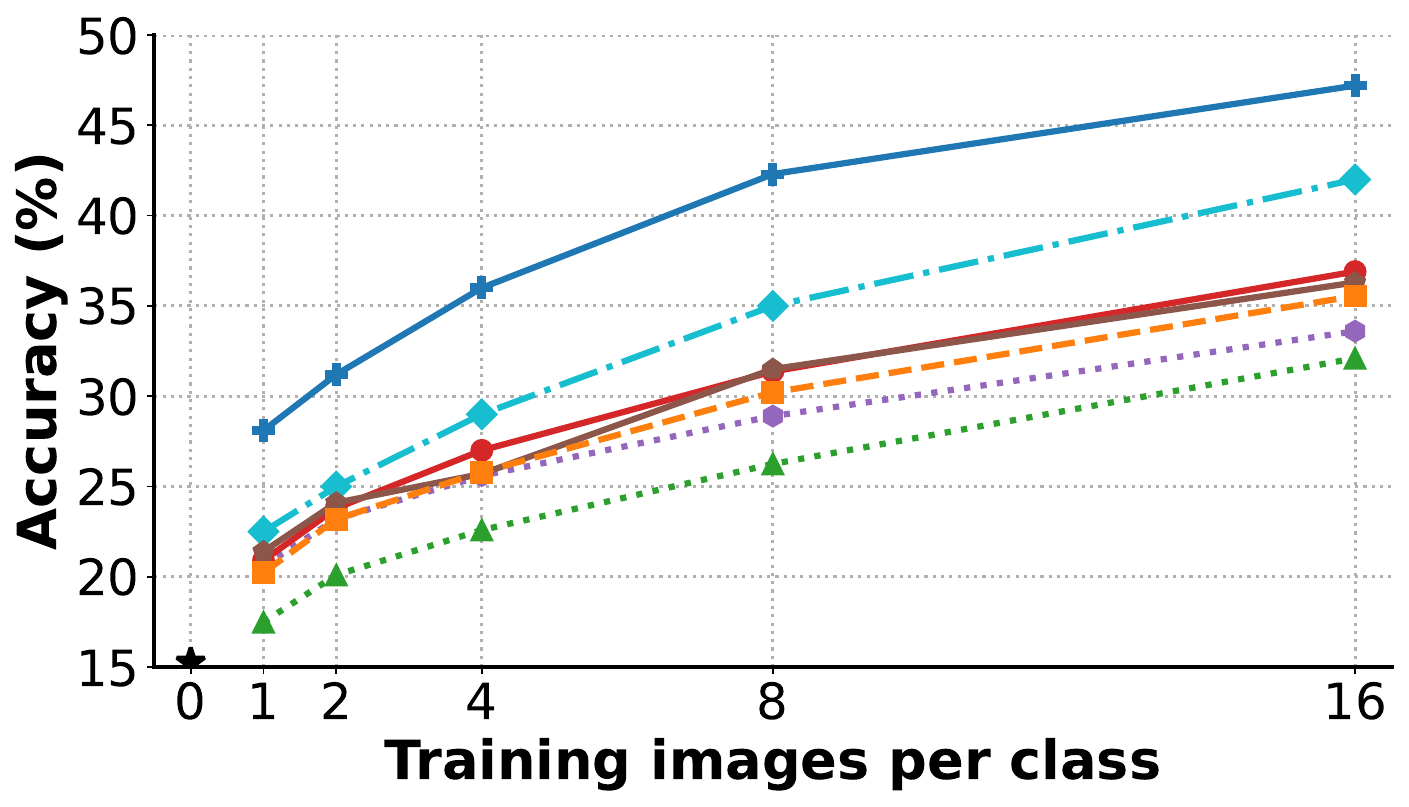}
    \caption{Aircraft}
    \label{fig:aircraft}
  \end{subfigure}
  \hfill
  \begin{subfigure}[b]{0.3\textwidth}
    \includegraphics[width=\textwidth]{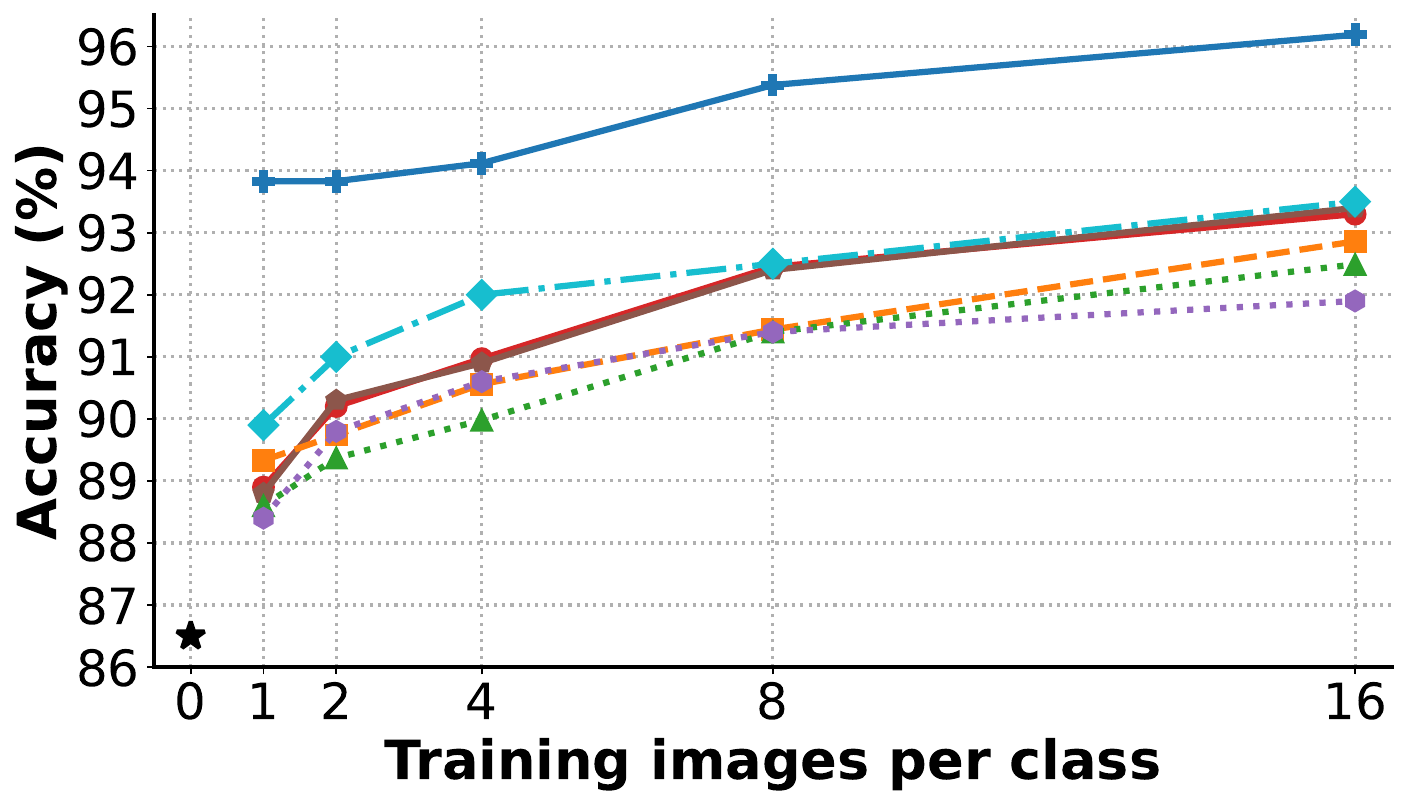}
    \caption{Caltech101}
    \label{fig:caltech101}
  \end{subfigure}
\hfill
  %
  \begin{subfigure}[b]{0.3\textwidth}
    \includegraphics[width=\textwidth]{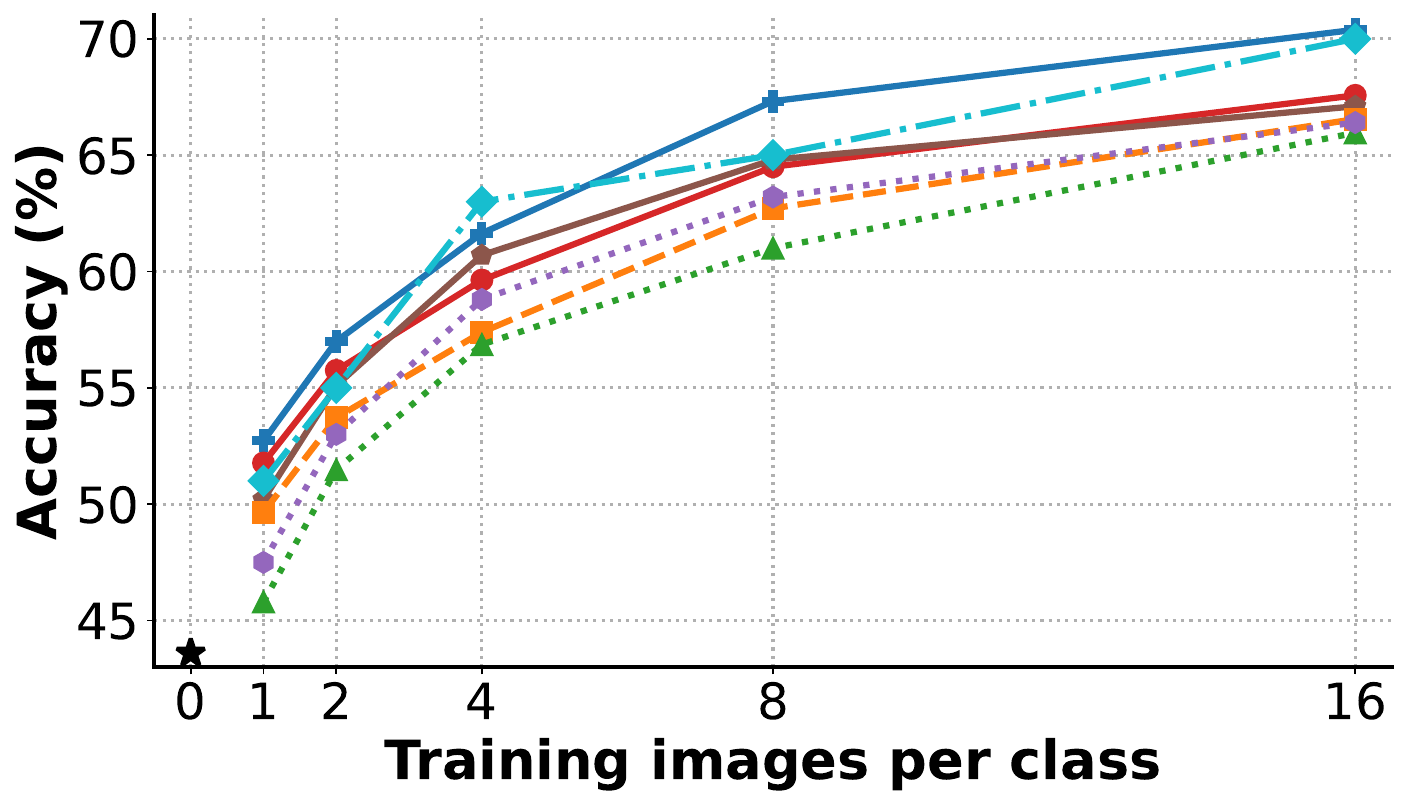}
    \caption{DTD}
    \label{fig:DTD}
  \end{subfigure}
   \hfill 
  \begin{subfigure}[b]{0.3\textwidth}
    \includegraphics[width=\textwidth]{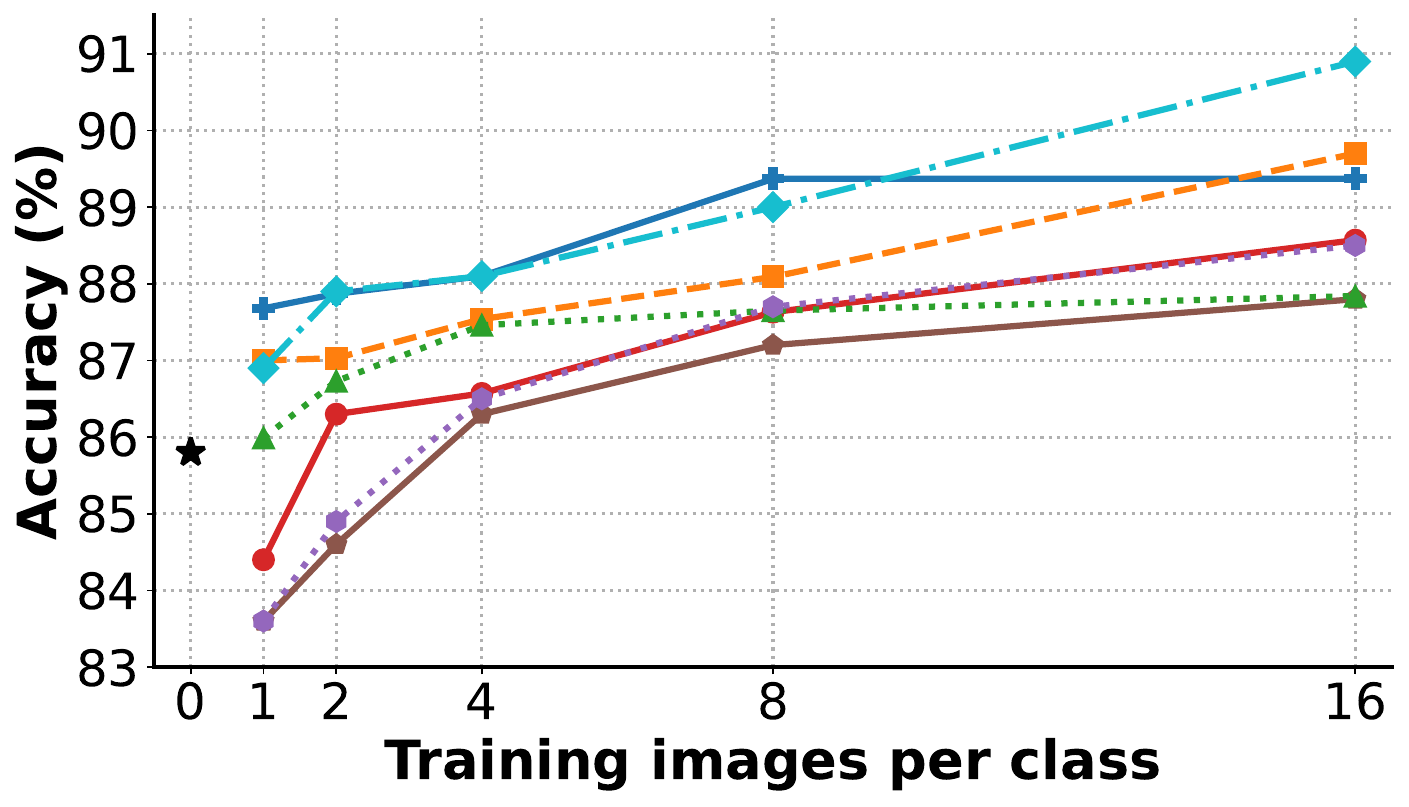}
    \caption{OxfordPets}
    \label{fig:pets}
  \end{subfigure} 
  \hfill
  \begin{subfigure}[b]{0.3\textwidth}
    \includegraphics[width=\textwidth]{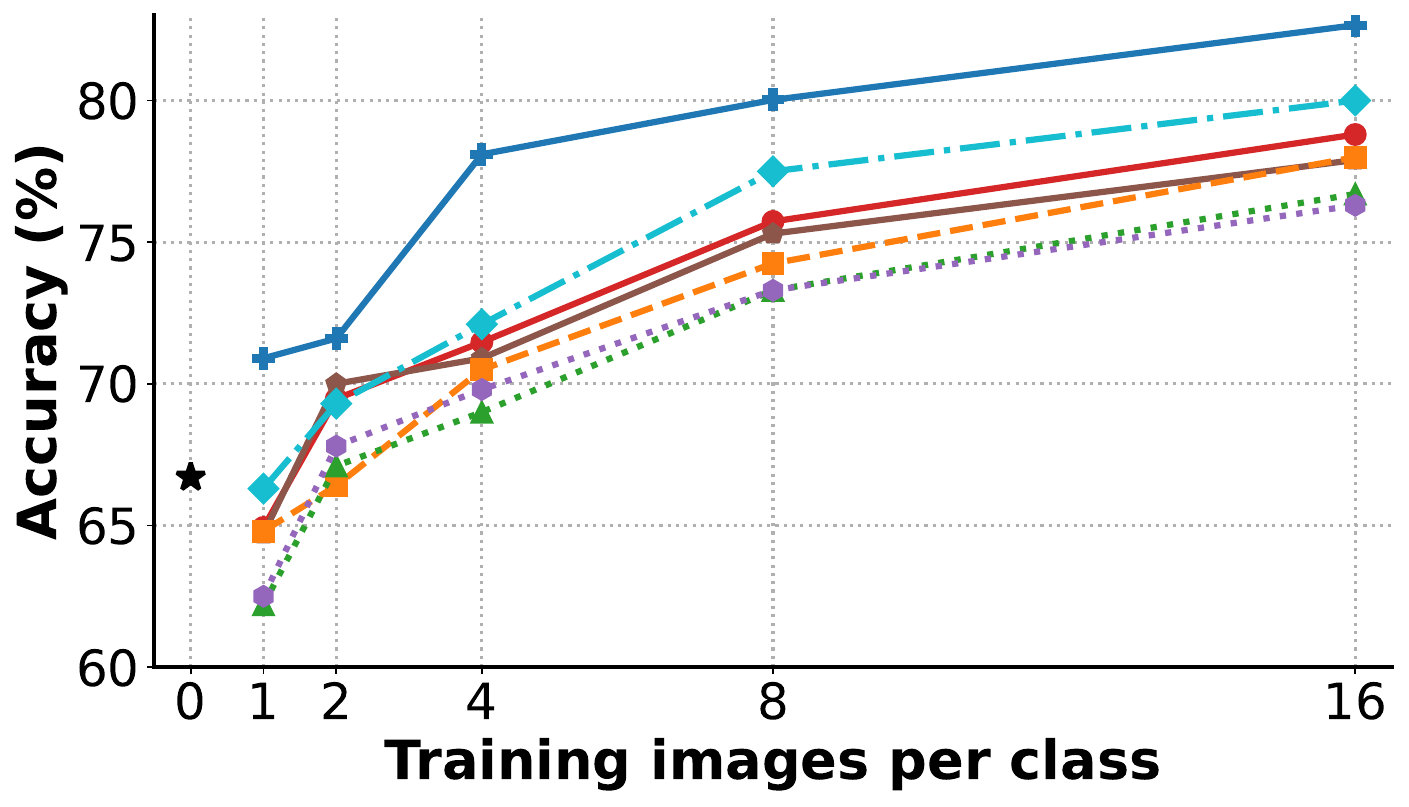}
    \caption{UCF101}
    \label{fig:UCF}
  \end{subfigure}
  \begin{subfigure}[b]{0.3\textwidth}
    \includegraphics[width=\textwidth]{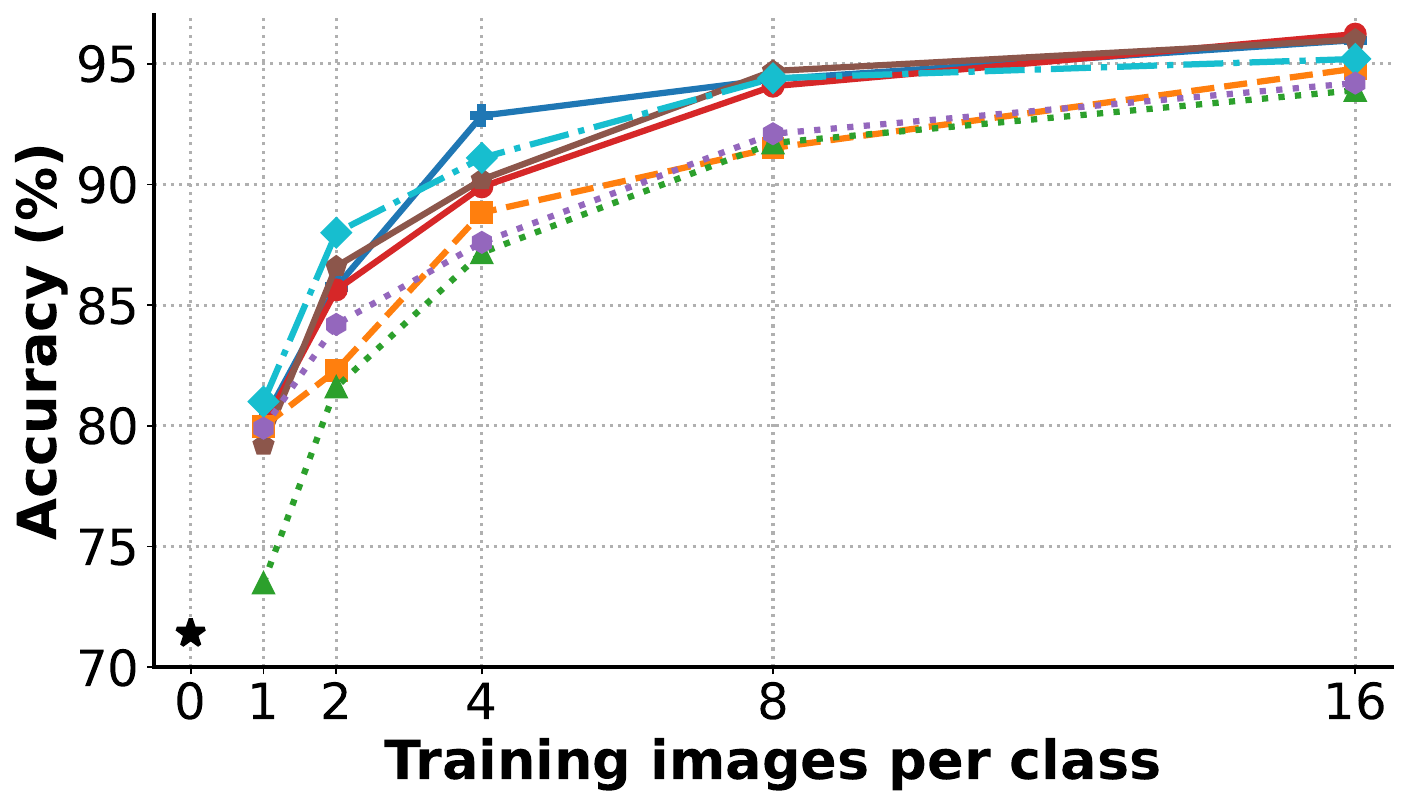}
    \caption{Flowers102}
    \label{fig:flowers}
  \end{subfigure}
  \hfill
  \begin{subfigure}[b]{0.3\textwidth}
    \includegraphics[width=\textwidth]{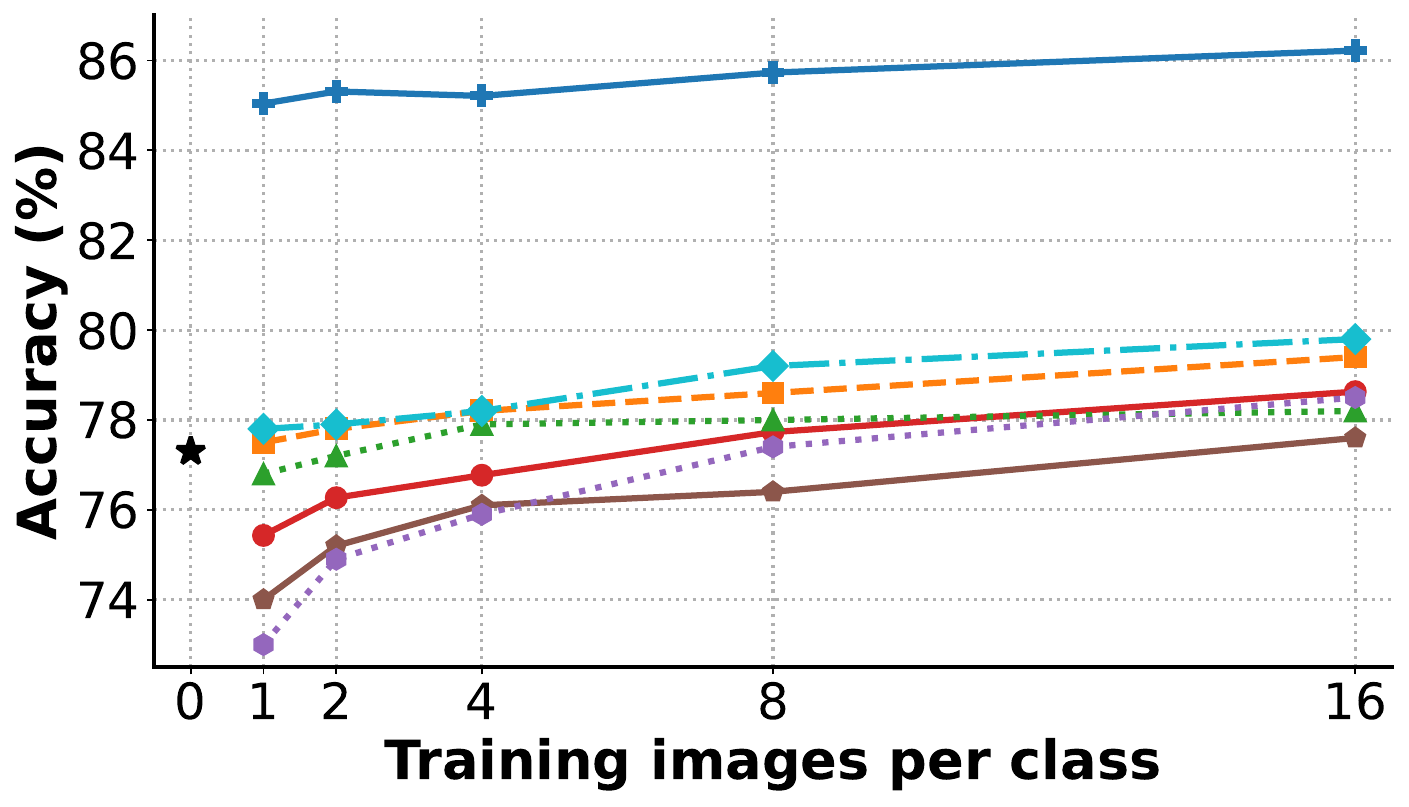}
    \caption{Food101}
    \label{fig:food101}
  \end{subfigure}
    \hfill
  \begin{subfigure}[b]{0.3\textwidth}
    \includegraphics[width=\textwidth]{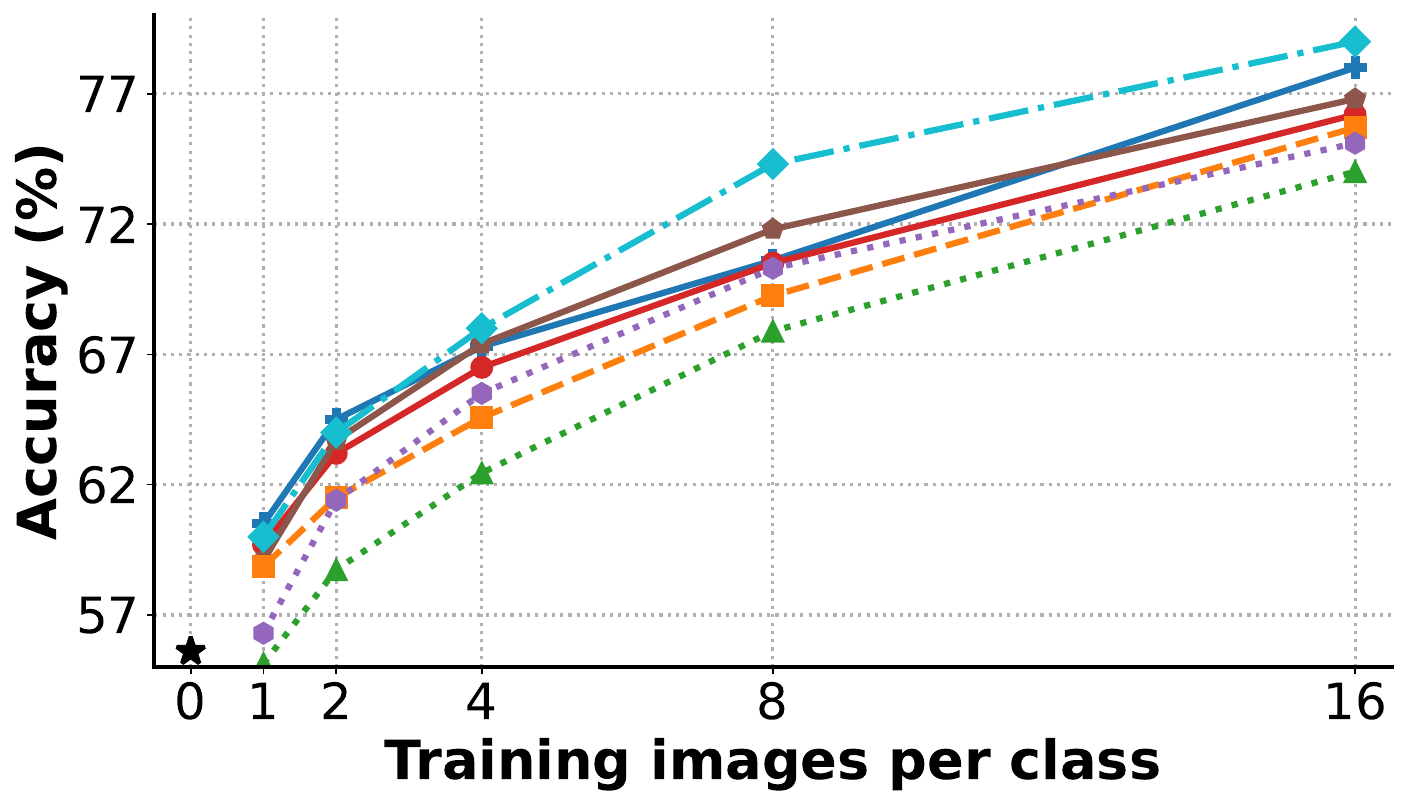}
    \caption{StanfordCars}
    \label{fig:cars}
  \end{subfigure}
  \begin{subfigure}[b]{0.3\textwidth}
    \includegraphics[width=\textwidth]{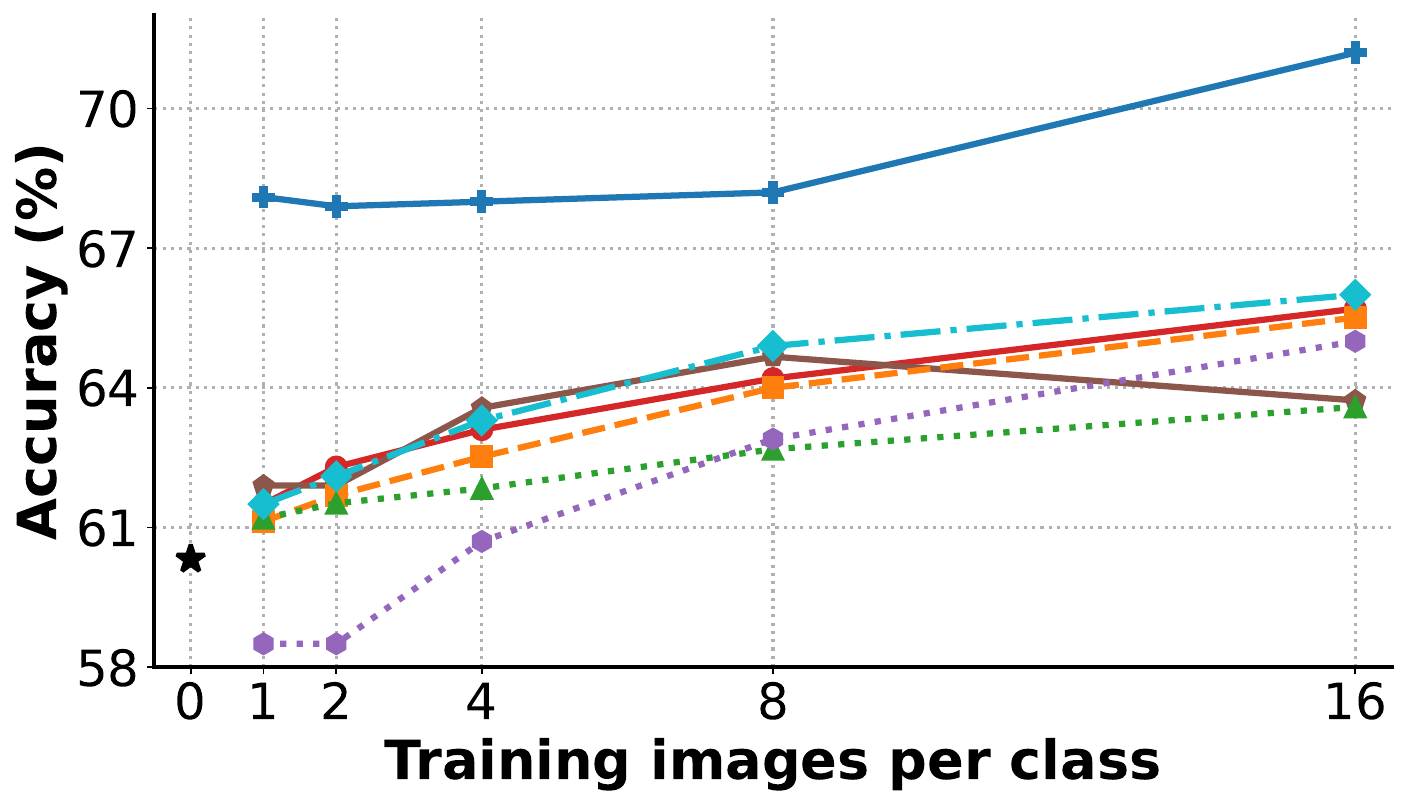}
    \caption{ImageNet}
    \label{fig:imagenet}
  \end{subfigure}
  \hfill
  \begin{subfigure}[b]{0.3\textwidth}
    \includegraphics[width=\textwidth]{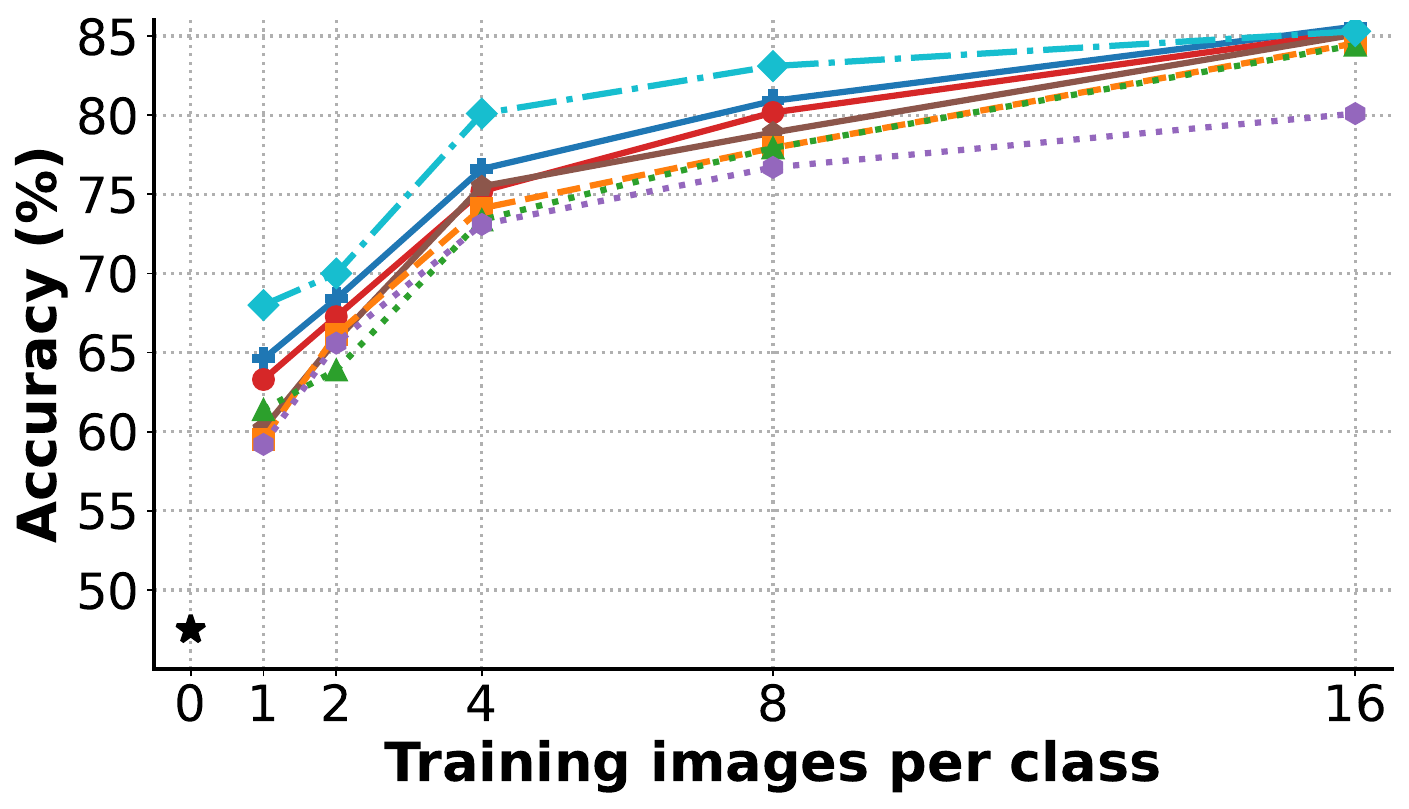}
    \caption{EuroSAT}
    \label{fig:food101}
  \end{subfigure}
  \hfill
    \begin{subfigure}[b]{0.3\textwidth}
    \includegraphics[width=\textwidth]{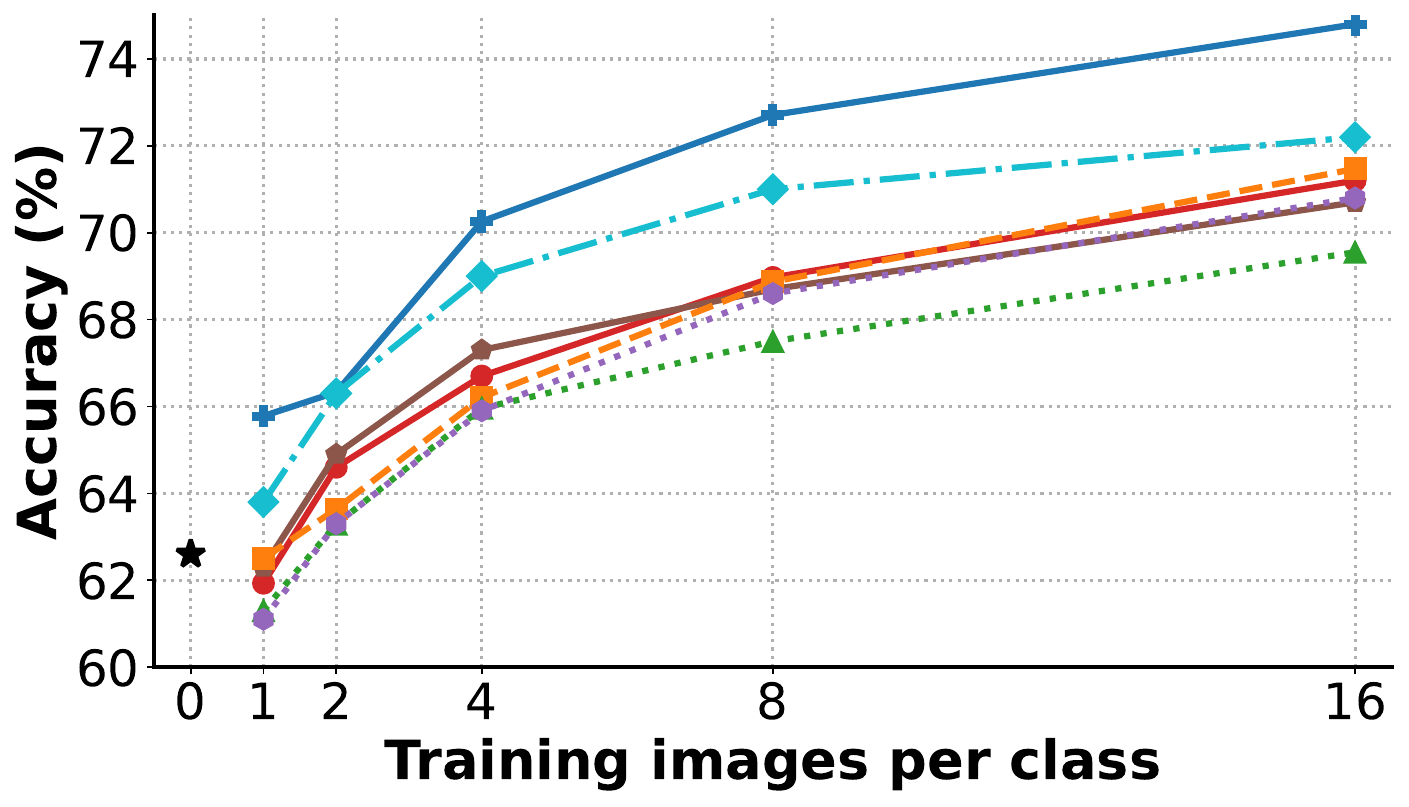}
    \caption{SUN397}
    \label{fig:sun397}
  \end{subfigure}
  \hfill
  \caption{Few-shot accuracy (\%) across \textbf{11 benchmarks datasets}. Our method consistently outperforms state-of-the-art (SOTA) adapter-based approaches and zero-shot CLIP.}
  \label{fig:fewshot}
\end{figure*}

\section{Experiment Results and Discussion} \label{sec:exp}
\paragraph{\textbf{Dataset and Experimental Setup}}
We evaluated our method on 11 standard benchmarks that span diverse visual domains. Aircraft~\cite{maji2013fine}, Flowers102~\cite{nilsback2008automated}, SUN397~\cite{xiao2010sun}, Food101~\cite{bossard2014food}, Caltech101~\cite{fei2004learning}, UCF101~\cite{soomro2012ucf101}, Stanford Cars~\cite{krause20133d}, DTD~\cite{cimpoi2014describing}, ImageNet~\cite{krizhevsky2012imagenet}, and EuroSAT~\cite{helber2019eurosat}. In addition, we evaluate on our newly created \emph{Injured vs.~Uninjured Soldier} dataset for casualty triage (see Sec.~\ref{sec:exp}-(d) for more details). These datasets vary in granularity, context, and visual diversity, offering a comprehensive testbed for a few-shot generalization. We follow standard evaluation protocols~\cite{zhang2022tip, zhou2022learning}, train with 1, 2, 4, 8, and 16 samples per class, and report mean accuracy on three independent runs. All models are optimized with AdamW with an initial learning rate of 0.001, decayed via cosine annealing. The experiments were carried out on a single NVIDIA A100 GPU (48GB).

\paragraph{\textbf{Implementation Details}} We implement our framework in PyTorch, adopting CLIP (ViT-B/16) as the vision backbone and a Transformer for text encoding. During training, the input images are augmented with random resizing, cropping, and horizontal flipping, then resized to $336 \times 336$. Patches are extracted using $(3{\times}3)$ and $(4{\times}4)$ grid-based tiling, where adjacent tiles are combined to generate multi-scale patches (see Fig. \ref{fig:patch_generation}). Each patch is resized to $224 \times 224$ and passed through the frozen CLIP encoder to obtain local features, which serve as graph nodes. The model includes two hyperparameters, $\alpha$ (see Eqn~(\ref{eq:logits_train})) and $\beta$ (see Eqn~(\ref{eq:affinity}))), tuned empirically. All experiments are conducted using the same optimizer settings for fair comparison.
Each dataset is trained independently using its own support and query sets.

\noindent\paragraph{\textbf{Comparison with SotA Methods}}
To assess whether our graph-driven patch-level relationship model and its relational attention mechanism provide a more effective way of steering CLIP toward the target few-shot task, we compare against recent adapter-based approaches, including TIP-Adapter-F~\cite{zhang2022tip}, TaskRes~\cite{yu2023task}, GraphAdapter~\cite{li2024graphadapter}, CLIP-Adapter~\cite{gao2024clip}, CLAP~\cite{silva2024closer}, Ta-Adapter~\cite{zhang2024ta} and CAA~\cite{jiang2025causal}. Table~\ref{tab:results} reports results on 11 benchmarks in five few-shot regimes (1, 2, 4, 8, 16 shots). Across all low-shot settings (1–8 shots), our relational cache consistently achieves the best average accuracy, indicating that enriching the cache with context-aware patch interactions, learned through edge-based relational attention, offers a more discriminative and data-efficient adaptation of CLIP than existing global feature adapters.

In the 1-shot regime, where only a single labeled example is available per class, our approach achieves an average accuracy of 68.8\%, surpassing the strongest prior adapter baseline, CAA (66.3\%), by +2.3\%. The largest gains are observed on Food101 (+7.2\%), ImageNet (+6.6\%), and Aircraft (+4.7\%). These datasets exhibit either fine-grained part-level variations (Food101, Aircraft) or substantial domain shift (ImageNet subsets), where relying solely on global CLIP embeddings is brittle. Here, our patch-level graph explicitly models relationships between local regions, while combined relational attention (\texttt{Attention 1} $\cdot$ \texttt{Attention 2}) selectively emphasizes semantically consistent and structurally relevant patch pairs. This allows the cache to focus on informative local evidence, such as textures, parts, or context, rather than treating all patches equally, which explains the stronger robustness in the extremely low-shot regime.

With increased labeled examples, a similar trend persists. In 2-shot, our method yields an average accuracy of 70.7\%, outperforming CAA (68.9\%), GraphAdapter (67.7\%), and CLAP (66.1\%). The greatest improvements again appear on Aircraft (+4.8\%), Food101 (+7.4\%), and Caltech101 (+2.8\%). Compared to GraphAdapter, which has global image-level or semantic graphs, our intra-image patch graph better captures fine-grained spatial structure. The relational attention module filters noisy or redundant patch interactions and preserves those that are both semantically similar (captured through dot-product similarity) and contextually compatible (captured through learned edge weights). This yields more stable cache keys even when only a few additional examples are available.

In the 4-shot setting, our model reaches 74.1\% average accuracy, clearly exceeding all baselines; for example, it improves upon CAA (72.2\%) by +1.2\%. In Flowers102 (92.9\%) and Caltech101 (94.2\%), the performance nearly saturates with only four labeled samples per class. This indicates that the combination of relational attention and learnable multi-aggregation pooling can rapidly distill patch-level cues into compact, discriminative cache representations, allowing the model to “lock in” the relevant structures of a class from very limited supervision.

At 8-shots, our approach maintains its lead with an average accuracy of 76.7\%, again the highest among all adapter-based methods. It performs particularly well on Pets (89.4\%), Caltech101 (95.4\%), and UCF101 (80.0\%). In these cases, the relational attention mechanism continues to refine patch interactions by down-weighting less informative regions (e.g., background or clutter) and reinforcing consistent pose or part configurations. This shows that the benefits of patch-level relational modeling are not confined to the extreme few-shot setting, but continue to yield stronger representations as modestly more labeled data become available.

In the 16-shot regime, Ta-Adapter achieves the highest average accuracy (83.6\%) by introducing task-aware adapters into both the vision and text branches of CLIP. 
Our method remains competitive with an average of 79.6\%, while still outperforming CAA (77.6\%), GraphAdapter (76.2\%), and TIP-Adapter-F (75.7\%). We also maintain an edge on Caltech101 (96.2\%) and Flowers102 (96.0\%), indicating that the cache refined through graph-based patch relationships and relational attention continues to generalize well when more supervision is available. It is important to emphasize that Ta-Adapter’s gains come from introducing task-aware adapter layers into both the vision and text branches of CLIP that remain active during inference, increasing test-time complexity. In contrast, our framework confines relational reasoning to training and keeps the CLIP backbone frozen, inducing no additional computational overhead at inference.

Overall, across all five few-shot regimes (1, 2, 4, 8, 16 shots), our model consistently ranks first in 1–8 shots and remains among the top performers at 16 shots. The systematic gains over global-feature adapters and graph-based baselines validate our central design: computing cache keys via graph-driven patch-level relationship modeling, equipped with a combined relational attention mechanism and multi-aggregation pooling, is an effective and scalable way to construct discriminative caches for both fine-grained and coarse-grained recognition under limited supervision.

\renewcommand{\thetable}{\arabic{table}}
\begin{table}[t]  
\centering
\caption{Few-shot classification accuracy (\%) across 11 benchmark datasets. Our method consistently outperforms SOTA adapter-based approaches in low-shot settings (1–8 shots) and remains competitive at 16-shot. Best results are shown in \textbf{Bold}. \textbf{Dataset abbreviations:} INet (ImageNet), SUN (SUN397), Air (FGVC-Aircraft),
Euro (EuroSAT), Cars (Stanford Cars), Food (Food101),
Pets (OxfordPets), Flow (Flowers102), Cal (Caltech101),
DTD (Describable Textures), and UCF (UCF101)
}
\resizebox{\textwidth}{!}{
\begin{tabular}{llccccccccccccc}
\toprule
Shots & Method & Venue & INet & SUN & Air & Euro & Cars & Food & Pets & Flow & Cal & DTD & UCF & Avg \\
\midrule
\multirow{1}{*}{0}   & CLIP \textsuperscript{\cite{radford2021learning}} & ICML'22 & 60.3 & 62.6 & 15.3 & 47.5 & 55.6 & 77.3 & 85.8 & 71.4 & 86.5 & 43.6 & 66.7 & 61.1 \\
\midrule
\addlinespace[0.05cm] 
\addlinespace[0.1cm] 
\multirow{5}{*}{1}  & TIP-Adapter-F \textsuperscript{\cite{zhang2022tip}} & ECCV'22 & 61.1 &
62.5 & 20.2 & 59.5 & 58.9 & 77.5 & 87.0 & 80.0 & 89.3 & 49.6 & 64.9 & 64.6 \\
  & TaskRes \textsuperscript{\cite{yu2023task}} & CVPR'23 & 61.9 &
62.3 & 21.4 & 61.7 & 59.1 & 74.0 & 83.6 & 79.2 & 88.8 & 50.2 & 64.8 & 64.3 \\
 &GraphAdapter\textsuperscript{~\cite{li2024graphadapter}} & NeurIPS'23 &61.5 & 61.9 & 20.9 & 63.3 & 59.7 & 75.4 & 84.4 & 80.0 & 88.9 & 51.8 & 64.9 & 64.8
 \\
 & CLIP-Adapter \textsuperscript{\cite{gao2024clip}} & IJCV'24 & 61.2 & 61.3 & 17.5 & 61.4 & 55.1 & 76.8 & 86.0 & 73.5 & 88.6 & 45.8 & 62.2 & 62.7 \\
 & CLAP\textsuperscript{\cite{silva2024closer}} & CVPR'24 & 58.5 & 61.1 & 20.6 & 59.2 & 56.3 & 73.0 & 83.6 & 79.9 & 88.4 & 47.5 & 62.5 & 62.8 \\
 & CAA\textsuperscript{\cite{jiang2025causal}} & ICCV'25 & 61.5 & 63.8 & 22.5 & 68.0 & 60.0 & 77.8 &86.9 & 81.0 & 89.9 & 51.0 & 66.3 & 66.3    \\
 & Proposed & - & 68.1 & 65.8 & 27.2 & 64.6 & 60.5
 & 85.0 & 87.7 & 80.2 & 93.8 & 52.7 & 70.9 & \textbf{68.8} \\
\hline
\addlinespace[0.05cm] 
\addlinespace[0.1cm] 
\multirow{5}{*}{2}  & TIP-Adapter-F \textsuperscript{\cite{zhang2022tip}} & ECCV'22 & 61.7
 & 63.6 & 23.2 & 66.1 & 61.5 & 77.8 & 87.0 & 82.3 & 89.7 & 53.7 & 66.4 & 66.6 \\
  & TaskRes \textsuperscript{\cite{yu2023task}} & CVPR'23 & 61.9 & 64.9 & 24.1 & 65.8 & 63.7 & 75.2 & 84.6 & 86.6 & 90.3 & 55.1 & 70.0 & 67.5 \\
  &GraphAdapter\textsuperscript{~\cite{li2024graphadapter}} & NeurIPS'23 & 62.3 & 64.6 & 23.8 & 67.3 & 63.2 & 76.3
 & 86.3 & 85.6 & 90.2 & 55.7 & 69.5 & 67.7 \\
  & CLIP-Adapter \textsuperscript{\cite{gao2024clip}} & IJCV'24 & 61.5 & 63.3 & 20.1 & 63.9 & 58.7 & 77.2 & 86.7 & 81.6 & 89.4 & 51.5 & 67.1 & 65.5 \\
 & CLAP\textsuperscript{\cite{silva2024closer}} & CVPR'24 & 58.5 & 63.3 & 23.2 & 65.6 & 61.4 & 74.9 & 84.9 & 84.2  & 89.8 & 53.0 & 67.8 & 66.1 \\
  & CAA\textsuperscript{\cite{jiang2025causal}} & ICCV'25 & 62.1 & 66.3 & 25.0 & 70.0 & 64.0 & 77.9 & 87.9 & 88.0 & 91.0 & 55.0 & 69.3 & 68.9       \\
  & Proposed & - & 67.9 & 66.3 & 29.8 & 68.4 & 64.5 & 85.3 & 87.9 & 85.8 & 93.8 & 56.9 & 71.6 & \textbf{70.7}\\
\hline
\addlinespace[0.1cm] 
 \multirow{5}{*}{4}   & TIP-Adapter-F \textsuperscript{\cite{zhang2022tip}} & ECCV'22 & 62.5 & 66.2 & 25.8 & 74.1 & 64.6 & 78.2 & 87.5 & 88.8 & 90.6 & 57.4 & 70.6 & 69.7 \\
  & TaskRes \textsuperscript{\cite{yu2023task}} & CVPR'23 & 63.6
& 67.3 & 25.7 & 73.8 & 67.4 & 76.1 & 86.3 & 90.2 & 91.0 & 60.7 & 70.9 & 70.3 \\
  &GraphAdapter\textsuperscript{~\cite{li2024graphadapter}} & NeurIPS'23 & 63.1 & 66.7 & 27.0 & 75.2 & 66.5 & 76.8  & 86.6 & 89.9  & 91.0 & 59.6 & 71.5 & 70.3\\
  & CLIP-Adapter \textsuperscript{\cite{gao2024clip}} & IJCV'24 & 62.7 & 66.0 & 22.6 & 73.4 & 62.4 & 77.9 & 87.5 & 87.2 & 90.0 & 56.9 & 69.1 & 68.7 \\
  & CLAP \textsuperscript{\cite{silva2024closer}} & CVPR'24 & 60.7 & 65.9 & 25.6 & 73.1 & 65.5 & 75.9 & 86.5 & 87.6 & 90.6 & 58.8 & 69.8 & 69.1 \\
  & CAA\textsuperscript{\cite{jiang2025causal}} & ICCV'25 & 63.3 & 69.0 & 29.0 & 80.1 & 68.0 & 78.2 & 88.1 & 91.1 & 92.0 & 63.0 & 72.1 & 72.2 \\
  & Proposed & - & 68.0
 & 70.3 & 32.7 & 76.6 & 67.3
 & 85.2 & 88.1  & 92.9 & 94.2 & 61.6 & 78.1 & \textbf{74.1} \\ 
\hline
\addlinespace[0.1cm] 
\multirow{5}{*}{8}  & TIP-Adapter-F \textsuperscript{\cite{zhang2022tip}} & ECCV'22 & 64.0
 & 88.9 & 30.2 & 77.9 & 69.2 & 78.6 & 88.1 & 91.5 & 91.4 & 62.7 & 74.2 & 74.2 \\
  & TaskRes \textsuperscript{\cite{yu2023task}} & CVPR'23 &  64.7 & 68.7 & 31.5 & 79.3  & 71.8 & 76.4 & 87.2 & 94.7 & 92.4 & 64.8 & 75.3 & 73.3 \\
  &GraphAdapter\textsuperscript{~\cite{li2024graphadapter}} & NeurIPS'23 & 64.2 & 68.9 & 31.4 & 80.2  & 70.5 & 77.7
 & 87.6 & 94.1  & 92.4  & 64.5 & 75.7 & 73.4 \\
  & CLIP-Adapter \textsuperscript{\cite{gao2024clip}} & IJCV'24 & 62.7 & 67.5 & 26.2 & 77.9 & 67.9 & 78.0 & 87.6
  & 91.7 & 91.4 & 61.0 & 73.3 & 71.4 \\
  & CLAP\textsuperscript{\cite{silva2024closer}} & CVPR'24 & 62.9 & 68.6 & 28.9 & 76.7 & 70.3 & 77.4 & 87.7 & 92.1 & 91.4 & 63.2 & 73.3 & 72.1 \\
  & CAA\textsuperscript{\cite{jiang2025causal}} & ICCV'25 & 64.9 &  71.0 & 35.0 &  83.1 & 74.3 & 79.2 & 89.0 & 94.4 & 92.5 & 65.0 & 77.5 & 75.0\\
  & Proposed & - & 68.2 & 72.7 & 39.7 & 80.9 & 70.6 & 85.7 & 89.4 & 94.4 & 95.4 & 67.3 & 80.0 & \textbf{76.7} \\
\hline
 \addlinespace[0.1cm] 
\addlinespace[0.1cm] 
 \multirow{5}{*}{16}  & TIP-Adapter-F \textsuperscript{\cite{zhang2022tip}} & ECCV'22 & 65.5 & 71.5 & 35.6 & 84.5 & 75.7 & 79.4 & 89.7 & 94.8 & 92.9 & 65.6 & 78.0 &  75.7 \\
  & TaskRes \textsuperscript{\cite{yu2023task}} & CVPR'23 & 63.7 & 70.7 & 36.3 & 84.0 & 76.8 & 77.6 & 87.8 & 96.0 & 93.4 & 67.1 & 78.0 & 75.8 \\
  &GraphAdapter\textsuperscript{~\cite{li2024graphadapter}} & NeurIPS'23 & 65.7 & 71.2 & 36.9 & 85.3 & 76.2 & 78.6 & 88.6 & 96.2 & 93.3 & 67.6 & 78.8 & 76.2 \\
 & CLIP-Adapter \textsuperscript{\cite{gao2024clip}} & IJCV'24 & 63.6 & 69.6 & 32.1 & 84.4 & 74.0 & 78.2 & 87.8 & 93.9 & 92.5 & 66.0 & 76.8 & 74.4 \\
  & CLAP \textsuperscript{\cite{silva2024closer}} & CVPR'24 & 65.0 & 70.8 & 33.6 & 80.1 & 75.1 & 78.5 & 88.5 & 94.2 & 91.9 & 66.4 & 76.3 & 74.6 \\
  & Ta-Adapter	\textsuperscript{\cite{zhang2024ta}} & PR'24 & 74.7	& 77.0	& 54.5 &	91.7	& 86.4 &	87.6	& 93.2 & 	97.9 &	96.4	& 73.6 & 	86.3	& \textbf{83.6} \\
  & CAA\textsuperscript{\cite{jiang2025causal}} & ICCV'25 & 66.0 & 72.2 & 42.0 & 85.3 & 79.0 & 79.8 & 90.9 & 95.2 & 93.5 & 70.0 & 80.0 & 77.6\\
  & Proposed & - & 71.2 & 74.8 & 45.3 & 85.6 & 78.0 & 86.2 & 89.4 & 96.0 & 96.2 & 70.4 & 82.7 & 79.6 \\
\hline
\hline
\end{tabular}
}
\label{tab:results}
\end{table}

\begin{figure}[h]
\centering
\begin{subfigure}[b]{0.16\columnwidth}
    \includegraphics[width=\columnwidth]{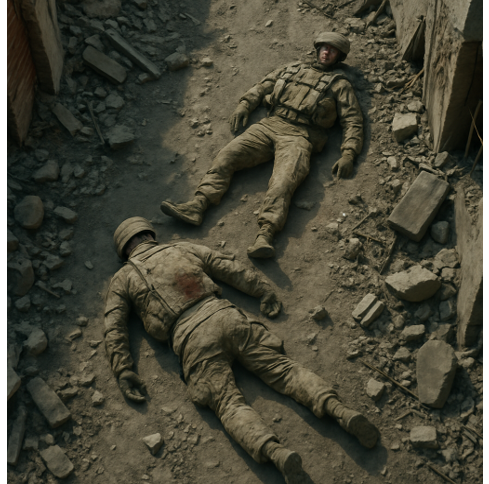}
\end{subfigure} \hspace{0.2cm}
\begin{subfigure}[b]{0.16\columnwidth}
    \includegraphics[width=\columnwidth]{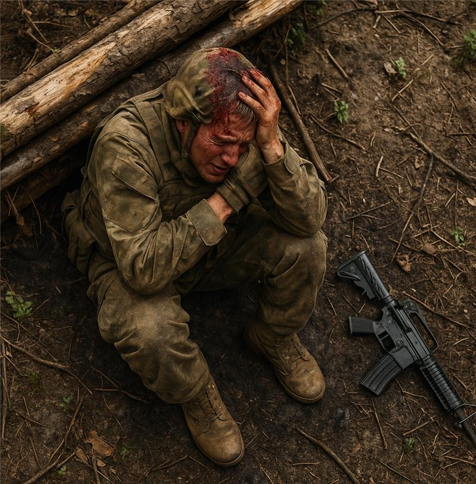}
\end{subfigure} \hspace{0.2cm}
\begin{subfigure}[b]{0.16\columnwidth}
    \includegraphics[width=\columnwidth]{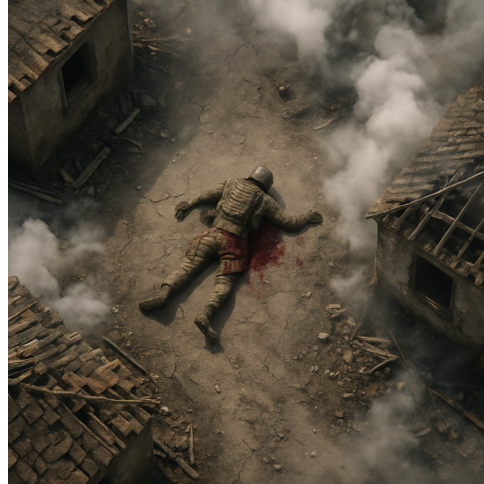}
\end{subfigure} \hspace{0.2cm}
\begin{subfigure}[b]{0.16\columnwidth}
    \includegraphics[width=\columnwidth]{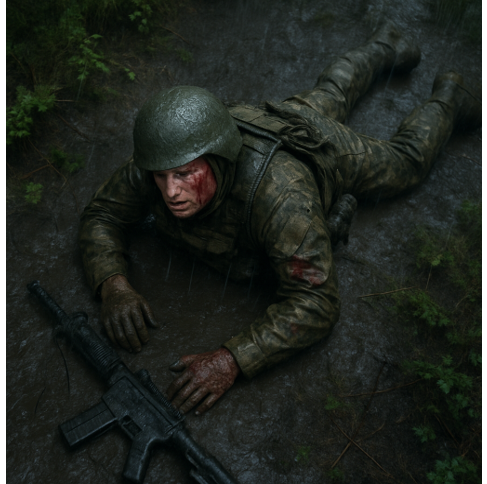}
\end{subfigure} \hspace{0.2cm}

{\centering \footnotesize (a) Injured soldiers}

\begin{subfigure}[b]{0.16\columnwidth}
    \includegraphics[width=\columnwidth]{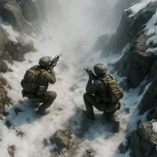}
\end{subfigure} \hspace{0.2cm}
\begin{subfigure}[b]{0.16\columnwidth}
    \includegraphics[width=\columnwidth]{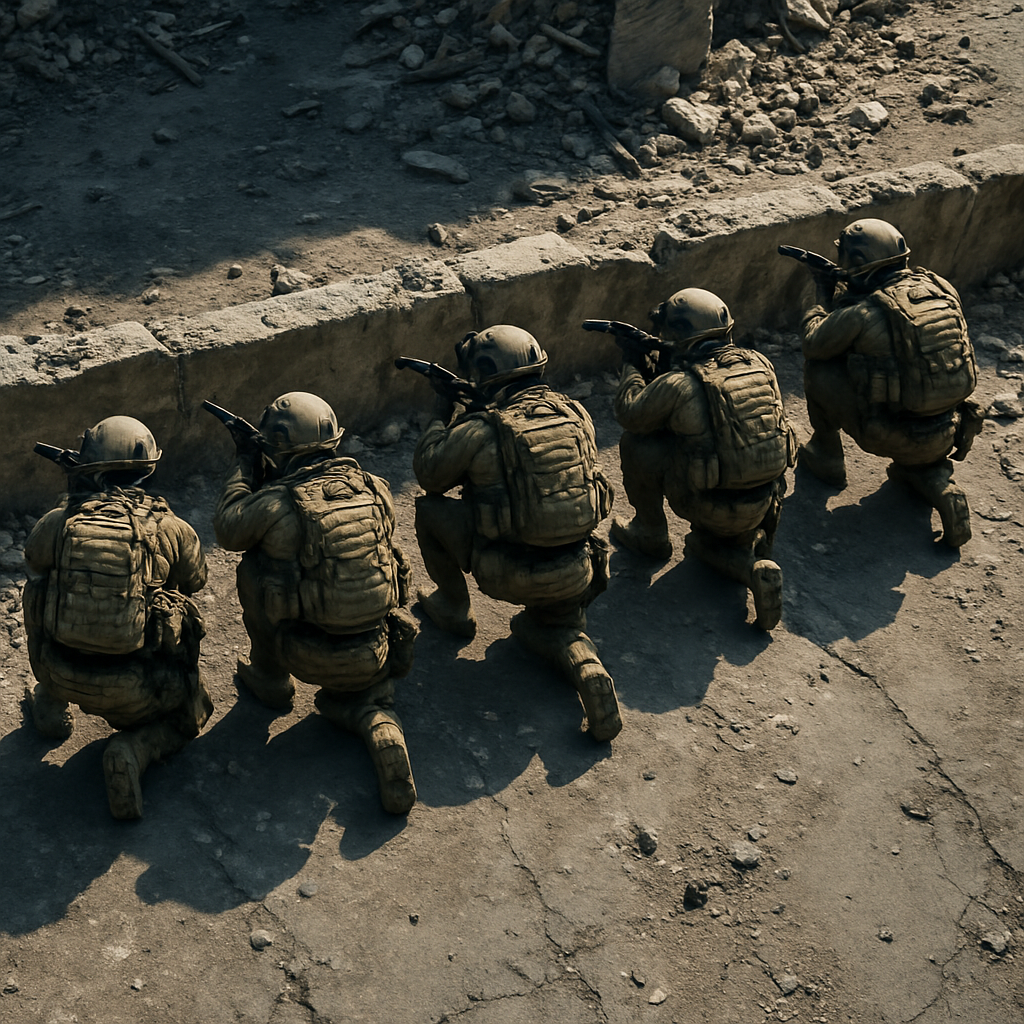}
\end{subfigure} \hspace{0.2cm}
\begin{subfigure}[b]{0.16\columnwidth}
    \includegraphics[width=\columnwidth]{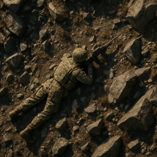}
\end{subfigure} \hspace{0.2cm}
\begin{subfigure}[b]{0.16\columnwidth}
    \includegraphics[width=\columnwidth]{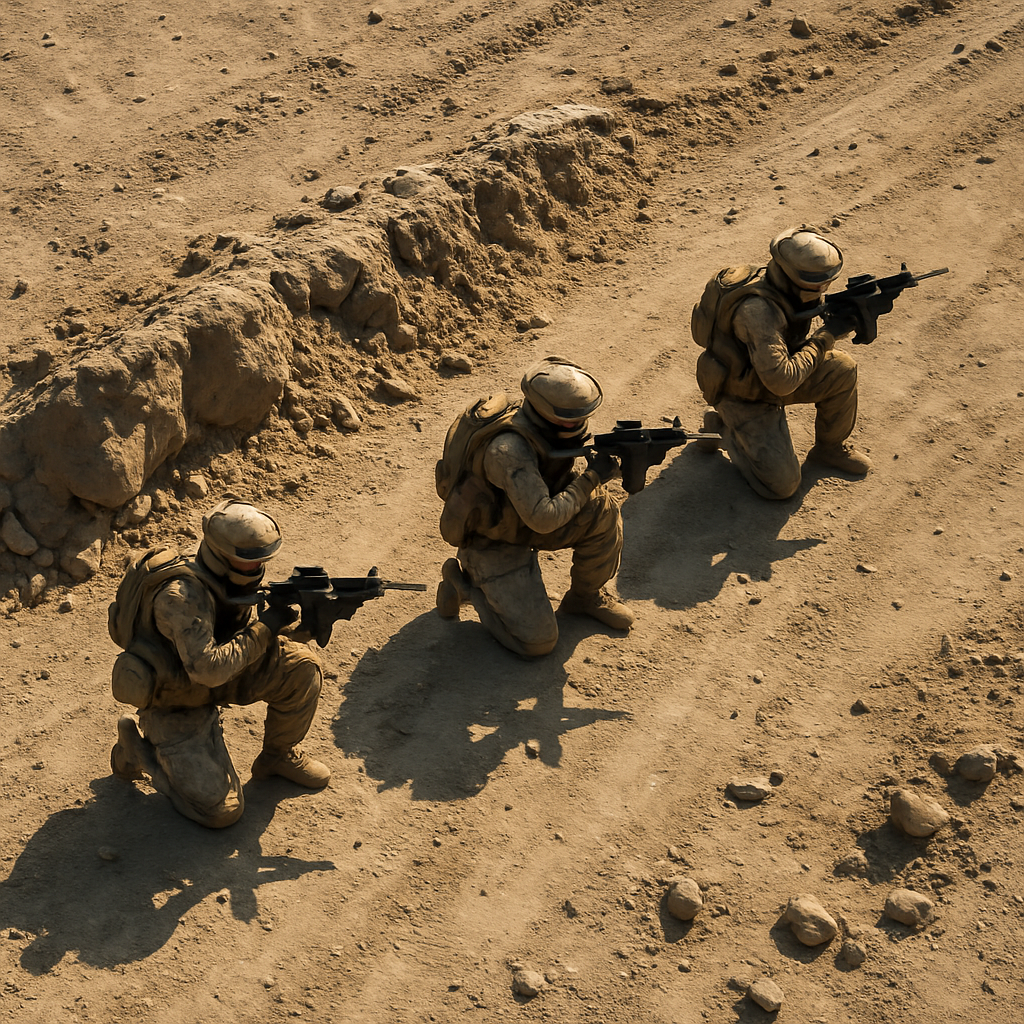}
\end{subfigure} \hspace{0.2cm}

{\centering \footnotesize (b) Uninjured soldiers}\par
\caption{Sample images from our Injured and Uninjured soldier dataset, illustrating the diversity and visual complexity.}
\label{fig:gpt_images}
\end{figure}

\noindent\paragraph{\textbf{Comparison on Our Dataset}}
In an increasingly unstable global landscape, the ability to rapidly identify and prioritize injured personnel is critical for both military operations and humanitarian search-and-rescue missions. Autonomous aerial platforms (e.g., UAVs and drones) are increasingly expected to support this capability, yet training reliable vision models requires data that reflect the realities of operational environments, such as camouflage and protective equipment, debris and clutter, smoke and dust, occlusions, harsh or low lighting, and highly variable terrain. To the best of our knowledge, no such dataset is currently publicly available that explicitly captures casualty-centric scenarios at this level of realism to support research on automated triage and casualty assessment. To help address this gap, we introduce a new \emph{Injured vs. Uninjured} soldiers, designed to stimulate research on casualty recognition and rapid triage under challenging battlefield-like conditions. The dataset contains 6,000 images (3,000 injured and 3,000 uninjured), generated with the GPT-4o image-generation model to cover various plausible combat situations. As illustrated in Fig.~\ref{fig:gpt_images}, the images span a wide range of viewpoints and poses, manifestations of injuries and blood patterns, clothing and protective gear, environmental background, and levels of occlusion and scene clutter. Collectively, these factors create a demanding and practically relevant testbed for evaluating few-shot recognition under domain shift, with direct relevance to time-critical decision support in the ``platinum minutes'' and the broader ``golden hour'' of casualty care.

Table~\ref{tab:our_dataset_main} summarizes the performance of our \emph{Injured vs.~Uninjured Soldier} dataset for 1-, 2-, 4-, 8-, and 16-shot supervision. Across all shots, our patch-level relational approach consistently outperforms Tip-Adapter-F~\cite{zhang2022tip}, highlighting the value of explicitly modeling inter-patch dependencies for casualty recognition in visually cluttered combat scenes. The advantage is most pronounced in the low-shot regime: with only one example per class, accuracy improves from 54.5\% to 67.8\% (+13.3\%), suggesting that the relational cache can extract meaningful injury-related cues even from minimal supervision. At 4 shots, accuracy increases from 74.6\% to 87.1\% (+12.5\%), and with 16-shots our method reaches 94.9\%, exceeding Tip-Adapter-F (85.1\%) by +9.8\%. These gains are particularly relevant for this dataset, where decisions often depend on subtle, highly localized evidence such as blood patterns, limb deformation, damage to protective gear, or collapsed postures, which are frequently observed under camouflage, debris, occlusion, or smoke. By reasoning over patch interactions, our graph-based attention can amplify such region-level signals and compose them into a robust representation, whereas holistic feature adaptation in Tip-Adapter-F tends to wash out these cues within a single global embedding. Overall, the improvements across all $K$-shot settings validate the suitability of our relational cache framework for triage-oriented recognition in challenging battlefield imagery.

\begin{table}[t!]
\centering
\caption{Comparison on our \emph{Injured vs. Uninjured} dataset for $K{=} 1, 2, 4, 8$ and $16$-shots. Best results are shown in \textbf{Bold}.}
\begin{tabular}{lcccccc}
\toprule
\textbf{\# Shots} & \textbf{1} & \textbf{2} & \textbf{4} & \textbf{8} & \textbf{16}  \\
\midrule
TIP-Adapter-F~\textsuperscript{\cite{zhang2022tip}} & 54.5  & 70.5 & 74.6 & 78.8 & 85.1 \\
Proposed   & \textbf{67.8}  & \textbf{75.6} & \textbf{87.1} & \textbf{91.9} & \textbf{94.9} \\
\bottomrule
\end{tabular}
\label{tab:our_dataset_main}
\end{table}

\subsection{Ablation Study}
We perform ablation studies to assess the contribution of each component and justify key design choices, including: a) Impact of Combined Attention for Relational Reasoning, b) Impact of Varying Patch Granularity, c) Effect of Inter-Patch Relational Edges, d) Effect of Different Backbone Architectures, e) Computational Efficiency and Scalability, f) Cache Hyper-parameters Tuning, and  g) Qualitative Analysis of Attention Mechanisms.

\paragraph{\textbf{Impact of Combined Attention for Relational Reasoning:}} We evaluated the effect of different attention mechanisms used in our graph-based relational reasoning module: (i) feedforward attention (\texttt{Attention 1}), (ii) dot-product attention (\texttt{Attention 2}), and (iii) combined attention (see Eqn~(\ref{eq:mx_attention})). Table~\ref{tab:ablation_attn_transposed} indicates that the combined attention variant consistently outperforms both the individual components across the datasets and the number of shots.

On UCF101, combined attention (\texttt{Attention 1}·\texttt{Attention 2}) achieves 70.9\%, 71.6\%, 78.1\%, 80.0\%, and 82.7\% accuracy for the 1-, 2-, 4-, 8-, and 16-shot settings, respectively, surpassing the strongest single-attention variant at every shot. For example, in the 4-shot regime, the combined attention reaches 78.1\%, compared to 76.8\% for \texttt{Attention 1} and 76.3\% for \texttt{Attention 2}. This steady improvement reflects the high intra-class variability and background clutter of UCF101: relying solely on contextual weighting (\texttt{Attention 1}) or solely on feature similarity (\texttt{Attention 2}) is insufficient, whereas their multiplicative interaction reliably suppresses spurious patch connections and preserves only those that are both contextually meaningful and visually aligned. On the fine-grained Aircraft dataset, the combined attention similarly provides consistent gains. In the 1-shot regime, the accuracy improves from 26.6\% (\texttt{Attention 1}) and 25.5\% (\texttt{Attention 2}) to 27.0\% with the combined variant. This advantage carries through all K-shot settings; for example, at 8 shots the combined attention achieves 39.7\%, surpassing 39.1\% (\texttt{Attention 1}) and 39.4\% (\texttt{Attention 2}). Aircraft categories differ only in subtle part-level cues; therefore, the gated design ensures that patch interactions are encouraged only when they satisfy both structural consistency (captured by \texttt{Attention 1}) and strong appearance similarity (captured by \texttt{Attention 2}), leading to more discriminative relational embeddings. Caltech101, being a relatively coarse-grained dataset, shows minimal differences between the attention variants once sufficient labeled examples are available (e.g., 95.9–96.2\% at 16 shots). However, at low shots, combined attention still provides meaningful improvements. In the 1-shot setting, it reaches 93.8\%, outperforming \texttt{Attention 1} (92.2\%) and \texttt{Attention 2} (92.8\%). These gains highlight that even for relatively easier datasets, constructing the cache from only a handful of support images benefits from suppressing noisy or weakly relevant patch interactions, precisely what multiplicative gating achieves. Overall, the ablation results validate our attention design: treating dot-product similarity (\texttt{Attention 2}) as a gating signal over learned contextual compatibility (\texttt{Attention 1}) produces more coherent, context-aware patch embeddings and, consequently, stronger cache keys for few-shot retrieval. 

\begin{table}[t!]
\centering
\caption{Few-shot accuracy (\%) across settings ($K{=}1$–16) on three benchmarks, comparing A1 (\texttt{Attention 1}), A2 (\texttt{Attention 2}) and their combination (A1 $\cdot$ A2) (see Eqn~(\ref{eq:mx_attention})). Best results are shown in \textbf{Bold}.
}
\resizebox{1\columnwidth}{!}{
\begin{tabular}{l|ccc|ccc|ccc}
\toprule
\multirow{2}{*}{\textbf{Shots}} 
& \multicolumn{3}{c|}{\textbf{UCF101}} 
& \multicolumn{3}{c|}{\textbf{Aircraft}} 
& \multicolumn{3}{c}{\textbf{Caltech}} \\
\cmidrule(lr){2-4} \cmidrule(lr){5-7} \cmidrule(lr){8-10}
& A1 & A2 & (A1$\cdot$A2) 
& A1 & A2 & (A1$\cdot$A2)  
& A1 & A2 & (A1$\cdot$A2) \\
\midrule
$1$  & 70.4 & 69.9 & \textbf{70.9} & 26.6 & 25.5 & \textbf{27.0} & 92.2 & 92.8 & \textbf{93.8} \\
$2$  & 70.5 & 70.2 & \textbf{71.6} & 28.9 & 29.0 & \textbf{29.8} & 93.0 & 93.1 & \textbf{93.8} \\
$4$  & 76.8 & 76.3 & \textbf{78.1} & 31.7 & 32.1 & \textbf{32.2} & 93.3 & 94.0 & \textbf{94.1} \\
$8$  & 79.2 & 79.8 & \textbf{80.0} & 39.5 & 39.4 & \textbf{39.7} & 95.2 & 94.9 & \textbf{95.4} \\
$16$ & 82.2 & 81.6 & \textbf{82.7} & 44.7 & 44.1 & \textbf{45.3} & 95.9 & 96.0 & \textbf{96.2} \\
\bottomrule
\end{tabular}
}
\label{tab:ablation_attn_transposed}
\end{table}
\begin{table}[t]
\small
\centering
\setlength{\tabcolsep}{3pt}
\renewcommand{\arraystretch}{1.05}
\caption{Few-shot accuracy (\%) across different patch counts on three benchmark datasets for shot settings $K{=}1$–16. Best results are shown in \textbf{Bold}.}
\begin{tabular}{l|cccccc|cccccc|cccccc}
\toprule
\multirow{2}{*}{\rotatebox{90}{K-Shots}}
& \multicolumn{6}{c|}{\textbf{UCF101}} 
& \multicolumn{6}{c|}{\textbf{Aircraft}} 
& \multicolumn{6}{c}{\textbf{Caltech}} \\
\cmidrule(lr){2-7} \cmidrule(lr){7-12} \cmidrule(lr){12-19}
& 9 & 13 & 17 & 26 & 32 & 36 & 9 & 13 & 17 & 26 & 32 & 36 & 9 & 13 & 17 & 26 & 32 & 36\\
\midrule
$1$  & 69.7 & 70.1 & 70.0 & \textbf{70.9} & 70.0 & 70.0 & 25.5 & 26.0 & 25.8 & \textbf{27.2} & 25.6 & 25.1 &  92.9 & 92.7 & 93.1 & \textbf{93.8} & \textbf{93.8} & 93.0\\
$2$  & 70.9 & 71.1 & 71.1 & \textbf{71.6} & 71.3 & 70.3 & 28.4 & 28.1 & 28.1 & \textbf{29.8} & 29.4 & 28.8 & 92.9 & 93.1 & 93.4 & \textbf{93.8} & 93.0 & 93.2\\
$4$  & 77.3 & 76.5 & 75.1 & \textbf{78.1} & 77.9 & 77.7 & 32.0 & 32.3 & 32.3 & \textbf{32.7} & 32.0 & 31.1 & 93.4 & 93.5 & 93.5 & \textbf{94.2} & 93.5 & 93.7\\
$8$  & 79.6 & 79.7 & 80.0 & \textbf{80.0} & 79.3 & 78.7 & 39.3 & 39.4 & 39.5 & \textbf{39.7} & 39.1 & 38.9  & 94.1 & 94.6 & 94.7 & \textbf{95.4} & 94.3 & 93.8\\
$16$ & 82.2 & 81.9 & 82.2 & \textbf{82.7} & 80.9 & 81.5 & 45.0 & 44.7 & 45.1 & \textbf{45.3} & 44.8 & 42.8  & 94.9 & 95.8 & 95.8 & \textbf{96.2} & 94.9 & 94.4\\
\bottomrule
\end{tabular}
\label{tab:ablation_patches}
\end{table}

\paragraph{\textbf{Impact of Varying Patch Granularity:}} We investigated the effect of varying the number of image patches used as node input for graph reasoning. Table~\ref{tab:ablation_patches} reports results for patch counts from 9 to 36 on UCF101, Aircraft, and Caltech101. In all datasets, accuracy generally improves when moving from coarse partitions (9 patches) to a medium-granularity setting (around 26 patches), after which the performance saturates or slightly decreases with 32–36 patches. On UCF101, 26 patches consistently yield the strongest results. For example, accuracy rises from 69.7\% (9 patches) to 70.9\% (26 patches) at 1-shot, and from 82.2\% to 82.7\% at 16-shots with 9 and 26 patches, respectively. This suggests that moderate granularity provides enough spatial diversity for relational reasoning, whereas too few patches under-represent action cues and too many produce highly correlated nodes that weaken message passing. The Aircraft, which is fine-grained, shows a pronounced optimum at 26 patches. Increasing from 9 to 26 improves 1-shot accuracy from 25.5\% to 27.2\% and 16-shots from 45.0\% to 45.3\%, while 36 patches reduces performance to 25.1\% and 42.8\% at 1- and 16-shots, respectively. The model benefits from having enough patches to isolate subtle part-level differences, but excessive partitioning fragments these cues and introduces redundant edges. Caltech101, being coarse-grained, shows milder improvements, but follows the same pattern. The 26-patch configuration delivers strong performance: 93.8\% at 1-shot and 96.2\% at 16-shot, and is consistently close to the best result at each shot count. Beyond the 26 patch configuration, adding more patches contributes little to new graph structure and mostly increases redundancy. Overall, the results indicate that moderate spatial granularity offers the best balance between local detail and relational coherence, allowing the patch-level graph to capture meaningful structure without being overwhelmed by noisy or redundant nodes.


\begin{table}[t]
\centering
\caption{Accuracy (\%) across all 5-shots without and \textbf{with} inter-patch relationship modeling. \textbf{Bold} indicates fully-connected relational graphs.}
\begin{tabular}{lcccccc}
\toprule
\textbf{K-Shots} & Edges & 1 & 2 & 4 & 8 & 16 \\
\midrule
\multirow{2}{*}{UCF101}  
 & w/o & 60.4  & 63.2  & 65.1  & 68.4 & 69.2 \\
 & w.  & \textbf{70.9} & \textbf{71.6} & \textbf{78.1} & \textbf{80.0} & \textbf{82.7}\\
\hline
\multirow{2}{*}{Aircraft}   
 & w/o & 22.8  & 23.4  & 25.4  & 29.1  & 32.5  \\
 & w.  & \textbf{27.2} & \textbf{29.8} & \textbf{32.7} & \textbf{39.7} & \textbf{45.3} \\
\hline
\multirow{2}{*}{Caltech} 
 & w/o & 73.7  & 73.9  & 74.4  & 74.6 & 76.1  \\
 & w.  & \textbf{93.8} & \textbf{93.8} & \textbf{94.2} & \textbf{95.4} & \textbf{96.2} \\
\bottomrule
\end{tabular}
\label{tab:selfloop_fewshot}
\end{table}

\paragraph{\textbf{Effect of Inter-Patch Relational Edges:}}
We compare our graph-driven approach with and without modeling inter-patch relationships in Table~\ref{tab:selfloop_fewshot}. Removing these relationships (edges) collapses the graph into a transformer-like operation applied independently to each patch, effectively treating patches as i.i.d. tokens with no contextual exchange.
On Aircraft, where subtle part-level cues define classes, removing edges reduces 1-shot accuracy from 27.2\% to 22.8\% and 16-shot performance from 45.3\% to 32.5\%. UCF101 shows even larger drops, reflecting its high variability and clutter: with edges, the model achieves 70.9\% (1-shot) and 82.7\% (16-shot), whereas removing them yields only 60.4\% and 69.2\%. Even Caltech101, although coarse-grained, benefits strongly from relational edges; its accuracy improves from 74.4\% to 94.2\% (+19.8\%) in the 4-shot setting, with consistent gains across all shots. These results collectively indicate that inter-patch edges play a vital role in stabilizing the final embedding space and enabling robust few-shot generalization. The results in Table~\ref{tab:selfloop_fewshot} show that removing these edges disrupts our dual-attention mechanism, which relies on the relational structure to determine meaningful patch interactions: both the gating signal (dot-product similarity) and the contextual weights (feedforward attention) in Eqn~(\ref{eq:mx_attention}) require patch-to-patch connectivity to guide aggregation. Without this relational connectivity, the model cannot reinforce dependencies between visually aligned regions (\texttt{Attention 2}) and their broader structural roles (\texttt{Attention 1}), causing embeddings to lose spatial coherence and ultimately degrading discriminative capability.


\begin{table}[t]
\centering
\caption{GPU memory and training time per epoch for varying patch counts on the Aircraft dataset ($K=16$).}
\begin{tabular}{lcccccc}
\toprule
\textbf{\# Patches} & \textbf{5} & \textbf{9} & \textbf{13} & \textbf{17} & \textbf{26}  \\
\midrule
Peak GPU (MiB) & 1308 & 1308 & 1310 & 1312 & 1314 \\
Time/Epoch (s)       & 95 & 107 & 119 & 142 & 170 \\
\bottomrule
\end{tabular}
\label{tab:resource_usage2}
\end{table}

\begin{table}[t]  
\centering
\caption{Comparison using different backbones of the CLIP visual encoder on Aircraft, Caltech101, UCF101, and our \emph{Injured vs.~Uninjured Soldier} dataset. The results are reported in accuracy (\%) for 1-, 2-, 4-, 8-, and 16-shot settings. Best results are shown in \textbf{bold}.}
\begin{tabular}{@{}llccccc@{}}
\toprule
Shots & & Backbone & Aircraft & Caltech & UCF101 & Soldier \\
\midrule
\addlinespace[0.05cm]
\multirow{4}{*}{1} & & RN50      & 18.5 & 88.7 & 62.8 & 60.2 \\
                    & & RN101     & 19.0 & 89.0 & 63.5 & 66.4 \\
                    & & ViT-B/32  & 23.5 & 90.8 & 70.0 & 61.2 \\
                    & & ViT-B/16  & \textbf{27.2} & \textbf{93.8} & \textbf{70.9} & \textbf{67.8} \\
\hline
\addlinespace[0.05cm]
\multirow{4}{*}{2} & & RN50      & 24.4 & 90.0 & 68.7 & 63.3 \\
                    & & RN101     & 25.5 & 91.8 & 70.0 & 66.9 \\
                    & & ViT-B/32  & 24.6 & 91.9 & 70.1 & 67.3 \\
                    & & ViT-B/16  & \textbf{29.8} & \textbf{93.8} & \textbf{71.6} & \textbf{75.6} \\
\hline
\addlinespace[0.05cm]
\multirow{4}{*}{4} & & RN50      & 24.8 & 88.9 & 71.5 & 73.2 \\
                    & & RN101     & 25.0 & 89.4 & 73.0 & 82.0 \\
                    & & ViT-B/32  & 30.0 & 94.0 & 77.4 & 75.2 \\
                    & & ViT-B/16  & \textbf{32.7} & \textbf{94.2} & \textbf{78.1} & \textbf{87.1} \\
\hline
\addlinespace[0.05cm]
\multirow{4}{*}{8} & & RN50      & 31.7 & 92.0 & 75.5 & 79.3 \\
                    & & RN101     & 33.2 & 93.0 & 77.5 & 83.7 \\
                    & & ViT-B/32  & 31.9 & 93.6 & 76.7 & 80.4 \\
                    & & ViT-B/16  & \textbf{39.7} & \textbf{95.4} & \textbf{80.0} & \textbf{91.9} \\
\hline
\addlinespace[0.1cm]
\multirow{4}{*}{16} & & RN50     & 35.8 & 92.7 & 77.4 & 90.5 \\
                     & & RN101    & 36.5 & 93.0 & 79.6 & 91.1 \\
                     & & ViT-B/32 & 39.5 & 95.5 & 81.0 & 91.0 \\
                     & & ViT-B/16 & \textbf{45.3} & \textbf{96.2} & \textbf{82.7} & \textbf{94.9} \\
\bottomrule
\end{tabular}
\label{tab:backbone_comparison}
\end{table}

\begin{figure*}[htbp]
  \centering
  \begin{subfigure}[b]{0.32\textwidth}
    \includegraphics[width=\textwidth]{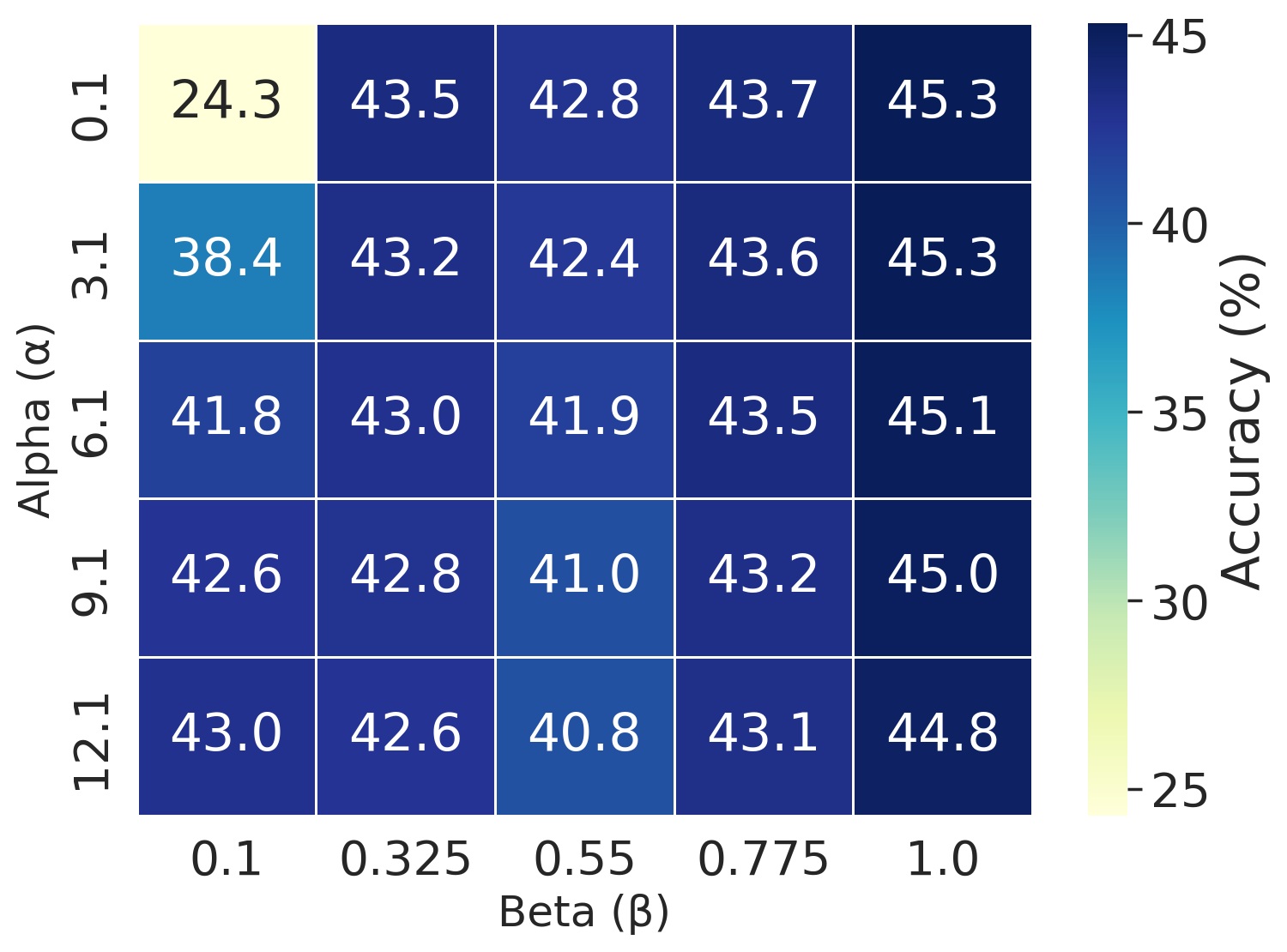}
    \caption{Aircraft}
    \label{fig:aircraft}
  \end{subfigure}\hfill 
   \begin{subfigure}[b]{0.29\textwidth}
    \includegraphics[width=\textwidth]{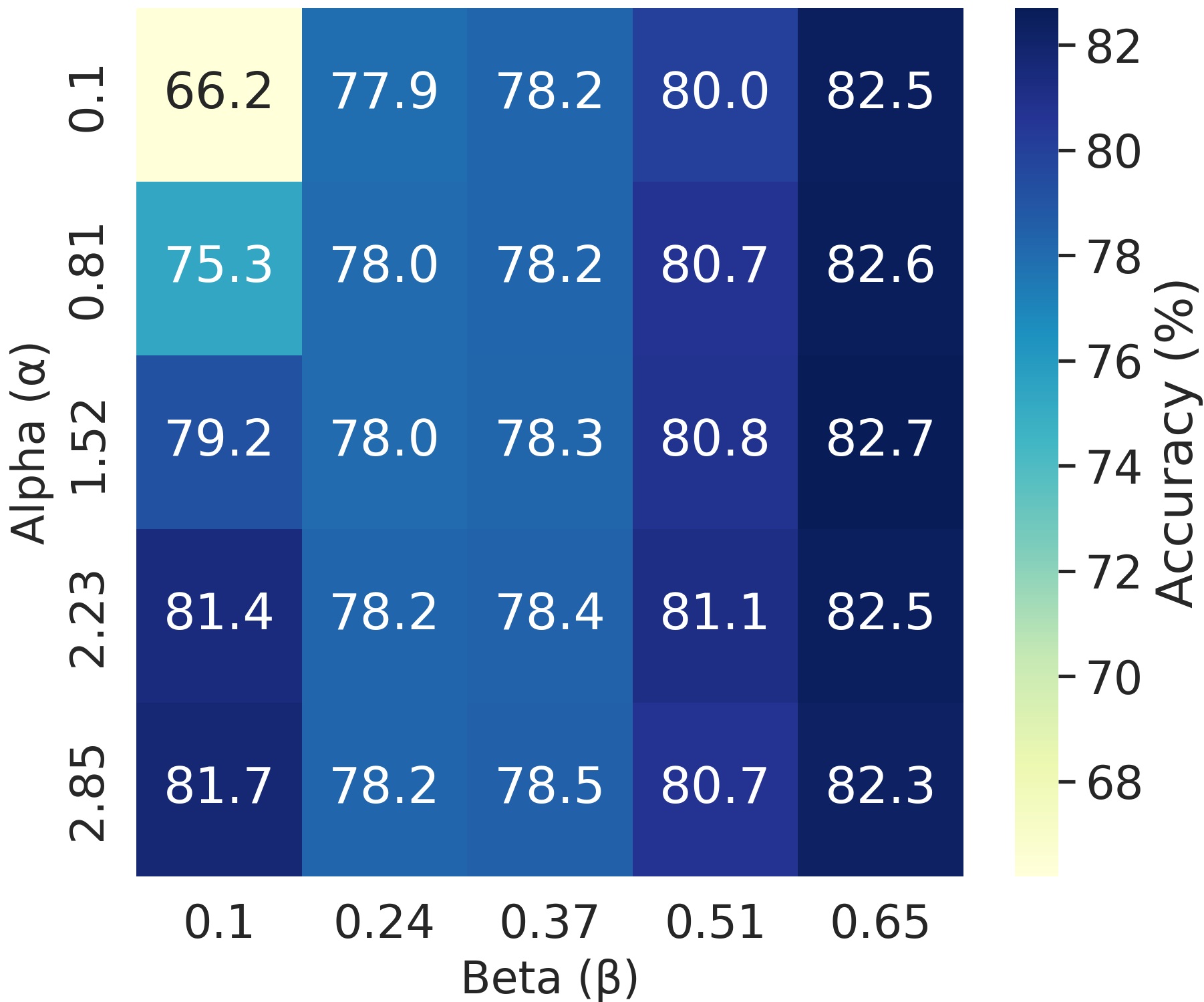}
    \caption{UCF101}
    \label{fig:UCF101}
  \end{subfigure}\hfill 
  \begin{subfigure}[b]{0.32\textwidth}
    \includegraphics[width=\textwidth]{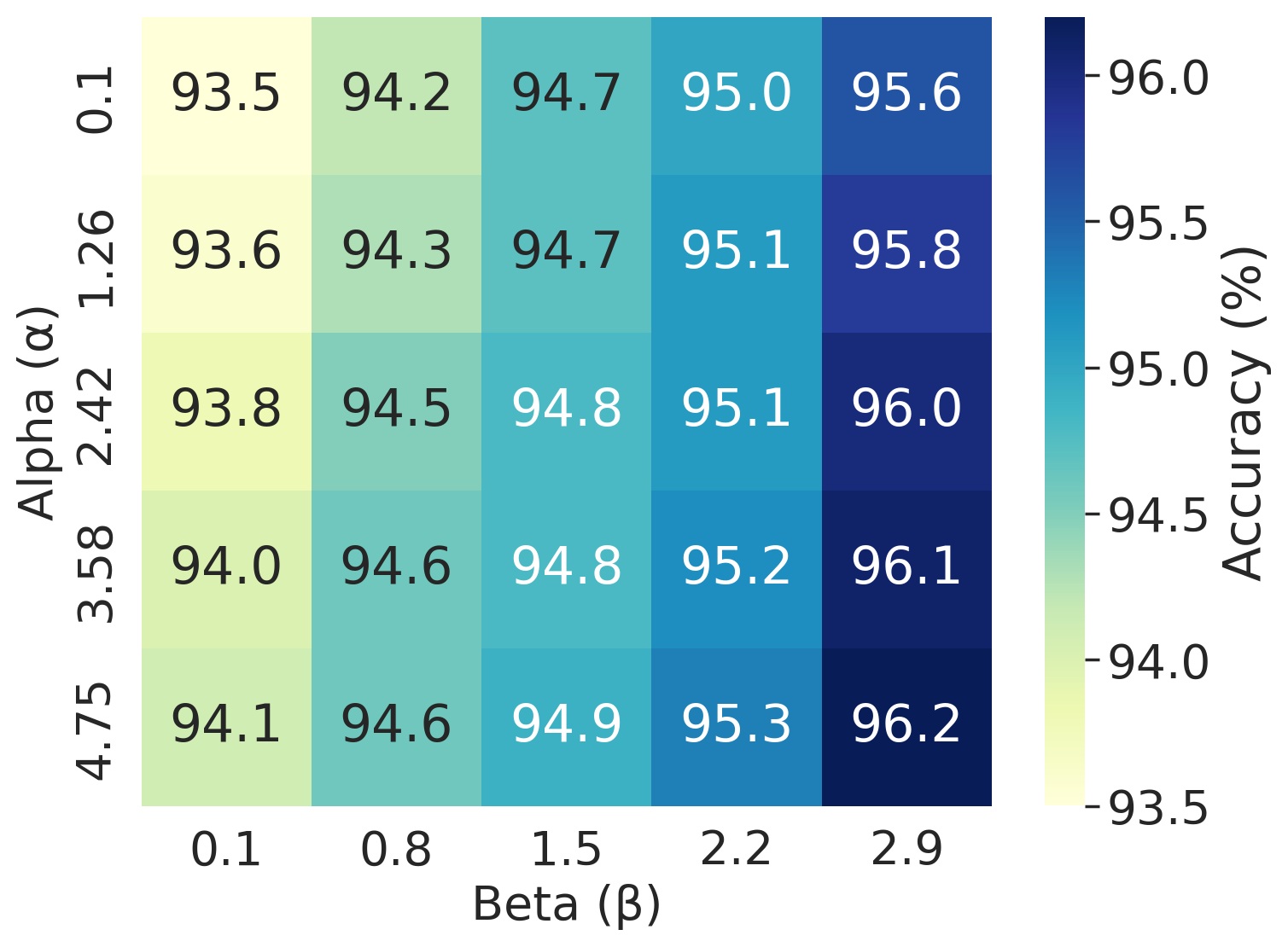}
    \caption{Caltech}
    \label{fig:CalTech}
  \end{subfigure}\hfill 
  \caption{Accuracy heatmaps from grid search over $\alpha$ and $\beta$ on (a) Aircraft, (b) UCF101, and (c) Caltech datasets illustrating the impact of zero-shot priors and affinity sharpness on cache performance.}
  \label{fig:heatmap}
\end{figure*}

\begin{figure*}[!t]
    \centering
    \begin{subfigure}[b]{0.15\textwidth}
        \includegraphics[width=\textwidth]{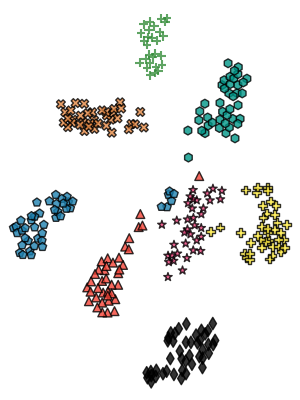}
        \caption{}
    \end{subfigure}
    \hfill
    \begin{subfigure}[b]{0.15\textwidth}
        \includegraphics[width=\textwidth]{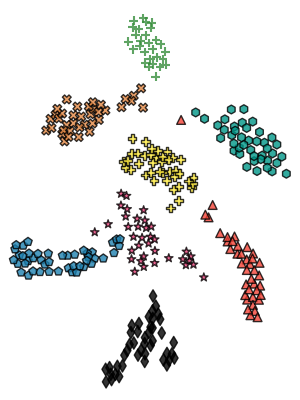}
        \caption{}
    \end{subfigure}
    \hfill
    \begin{subfigure}[b]{0.15\textwidth}
        \includegraphics[width=\textwidth]{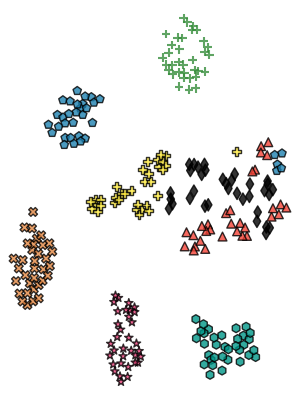}
        \caption{}
    \end{subfigure}
    \hfill
    \begin{subfigure}[b]{0.15\textwidth}
        \includegraphics[width=\textwidth]{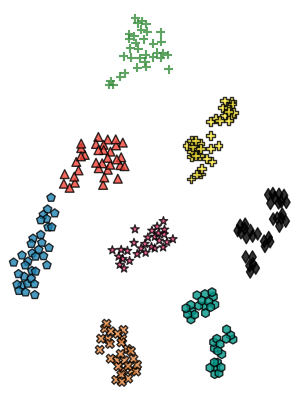}
        \caption{}
    \end{subfigure}

     \begin{subfigure}[b]{0.11\textwidth}
        \includegraphics[width=\textwidth]{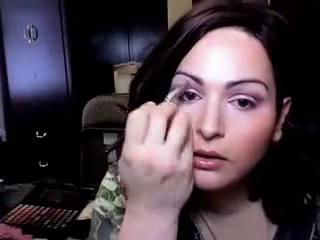}
    \end{subfigure}
    \hfill
    \begin{subfigure}[b]{0.11\textwidth}
        \includegraphics[width=\textwidth]{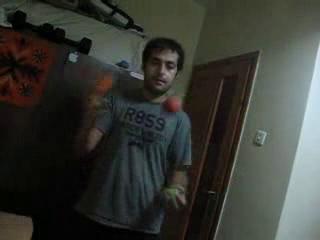}
    \end{subfigure}
    \hfill
    \begin{subfigure}[b]{0.11\textwidth}
        \includegraphics[width=\textwidth]{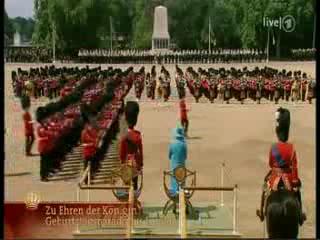}
    \end{subfigure}
        \hfill
     \begin{subfigure}[b]{0.11\textwidth}
         \includegraphics[width=\textwidth]{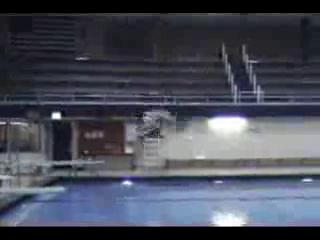}
    \end{subfigure}
        \hfill
     \begin{subfigure}[b]{0.11\textwidth}
        \includegraphics[width=\textwidth]{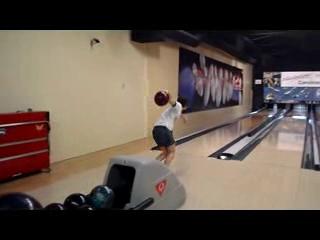}
    \end{subfigure}
        \hfill
    \begin{subfigure}[b]{0.11\textwidth}
        \includegraphics[width=\textwidth]{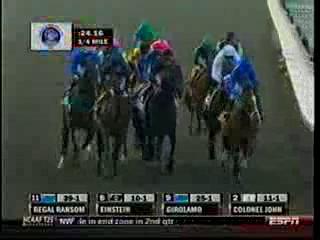}
    \end{subfigure}
        \hfill
    \begin{subfigure}[b]{0.11\textwidth}
        \includegraphics[width=\textwidth]{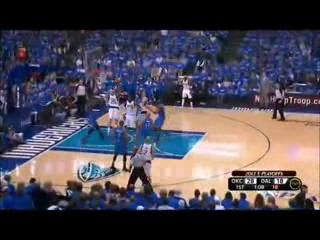}
    \end{subfigure}
        \hfill
     \begin{subfigure}[b]{0.11\textwidth}
        \includegraphics[width=\textwidth]{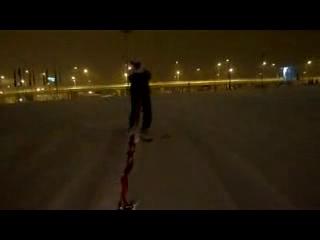}
    \end{subfigure}
    \hfill
    \caption{t-SNE visualizations of logit (see Eqn~(\ref{eq:logits_train})) features from 8 randomly selected UCF101 classes using (a) Tip-Adapter, (b) Tip-Adapter-F, (c) ours with \texttt{Attention 1}, and (d) ours with combined attention (\texttt{Attention 1} $\cdot$ \texttt{Attention 2}). Our method yields more compact and well-separated clusters, especially with combined attention, effectively capturing inter-patch temporal dependencies. Bottom row shows representative video frames illustrating the dataset’s visual diversity.
    }
    \label{fig:t-sne-UCF}
\end{figure*}


\paragraph{\textbf{Effect of Different Backbone Architectures:}}
We further investigate how different CLIP visual encoders influence the performance of our framework. Table~\ref{tab:backbone_comparison} presents results obtained with ResNet-50 (RN50), ResNet-101 (RN101), ViT-B/32, and ViT-B/16 backbones under 1-, 2-, 4-, 8- and 16-shot settings. Across three datasets, a clear upward trend is observed as the backbone capacity increases. On the fine-grained Aircraft dataset, upgrading from ResNet-50 to ViT-B/16 substantially improves performance: 1-shot accuracy increases from 18.5\% to 27.2\%, and 16-shot accuracy from 35.8\% to 45.3\%. This indicates that stronger encoders provide richer patch descriptors capturing subtle part and texture cues that our graph-driven patch-level relationship model can exploit through relational attention, yielding more discriminative cache keys. Caltech101, which is comparatively easier, still benefits from higher-capacity backbones, though gains are smaller due to near-saturation. The accuracy rises from 88.7\% (RN50) to 93.8\% (ViT-B/16) in the 1-shot setting and from 92.7\% to 96.2\% at 16-shot. Here, better backbones mainly refine global category structure, while our relational cache consolidates these stronger patch features into compact class-specific representations. UCF101 shows similar behavior. Moving from RN50 to ViT-B/16 improves 1-shot performance from 62.8\% to 70.9\% and 16-shot performance from 77.4\% to 82.7\%, suggesting that more expressive encoders help capture diverse pose and background cues in action frames. Our inter-patch attention then integrates these cues across patches, producing stable, context-aware embeddings for few-shot classification. Overall, this ablation shows that our graph-based relational cache effectively leverages stronger CLIP backbones: as patch features improve, the relational attention and cache refinement stages convert this added representational power into consistent few-shot gains.

\begin{figure*}[!t]
 \vspace{-0.3cm}
    \centering
    \begin{subfigure}[b]{0.15\textwidth}
        \includegraphics[width=\textwidth]{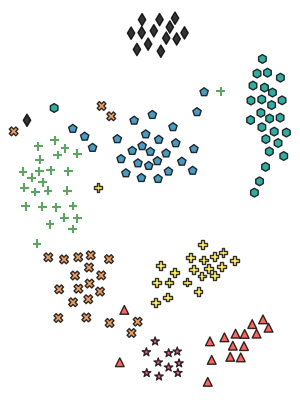}
        \caption{}
    \end{subfigure}
    \hfill
    \begin{subfigure}[b]{0.15\textwidth}
        \includegraphics[width=\textwidth]{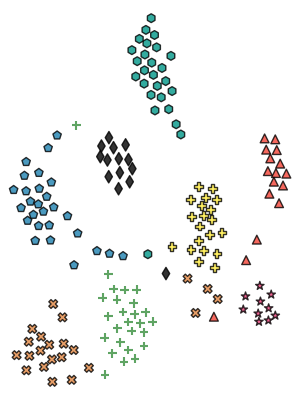}
        \caption{}
    \end{subfigure}
    \hfill
    \begin{subfigure}[b]{0.15\textwidth}
        \includegraphics[width=\textwidth]{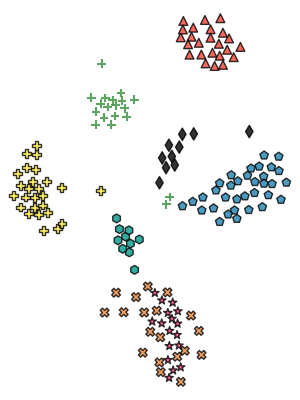}
        \caption{}
    \end{subfigure}
    \hfill
    \begin{subfigure}[b]{0.15\textwidth}
        \includegraphics[width=\textwidth]{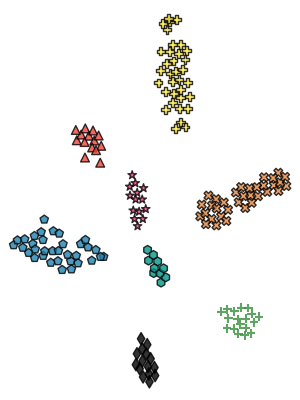}
        \caption{}
    \end{subfigure}
    \vspace{-1mm}   

     \vspace{0.2cm}
     \begin{subfigure}[b]{0.12\textwidth}
        \includegraphics[width=\textwidth]{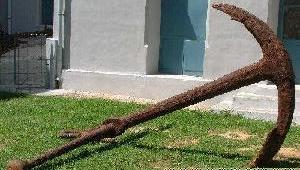}
    \end{subfigure}
    \hfill
    \begin{subfigure}[b]{0.11\textwidth}
        \includegraphics[width=\textwidth]{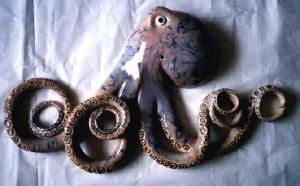}
    \end{subfigure}
    \hfill
    \begin{subfigure}[b]{0.11\textwidth}
        \includegraphics[width=\textwidth]{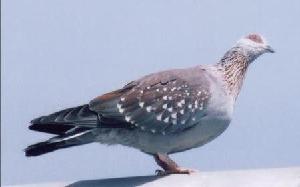}
    \end{subfigure}
    \hfill
     \begin{subfigure}[b]{0.106\textwidth}
         \includegraphics[width=\textwidth]{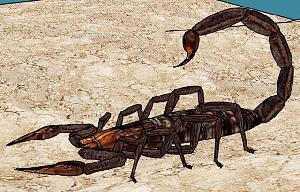}
    \end{subfigure}
    \hfill
     \begin{subfigure}[b]{0.106\textwidth}
        \includegraphics[width=\textwidth]{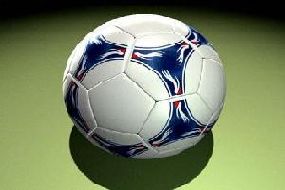}
    \end{subfigure}
    \hfill
    \begin{subfigure}[b]{0.103\textwidth}
        \includegraphics[width=\textwidth]{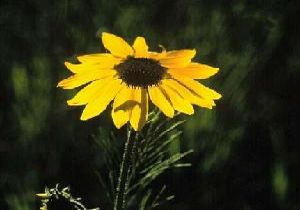}
    \end{subfigure}
    \hfill
    \begin{subfigure}[b]{0.11\textwidth}
        \includegraphics[width=\textwidth]{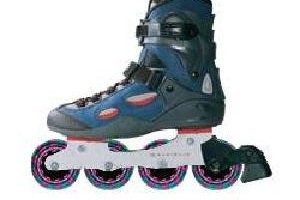}
    \end{subfigure}
    \hfill
     \begin{subfigure}[b]{0.11\textwidth}
        \includegraphics[width=\textwidth]{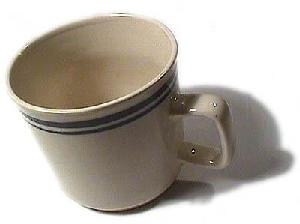}
    \end{subfigure}
    \hfill
    \caption{t-SNE visualizations of logit (see Eqn~(\ref{eq:logits_train})) features from 8 randomly selected Caltech101 classes using (a) Tip-Adapter, (b) Tip-Adapter-F, (c) ours with \texttt{Attention 1}, and (d) ours with combined attention (\texttt{Attention 1} $\cdot$ \texttt{Attention 2}). Despite Caltech101 being a coarse-grained dataset, our combined attention achieves more compact and well-separated clusters, enhancing class discrimination. The bottom row shows representative images from selected classes.}
    \label{fig:t-sne-CalTech}
\end{figure*}

\begin{table}[h]
\centering
\caption{Peak GPU memory and per-epoch training time of Ours vs. Tip-Adapter-F on the 16-shot Aircraft dataset.}
\begin{tabular}{l c ccccc}
\toprule
 & & Peak GPU (MiB)  & Time/Epoch (s)\\
\midrule
 & Ours  & 1314 & 170 \\
& Tip-Adapter-F~\cite{zhang2022tip} & 1296   &  7    \\
\bottomrule
\end{tabular}
\label{tab:resource_usage}
\end{table}
\paragraph{\textbf{Computational Efficiency and Scalability:}} 
We assess the computational trade-offs of our patch-level relational modeling in terms of GPU memory and training time on the Aircraft dataset with a 16-shot setting using an NVIDIA A40 GPU. Table~\ref{tab:resource_usage2} reports resource usage as we vary the number of input patches from 5 to 26. The peak GPU memory remains essentially stable, increasing only slightly from 1308\,MiB to 1314\,MiB as the number of patches grows from 5 to 26, while the training time per epoch rises from 95\,s to 170\,s, exhibiting a near-linear trend with respect to patch count. This indicates that our graph-based module scales gracefully with spatial granularity and does not incur prohibitive memory overhead even at higher patch resolutions. Table~\ref{tab:resource_usage} compares our method with Tip-Adapter-F~\cite{zhang2022tip} at the recommended 26-patch configuration. Modeling pairwise relations across patches leads to a modest increase in peak memory (1314\,MiB vs. 1296\,MiB) but a higher training time per epoch (170\,s vs. 7\,s), reflecting the additional cost of relational reasoning that captures fine-grained structural dependencies. Crucially, this overhead is confined to the training phase. At inference, our approach falls back to standard cache-based retrieval with a frozen CLIP encoder, resulting in test-time latency and memory consumption equivalent to Tip-Adapter-F. Thus, the method remains practical for real-world deployment while offering improved spatial discriminability in few-shot settings.

\paragraph{\textbf{Cache Hyper-parameters Tuning:}} Fig.~\ref{fig:heatmap} presents accuracy heatmaps for the $\alpha$--$\beta$ grid search on three representative datasets: (a) Aircraft, (b) UCF101, and (c) Caltech. These plots illustrate the effect of residual weight ($\alpha$), which balances learned features and zero-shot priors, and affinity sharpness ($\beta$), which controls the selectivity of cache-based similarity. In Aircraft, a fine-grained dataset, performance is highly sensitive to $\alpha$ and less so to $\beta$. The highest accuracy (45.3\%) occurs at $\alpha = 0.1$ and $\beta = 1.0$, where minimal residual bias is retained but sharper affinities aid in class separation. This suggests that relational priors already capture the detailed structure and that excessive residual weight may overfit. In UCF101, a dynamic and moderately complex dataset, accuracy steadily improves with increasing $\alpha$ and $\beta$, reaching 82.7\% for $\alpha = 1.52$ and $\beta = 0.65$. This reflects the need for stronger supervised adaptation and moderately sharp affinities in action domains. For Caltech, a coarse-grained object recognition dataset, the performance is relatively stable across parameter settings but still improves with sharper affinities. The best result (96.2\%) is observed at $\alpha = 4.75$ and $\beta = 2.9$, confirming that cache sharpness is particularly beneficial when intra-class variance is low. These results demonstrate that, while our method is robust in a range of hyperparameters, a modest adjustment of $\alpha$ and $\beta$ yields consistent accuracy gains. To maintain clarity and readability, we limit the visualization to the selected range of $\alpha$ and $\beta$ in Fig. \ref{fig:heatmap}, as accuracy tends to decline beyond these values. The cache mechanism effectively adapts to the complexity and granularity of the dataset, underscoring its flexibility and ease of deployment.


\paragraph{\textbf{Qualitative Analysis of Attention Mechanisms:}}
Figs.~\ref{fig:t-sne-UCF} and~\ref{fig:t-sne-CalTech} provide a qualitative analysis of different attention mechanisms by visualizing the logit features (see Eqn~(\ref{eq:logits_train})) of eight randomly selected classes from UCF101 and Caltech101 using t-SNE. We compare four variants: (a) Tip-Adapter, (b) Tip-Adapter-F, (c) our method with only \texttt{Attention 1}, and (d) our full model with combined attention (\texttt{Attention 1} $\cdot$ \texttt{Attention 2}). Tip-Adapter~\cite{zhang2022tip} and Tip-Adapter-F~\cite{zhang2022tip} produce relatively loose clusters with notable class overlaps, indicating limited discriminability. Using only \texttt{Attention 1} yields more structured embeddings, with tighter clusters and clearer inter-class boundaries, reflecting the benefit of context-aware edge weighting in our patch-level relational graph. The combined attention further sharpens this effect: \texttt{Attention 1} and \texttt{Attention 2} jointly gate patch interactions so that only contextually consistent and visually similar regions exchange information, producing the most compact and well-separated clusters.

In UCF101 (Fig.~\ref{fig:t-sne-UCF}), which exhibits high intra-class variation and background clutter, our full model markedly reduces dispersion within each class and suppresses cross-class overlap, showing that relational attention effectively captures discriminative inter-patch configurations for action categories. In Caltech101 (Fig.~\ref{fig:t-sne-CalTech}), where baseline clusters are already relatively distinct, combined attention still improves compactness and class separation, especially along previously ambiguous boundaries. Overall, these visualizations support our quantitative findings, demonstrating that relational attention strengthens the semantic organization of the feature space and produces more discriminative cache representations.

\section{Conclusion} This paper presented a patch-driven relational graph framework that strengthens cache-based few-shot adaptation of CLIP by injecting structured, intra-image reasoning into the training process. Departing from prior approaches that rely primarily on a single global embedding, our method explicitly models inter-patch dependencies using edge-aware gated attention and a learnable multi-aggregation pooling strategy, yielding more discriminative and context-aware representations for cache retrieval. Importantly, this relational refinement is used only during training and distilled into the cache/adapters, preserving the hallmark advantage of cache-based methods—fast, lightweight inference with no additional test-time overhead beyond standard CLIP and cache lookup.

Extensive experiments across 11 benchmark datasets demonstrate consistent improvements over strong adapter and cache-based baselines, with particularly notable gains in the most challenging regimes: low-shot supervision, fine-grained recognition, and domain-shifted scenarios where decisions depend on subtle, localized evidence. Our ablation studies further validate the contribution of each design choice, including the benefits of relational attention, the role of patch granularity, and the generality of the approach across different CLIP backbones, underscoring the robustness and scalability of the proposed framework across diverse visual domains.

Looking ahead, we see several promising directions. Extending patch-level relational caching to video could enable temporally coherent few-shot recognition with motion-aware relational cues, while broader multimodal extensions may further improve robustness by incorporating complementary signals beyond static imagery. Overall, our findings suggest that relational modeling and cache-based efficiency can be reconciled in a single framework, offering a practical path toward reliable few-shot classification in real-world, time-critical applications.

\section*{Acknowledgment}
This work was supported by UK Research and Innovation (UKRI) - Engineering and Physical Sciences Research Council (EPSRC) under Grant EP/X028631/1 (ATRACT project). The authors thank Benjamin Cave and Kristie Tingley for creating the \emph{Injured vs.~Uninjured Soldier} dataset for this research.



\FloatBarrier

\section*{Supplementary Material}
This supplementary document provides additional results and analyses that complement the main manuscript. Specifically, it includes: (i) additional details on the preparation of our proposed \emph{Injured vs.~Uninjured Soldier} dataset; (ii) further quantitative analysis of the combined attention mechanism on this dataset; and (iii) additional t-SNE visualizations for Flowers102, EuroSAT, DTD, and the Injured vs.~Uninjured Soldier dataset.

\begin{table}[h]
\centering
\caption{Few-shot accuracy (\%) across different patch counts on our \emph{Injured vs. Uninjured soldier} dataset for shot settings $K{=}1$–16. Best results are shown in \textbf{Bold}.}
\setlength{\tabcolsep}{4pt}
\renewcommand{\arraystretch}{1.2}
{
\begin{tabular}{l |cccccc }
\toprule
\multirow{1}{*}{\textbf{K-Shots} }
& \textbf{9} & \textbf{13} & \textbf{17} & \textbf{26} & \textbf{32} &\textbf{36}   \\
\midrule
$1$  & 61.8 & 65.2 & 65.9 & \textbf{67.8} & 67.2 & 67.0 \\
$2$  & 65.2 & 66.4 & 67.2 & \textbf{75.6} & 69.1 &  68.8 \\
$4$  & 84.0 & 84.3 & 84.8 & \textbf{87.1} & 82.4 &  82.1 \\
$8$  & 83.4 & 85.3 & 85.8 & \textbf{91.9} & 84.8 &  84.4 \\
$16$ & 92.3 & 92.5 & 94.0 & \textbf{94.9} & 93.0 & 93.0 \\
\bottomrule
\end{tabular}
}
\label{tab:ablation_patches_injured}
\end{table}

\paragraph{Additional Details for Injured vs.~Uninjured Dataset Preparation:} 
To complement the description in the main paper, we provide further details on the Injured vs.\ Uninjured soldier dataset. Our dataset contains 3000 injured and 3000 uninjured soldier images generated using OpenAI’s image-generation API. All images were generated using OpenAI's image-generation API following a structured prompt for randomly varying the attributes, like \emph{location}, \emph{weather}, \emph{environmental effects}, \emph{altitude}, \emph{lighting}, and \emph{rendering style}. This approach introduces broad diversity in environmental context, viewpoint, and atmospheric appearance while maintaining the consistent image quality and scene composition.

The generation pipeline specifies a high-altitude surveillance perspective and varies factors such as terrain (e.g., \emph{field, forest, desert, riverbank}), weather (e.g., \emph{clear sky, foggy weather, light rain, snowfall}), and environmental effects (e.g., \emph{dust and smoke rising, haze obscuring details, heatwave distortion}). Injured-soldier prompts additionally include textual cues describing distress, limited movement, or harsh environmental struggle, while uninjured-soldier prompts omit injury-related cues but retain the same scene-level randomness. Soldier count was sampled from a fixed range, with a maximum of six soldiers specified in the template; although the API could theoretically generate deviations, no images exceeding this count were observed in practice.

Because each scene attribute is sampled independently, the dataset naturally covers a broad range of visual variations, including changes in lighting, weather, terrain texture, and different forms of environmental occlusion. This diversity results in substantial intra-class variation for both injured and uninjured categories. Consistent with the main paper Sec.~IV-d and Supplementary Tables \ref{tab:ablation_patches_injured}–\ref{tab:injured_attention}, the task demands sensitivity to localized cues (e.g., blood patterns, limb deformation, collapsed posture) while preserving robustness to strong contextual variance. The structured yet stochastic prompting strategy thereby provides a challenging benchmark for evaluating few-shot generalization under realistic battlefield conditions. We will release more details on the data generation at \textit{\url{github.com/tasveerahmad/Patch-Relational-Graph-Attention}}

\paragraph{Additional Quantitative Results for Table IV:}
Table~\ref{tab:ablation_patches_injured} presents the effect of varying the number of patches on our \emph{Injured vs.~Uninjured Soldier} dataset. Across all shot settings, we observe that increasing patch granularity improves performance up to an optimal point, after which gains saturate or slightly decline. In particular, using 26 patches consistently yields the best results across all shot settings, achieving 67.8\% (1-shot), 75.6\% (2-shot), 87.1\% (4-shot), 91.9\% (8-shot), and 94.9\% (16-shot). Compared to coarser patch grids, this setting provides stronger localization for injury evidence while still preserving coherent relational aggregation. For example, at 1 shot accuracy rises from 61.8\% (9 patches) to 67.8\% (26 patches), and at 16 shots from 92.3\% to 94.9\%.

This trend aligns with the visual nature of the task: injury recognition in cluttered battlefield imagery often depends on subtle, spatially localized cues (e.g., blood patterns, posture collapse, or limb deformation) that can be partially occluded by debris or smoke. Increasing the number of patches up to a moderate level allows our graph-driven relational module to better isolate and connect these informative regions through combined attention, producing more discriminative cache keys. However, overly fine patching (e.g., 32 or 36 patches) can introduce redundant or noisy nodes, which weakens message passing and slightly reduces accuracy (e.g., 67.2\%/67.0\% at 1 shot and 93.0\%/93.0\% at 16 shots). Overall, these results confirm that a moderate patch granularity provides the best trade-off between spatial detail and relational coherence for few-shot cache learning on this dataset.

\begin{table}[h]
\centering
\caption{Few-shot accuracy (\%) across settings ($K{=}1$–16) on our \emph{Injured vs. Uninjured soldier} dataset, comparing A1 (\texttt{Attention 1}), A2 (\texttt{Attention 2}) and their combination (A1 $\cdot$ A2) (see Eqn~(2)). Best results are shown in \textbf{Bold}.
}
\setlength{\tabcolsep}{6pt}
{
\begin{tabular}{l|ccc}
\toprule
\textbf{Shots} & A1 & A2 & (A1$\cdot$A2)  \\
\midrule
$1$  & 54.4 & 56.4  &  \textbf{67.8}  \\
$2$  & 65.1 & 60.1 &  \textbf{75.6}  \\
$4$  & 85.3 & 74.6 &  \textbf{87.1} \\
$8$  & 87.2 & 86.1 &  \textbf{91.9}  \\
$16$ & 94.0 & 92.5 &  \textbf{94.9} \\
\bottomrule
\end{tabular}
}
\label{tab:injured_attention}
\end{table}


\paragraph{Additional Quantitative Results for Table III:}
Table~\ref{tab:injured_attention} further analyzes the impact of our attention design on the \emph{Injured vs.~Uninjured Soldier} dataset by comparing \texttt{Attention 1}, \texttt{Attention 2}, and their multiplicative combination (\texttt{Attention 1}$\cdot$\texttt{Attention 2}). The combined attention consistently achieves the best performance across all shot settings, showing that injury recognition benefits from jointly modeling contextual compatibility and appearance similarity between patches. In the most challenging 1-shot regime, accuracy improves from 54.4\% (\texttt{Attention 1}) and 56.4\% (\texttt{Attention 2}) to 67.8\% with \texttt{Attention 1}$\cdot$\texttt{Attention 2}, indicating that neither component alone is sufficient to reliably capture sparse injury evidence from a single example. The same trend holds at 2 shots, where the combined attention reaches 75.6\%, compared to 65.1\% and 60.1\% for \texttt{Attention 1} and \texttt{Attention 2}, respectively. As supervision increases, the combined design continues to provide consistent gains, e.g., 87.1\% at 4 shots and 91.9\% at 8 shots, ultimately reaching 94.9\% at 16 shots.

These improvements are aligned with the visual nature of the dataset: distinguishing injured from uninjured soldiers often depends on subtle, localized cues (e.g., blood patterns, limb deformation, posture changes) that may appear under occlusion, camouflage, or background clutter. \texttt{Attention 2} alone can capture local similarity but may over-connect visually similar yet irrelevant regions, while \texttt{Attention 1} alone provides contextual weighting but may not sufficiently gate interactions by appearance. Their multiplicative combination acts as a gated mechanism that preserves patch interactions that are both visually aligned and contextually plausible, yielding more coherent patch embeddings and, in turn, stronger cache representations for few-shot triage.

\begin{figure*}[ht]
    \centering

    \begin{subfigure}[b]{0.22\textwidth}
        \includegraphics[width=\textwidth]{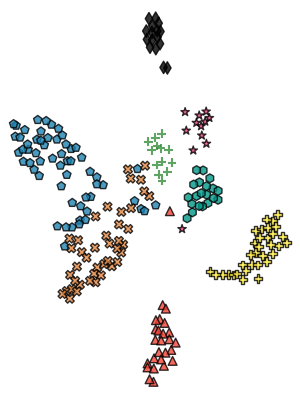}
        \caption{}
    \end{subfigure}
    \hfill
    \begin{subfigure}[b]{0.22\textwidth}
        \includegraphics[width=\textwidth]{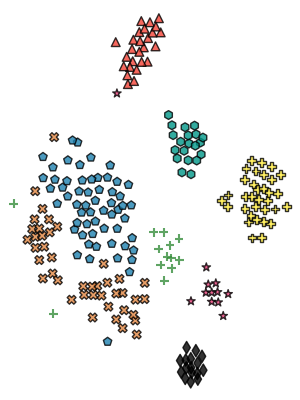}
        \caption{}
    \end{subfigure}
    \hfill
    \begin{subfigure}[b]{0.22\textwidth}
        \includegraphics[width=\textwidth]{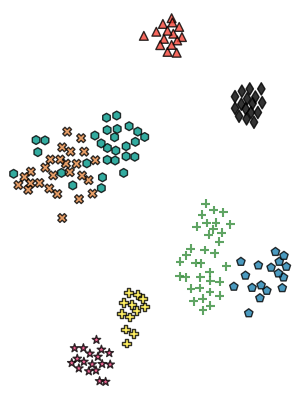}
        \caption{}
    \end{subfigure}
    \hfill
    \begin{subfigure}[b]{0.22\textwidth}
        \includegraphics[width=\textwidth]{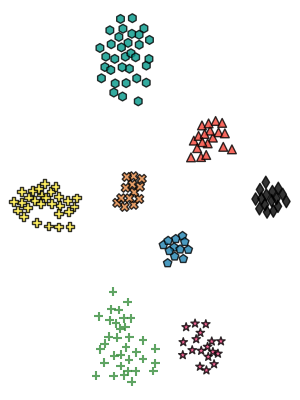}
        \caption{}
    \end{subfigure}

    \begin{subfigure}[b]{0.102\textwidth}
        \includegraphics[width=\textwidth]{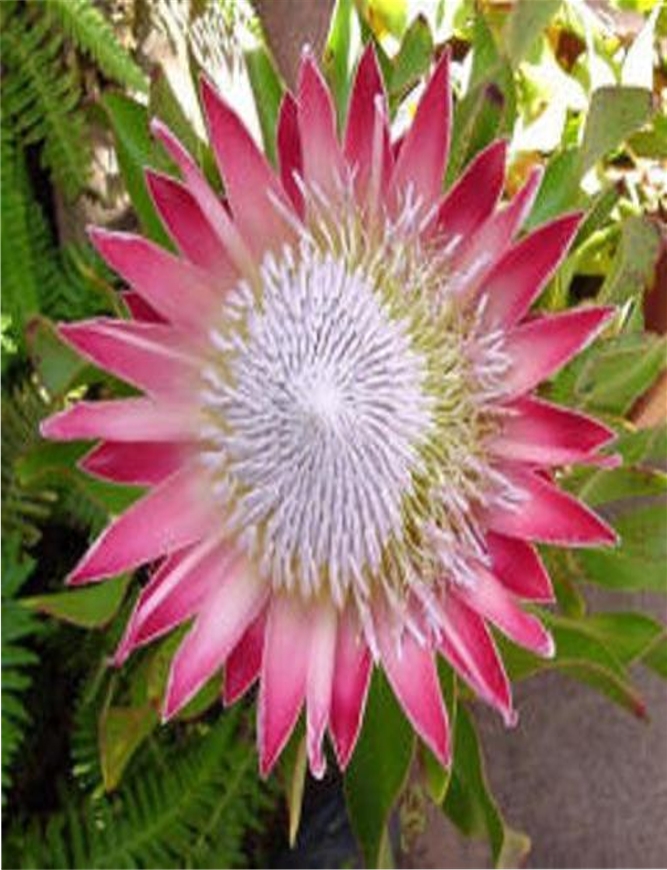}
    \end{subfigure} \hfill
    \begin{subfigure}[b]{0.102\textwidth}
        \includegraphics[width=\textwidth]{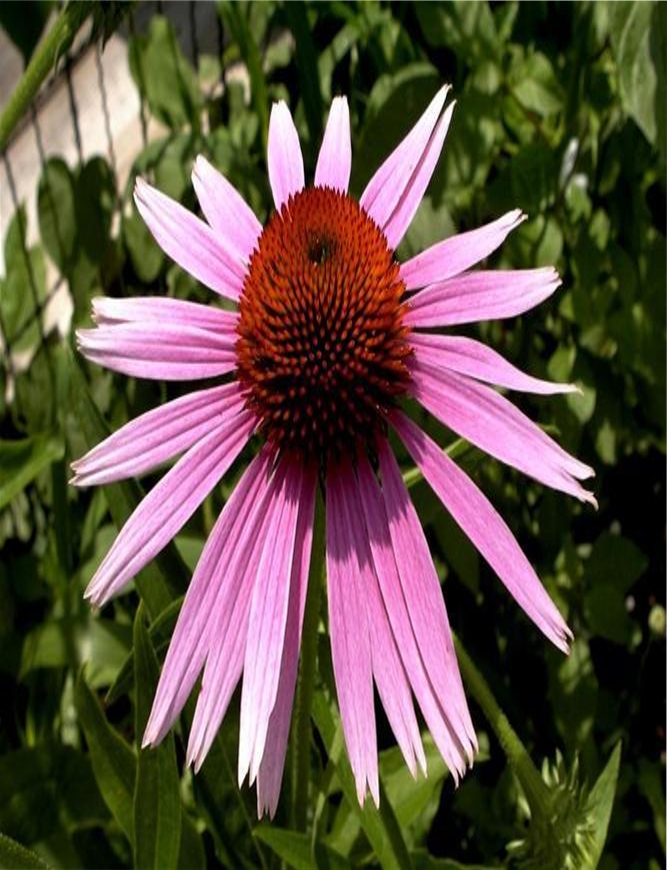}
    \end{subfigure} \hfill
    \begin{subfigure}[b]{0.099\textwidth}
        \includegraphics[width=\textwidth]{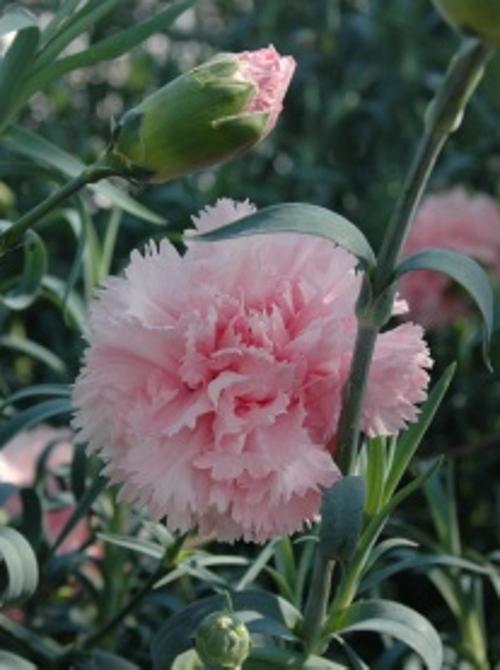}
    \end{subfigure} \hfill
    \begin{subfigure}[b]{0.102\textwidth}
        \includegraphics[width=\textwidth]{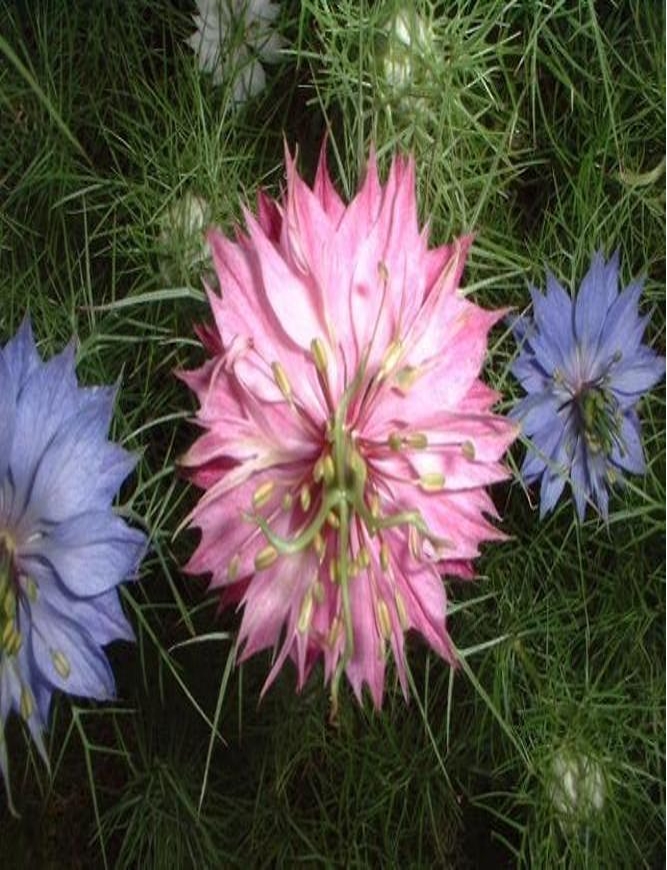}
    \end{subfigure} \hfill
    \begin{subfigure}[b]{0.099\textwidth}
        \includegraphics[width=\textwidth]{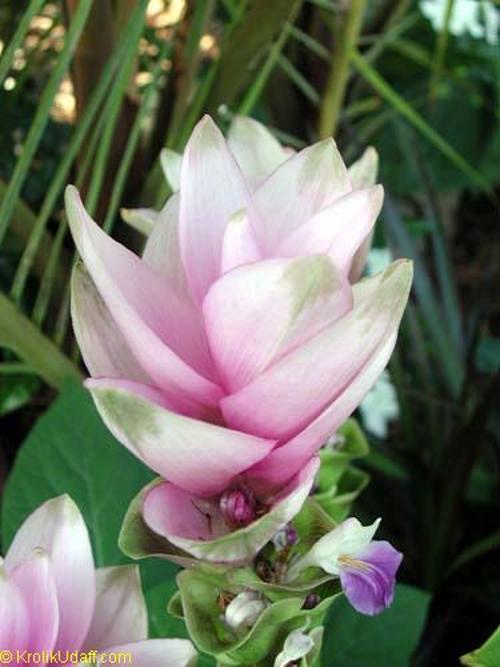}
    \end{subfigure} \hfill
    \begin{subfigure}[b]{0.102\textwidth}
        \includegraphics[width=\textwidth]{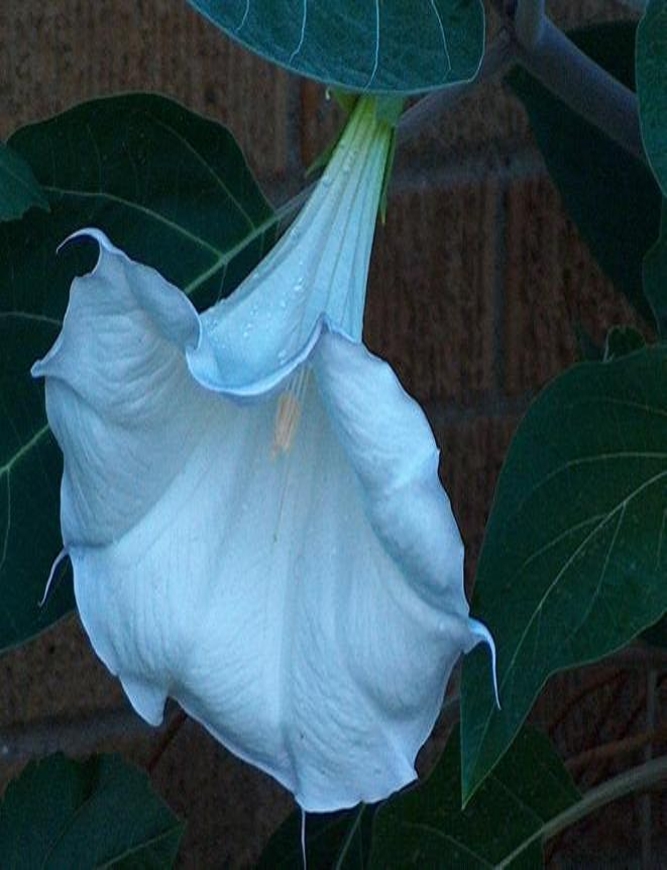}
    \end{subfigure} \hfill
    \begin{subfigure}[b]{0.102\textwidth}
        \includegraphics[width=\textwidth]{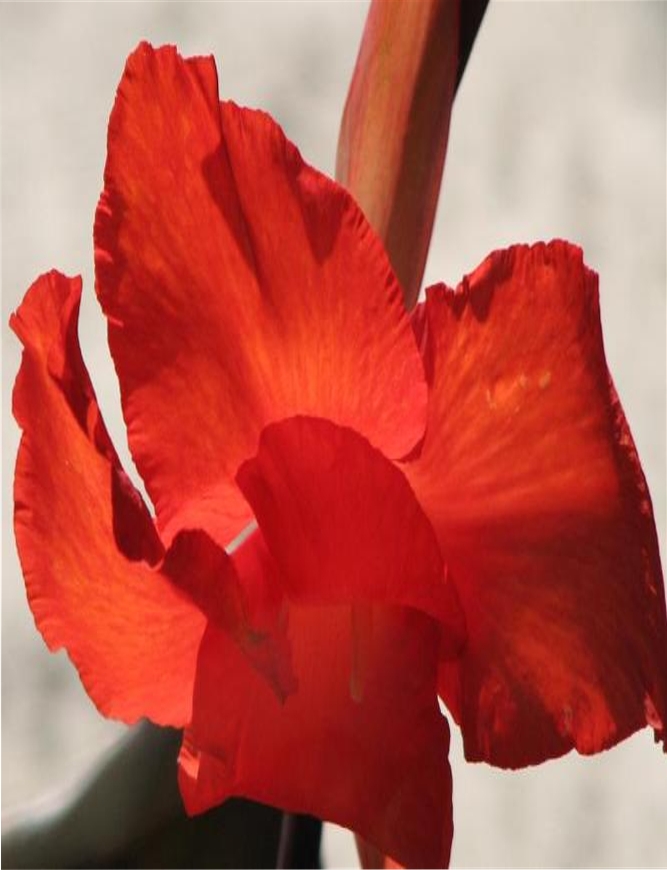}
    \end{subfigure} \hfill
    \begin{subfigure}[b]{0.102\textwidth}
        \includegraphics[width=\textwidth]{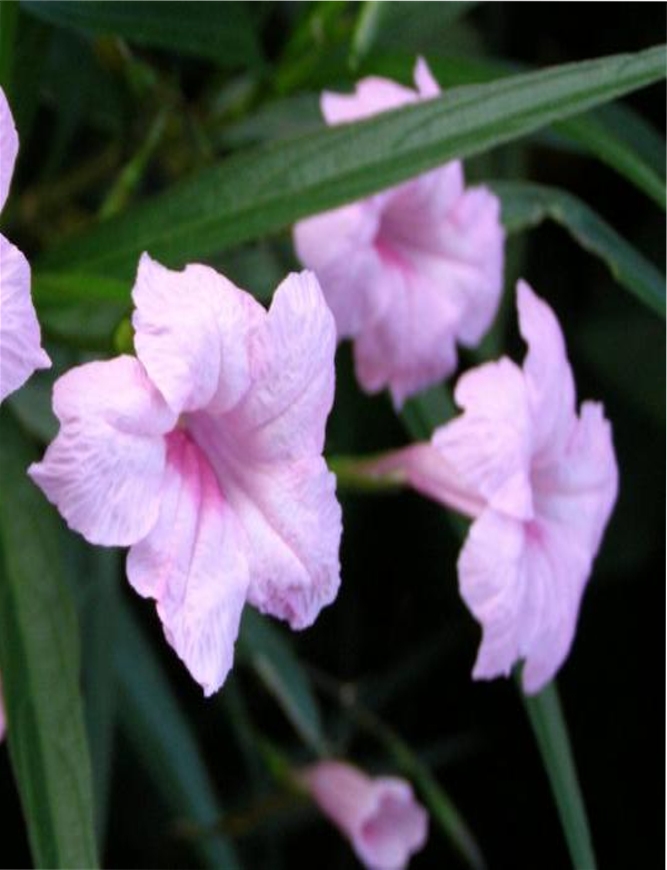}
    \end{subfigure} \hfill
    \caption{t-SNE visualizations of logit (see Eqn~(5)) features from 8 randomly selected Flowers102 classes using (a) Tip-Adapter, (b) Tip-Adapter-F, (c) ours with \texttt{Attention 1}, and (d) ours with combined attention (\texttt{Attention 1}$\cdot$\texttt{Attention 2}). The combined attention variant reduces overlap between highly similar flower categories by emphasizing discriminative part-level cues (e.g., petal arrangement and color gradients). The bottom row shows representative images from the selected classes.
} 
    \label{fig:t-sne-flowers}
\end{figure*}

\begin{figure*}[ht]
    \centering

    \begin{subfigure}[b]{0.22\textwidth}
        \includegraphics[width=\textwidth]{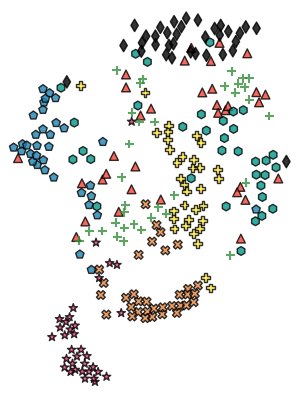}
        \caption{}
    \end{subfigure}
    \hfill
    \begin{subfigure}[b]{0.22\textwidth}
        \includegraphics[width=\textwidth]{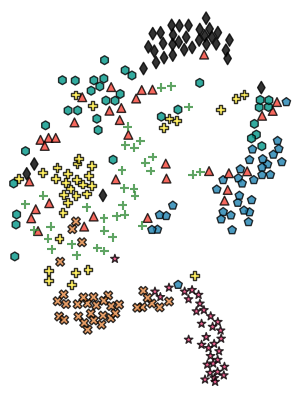}
        \caption{}
    \end{subfigure}
    \hfill
    \begin{subfigure}[b]{0.22\textwidth}
        \includegraphics[width=\textwidth]{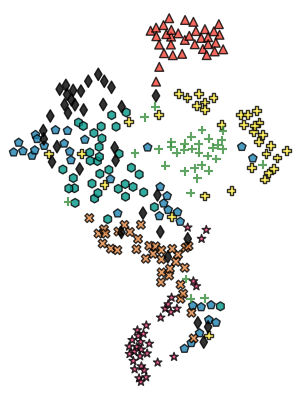}
        \caption{}
    \end{subfigure}
    \hfill
    \begin{subfigure}[b]{0.22\textwidth}
        \includegraphics[width=\textwidth]{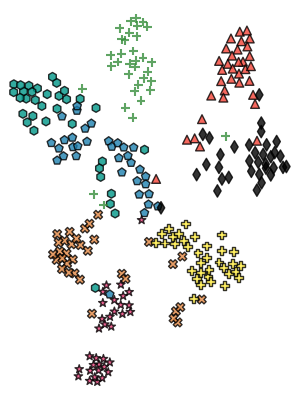}
        \caption{}
    \end{subfigure}

    \begin{subfigure}[b]{0.11\textwidth}
        \includegraphics[width=\textwidth]{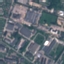}
    \end{subfigure} \hfill
    \begin{subfigure}[b]{0.11\textwidth}
        \includegraphics[width=\textwidth]{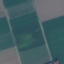}
    \end{subfigure} \hfill
    \begin{subfigure}[b]{0.11\textwidth}
        \includegraphics[width=\textwidth]{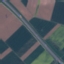}
    \end{subfigure} \hfill
    \begin{subfigure}[b]{0.11\textwidth}
        \includegraphics[width=\textwidth]{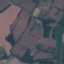}
    \end{subfigure} \hfill
    \begin{subfigure}[b]{0.11\textwidth}
        \includegraphics[width=\textwidth]{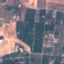}
    \end{subfigure} \hfill
    \begin{subfigure}[b]{0.11\textwidth}
        \includegraphics[width=\textwidth]{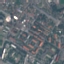}
    \end{subfigure} \hfill
    \begin{subfigure}[b]{0.11\textwidth}
        \includegraphics[width=\textwidth]{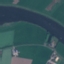}
    \end{subfigure} \hfill
    \begin{subfigure}[b]{0.11\textwidth}
        \includegraphics[width=\textwidth]{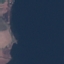}
    \end{subfigure} \hfill
    \caption{t-SNE visualizations of logit (see Eqn~(5)) features from 8 randomly selected EuroSAT classes using (a) Tip-Adapter, (b) Tip-Adapter-F, (c) ours with \texttt{Attention 1}, and (d) ours with combined attention (\texttt{Attention 1}$\cdot$\texttt{Attention 2}). With combined attention, our method yields clearer separation among land-cover categories by emphasizing consistent region-level cues (e.g., texture and spatial layout) and reducing overlap between visually similar classes. The bottom row shows representative satellite-image samples from the selected classes.
}
    \label{fig:t-sne-eurosat}
\end{figure*}

\begin{figure*}[ht]
    \centering

    \begin{subfigure}[b]{0.22\textwidth}
        \includegraphics[width=\textwidth]{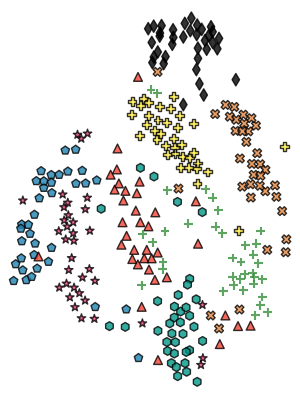}
        \caption{}
    \end{subfigure}
    \hfill
    \begin{subfigure}[b]{0.22\textwidth}
        \includegraphics[width=\textwidth]{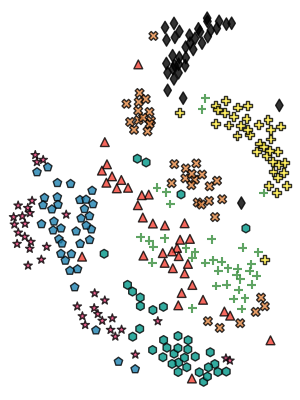}
        \caption{}
    \end{subfigure}
    \hfill
    \begin{subfigure}[b]{0.22\textwidth}
        \includegraphics[width=\textwidth]{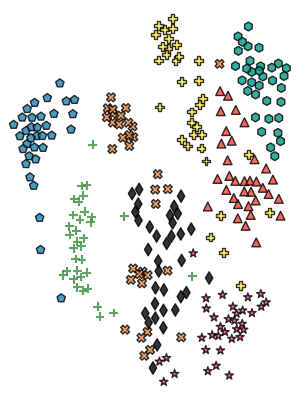}
        \caption{}
    \end{subfigure}
    \hfill
    \begin{subfigure}[b]{0.22\textwidth}
        \includegraphics[width=\textwidth]{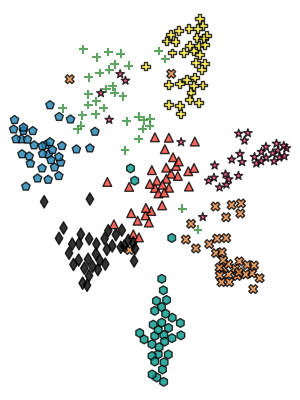}
        \caption{}
    \end{subfigure}

    \begin{subfigure}[b]{0.102\textwidth}
        \includegraphics[width=\textwidth]{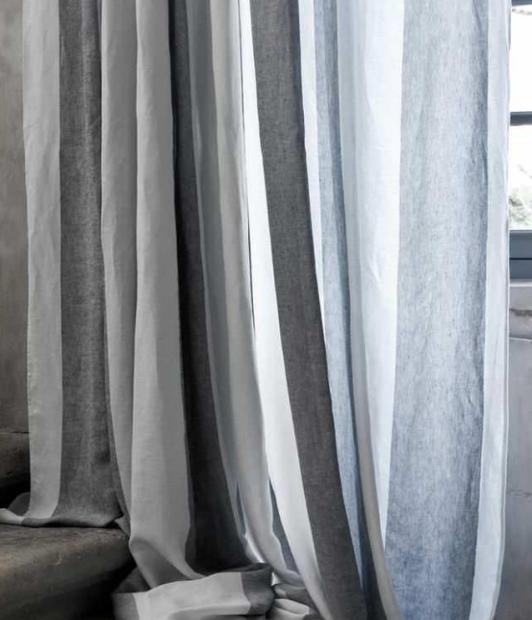}
    \end{subfigure} \hfill
    \begin{subfigure}[b]{0.102\textwidth}
        \includegraphics[width=\textwidth]{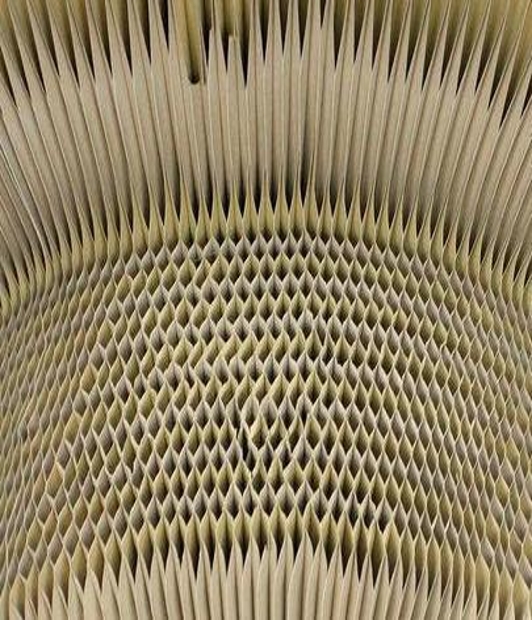}
    \end{subfigure} \hfill
    \begin{subfigure}[b]{0.102\textwidth}
        \includegraphics[width=\textwidth]{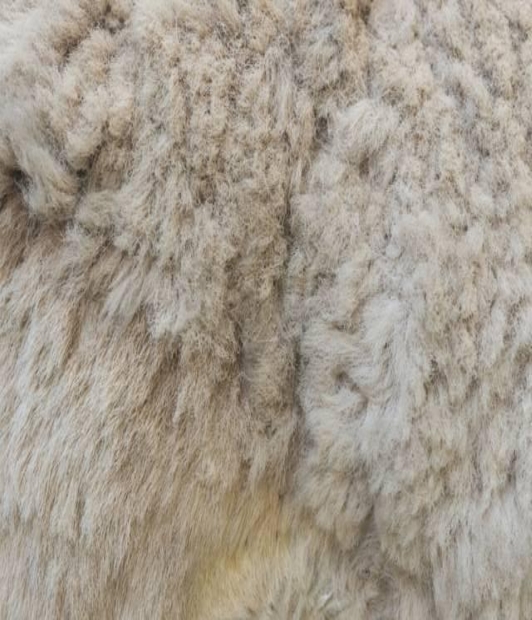}
    \end{subfigure} \hfill
    \begin{subfigure}[b]{0.102\textwidth}
        \includegraphics[width=\textwidth]{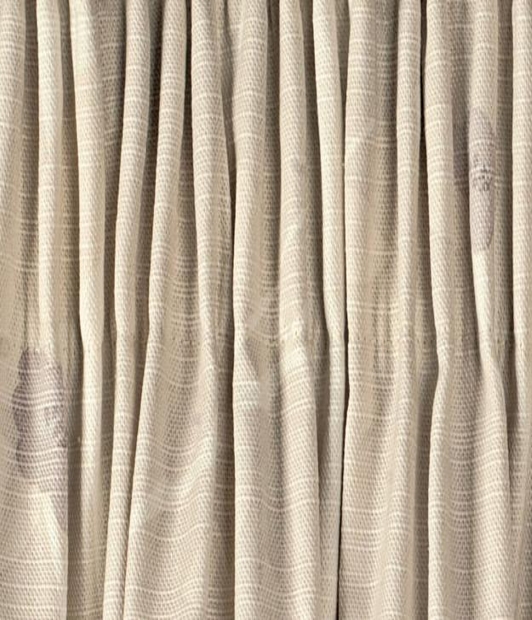}
    \end{subfigure} \hfill
    \begin{subfigure}[b]{0.102\textwidth}
        \includegraphics[width=\textwidth]{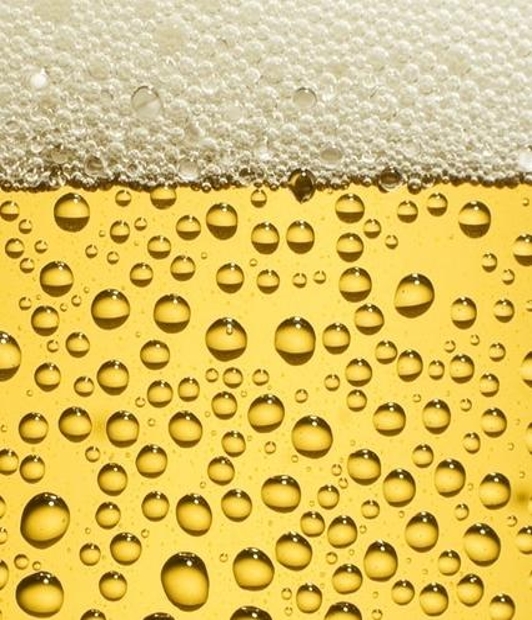}
    \end{subfigure} \hfill
    \begin{subfigure}[b]{0.102\textwidth}
        \includegraphics[width=\textwidth]{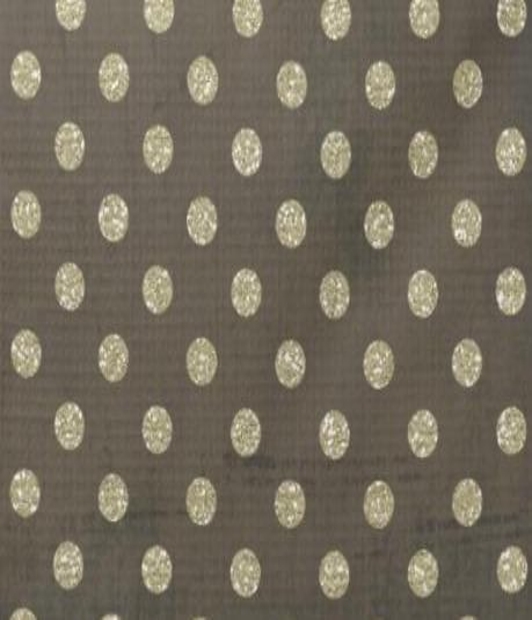}
    \end{subfigure} \hfill
    \begin{subfigure}[b]{0.102\textwidth}
        \includegraphics[width=\textwidth]{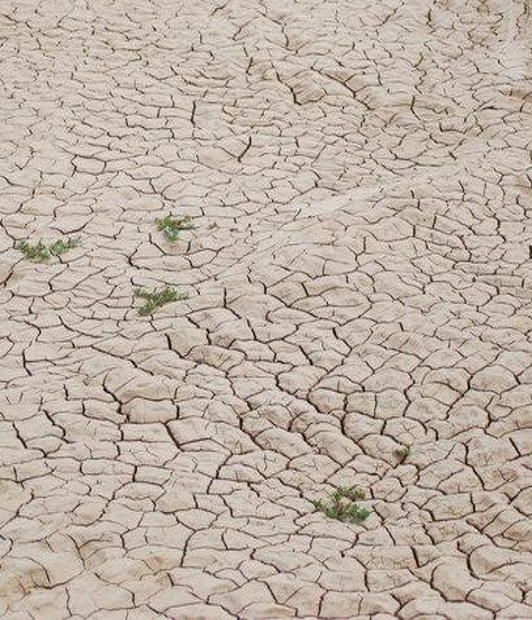}
    \end{subfigure} \hfill
    \begin{subfigure}[b]{0.11\textwidth}
        \includegraphics[width=\textwidth]{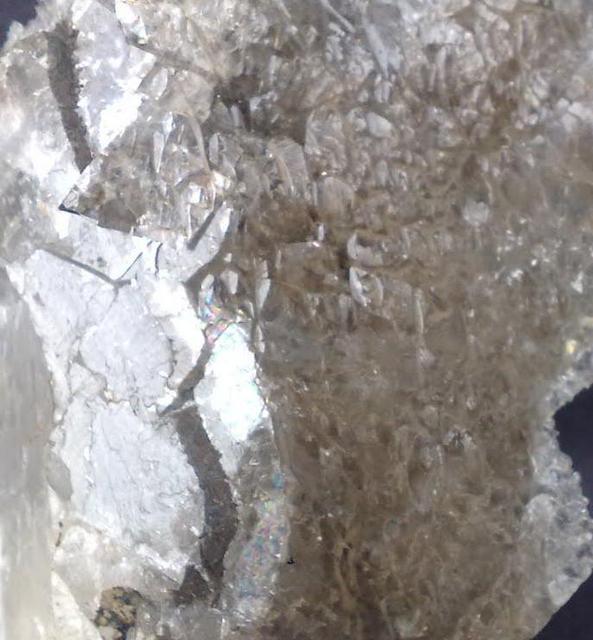}
    \end{subfigure} \hfill
    \caption{t-SNE visualizations of logit (see Eqn~(5)) features from 8 randomly selected DTD classes using (a) Tip-Adapter, (b) Tip-Adapter-F, (c) ours with \texttt{Attention 1}, and (d) ours with combined attention (\texttt{Attention 1}$\cdot$\texttt{Attention 2}). The combined attention produces more compact and distinctly separated clusters by emphasizing discriminative texture primitives and suppressing noisy patch interactions, which is crucial for separating visually similar material patterns in DTD. The bottom row shows representative texture samples from the selected classes.
}
    \label{fig:t-sne-dtd}
\end{figure*}

\begin{figure*}[ht]
    \centering

    \begin{subfigure}[b]{0.24\textwidth}
        \includegraphics[width=\textwidth]{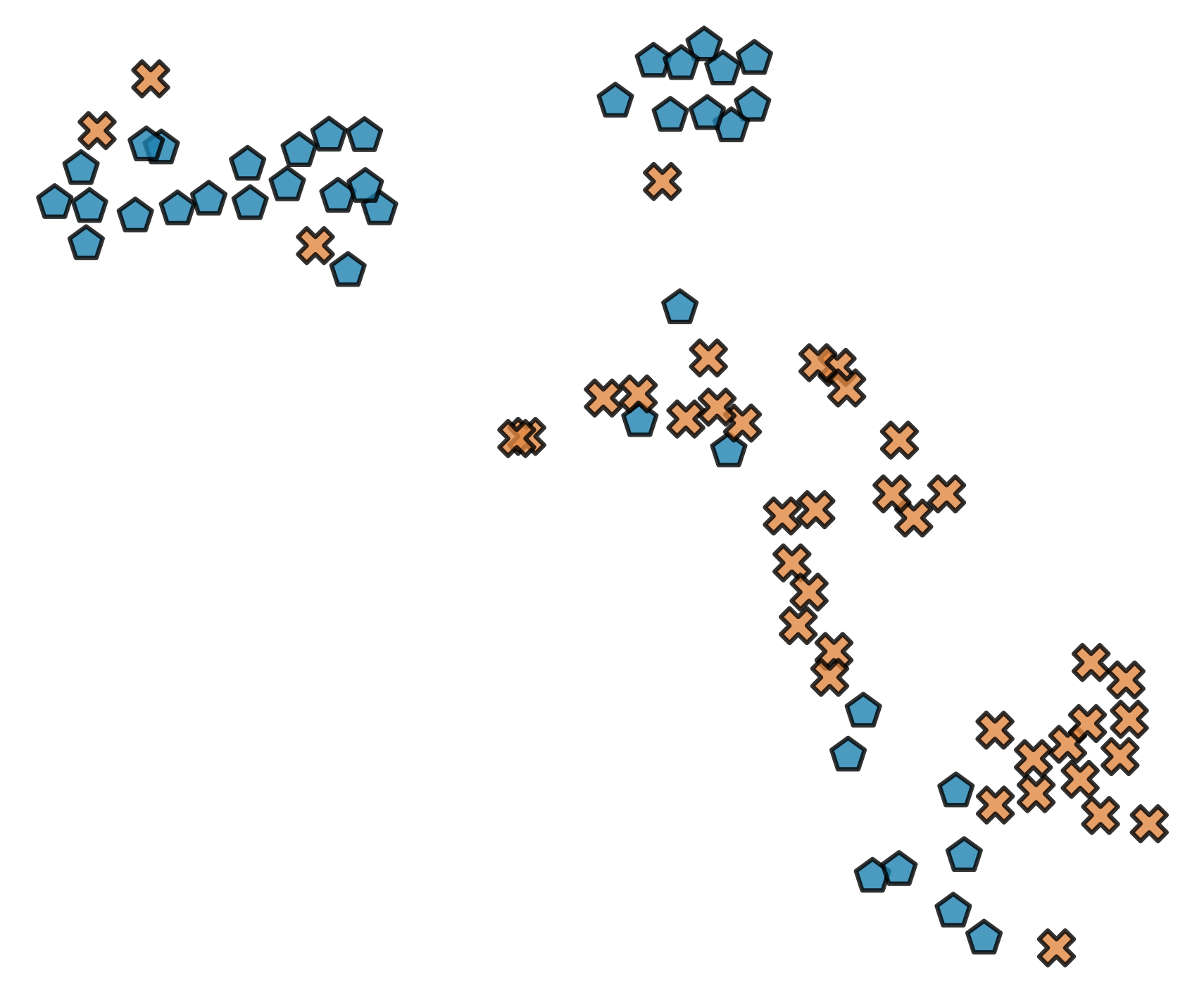}
        \caption{}
    \end{subfigure}
    \hfill
    \begin{subfigure}[b]{0.24\textwidth}
        \includegraphics[width=\textwidth]{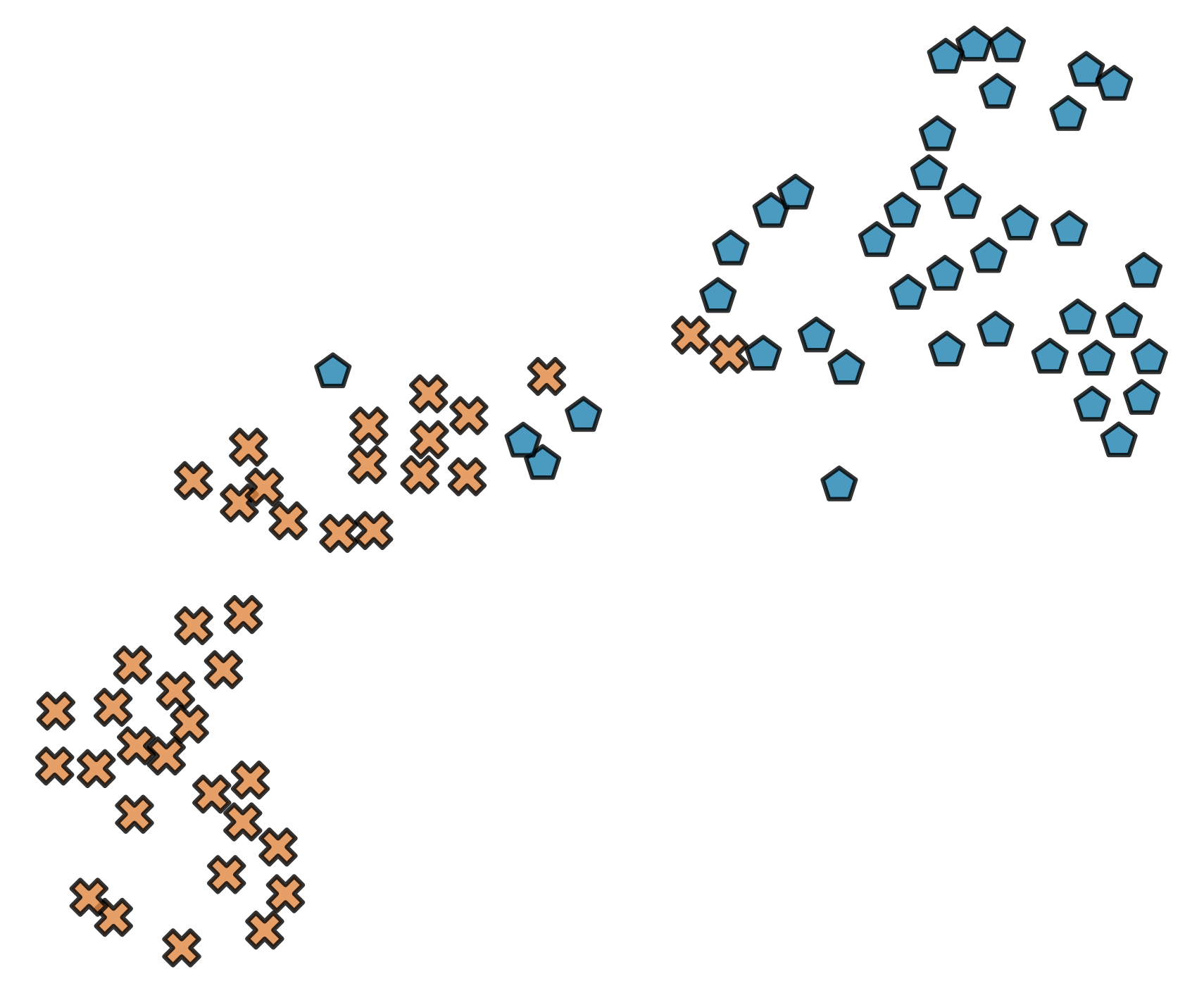}
        \caption{}
    \end{subfigure}
    \hfill
    \begin{subfigure}[b]{0.24\textwidth}
        \includegraphics[width=\textwidth]{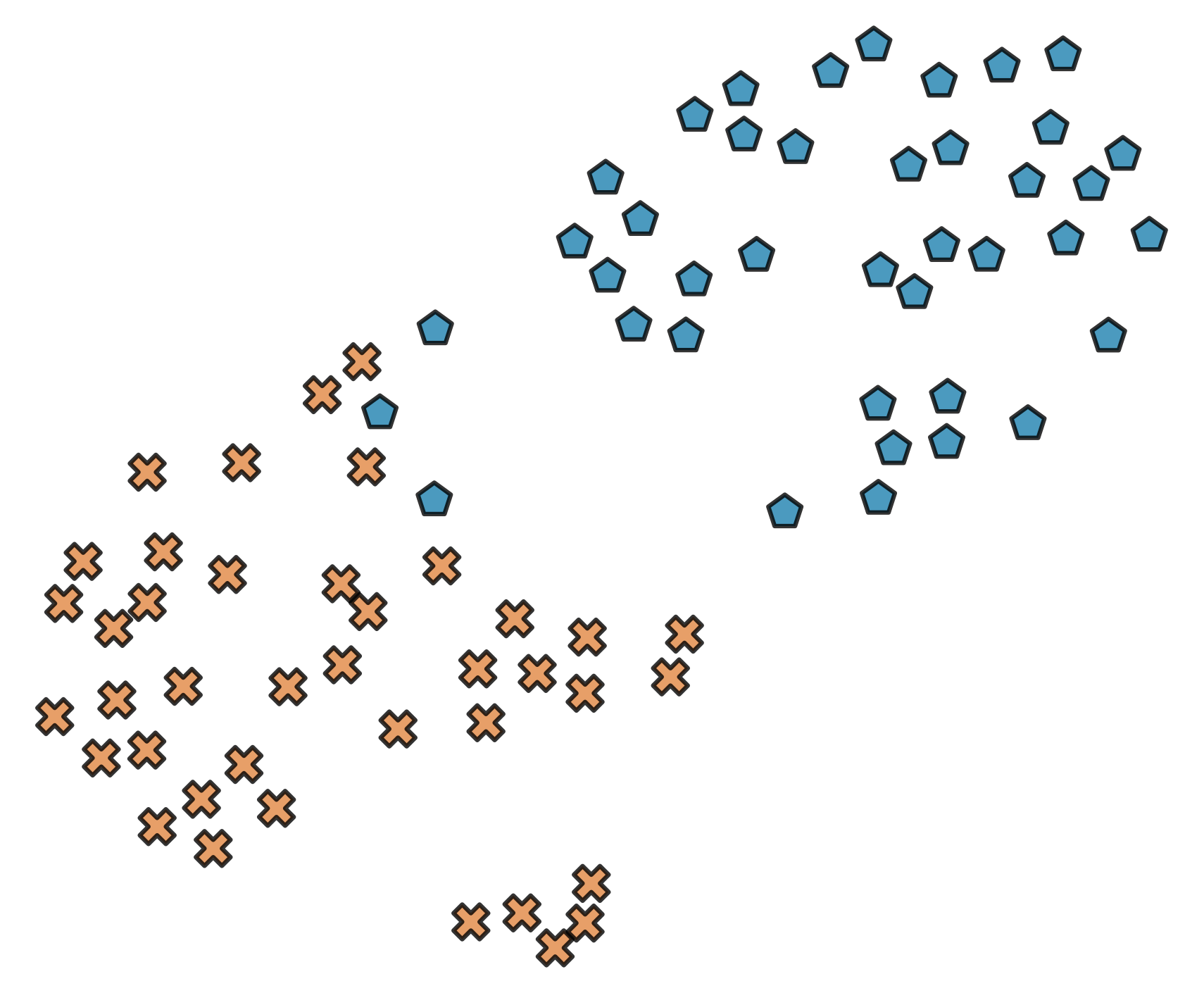}
        \caption{}
    \end{subfigure}
    \hfill
    \begin{subfigure}[b]{0.24\textwidth}
        \includegraphics[width=\textwidth]{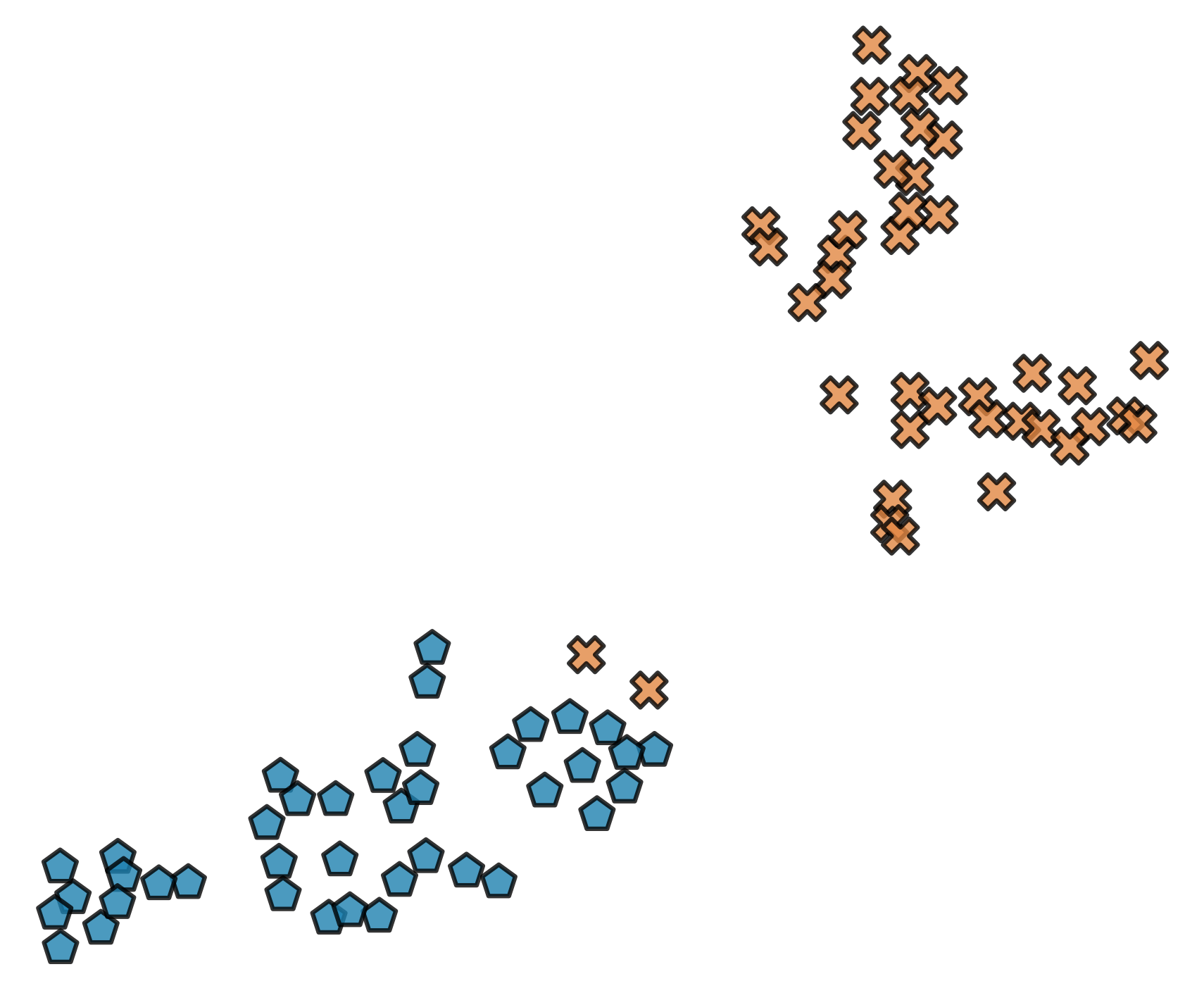}
        \caption{}
    \end{subfigure}

    \begin{subfigure}[b]{0.11\textwidth}
        \includegraphics[width=\textwidth]{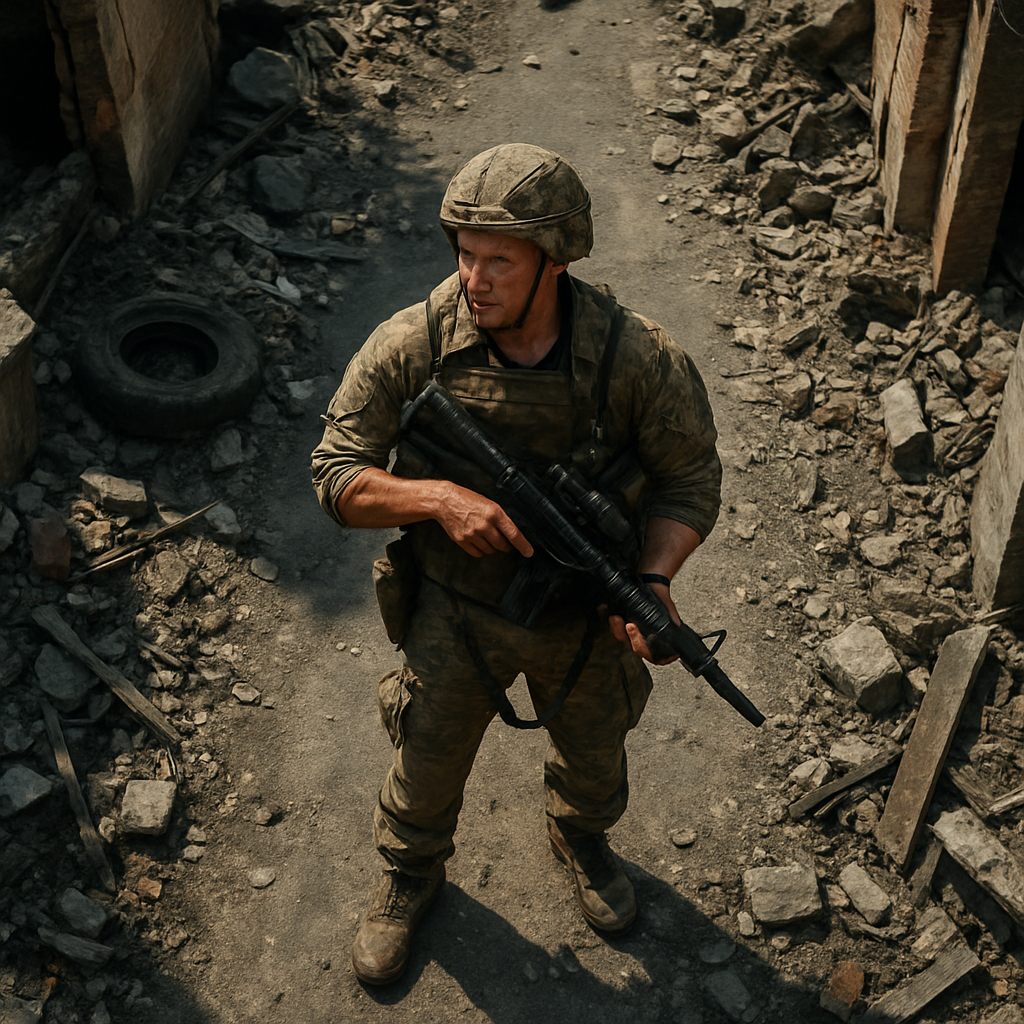}
    \end{subfigure} \hfill
    \begin{subfigure}[b]{0.11\textwidth}
        \includegraphics[width=\textwidth]{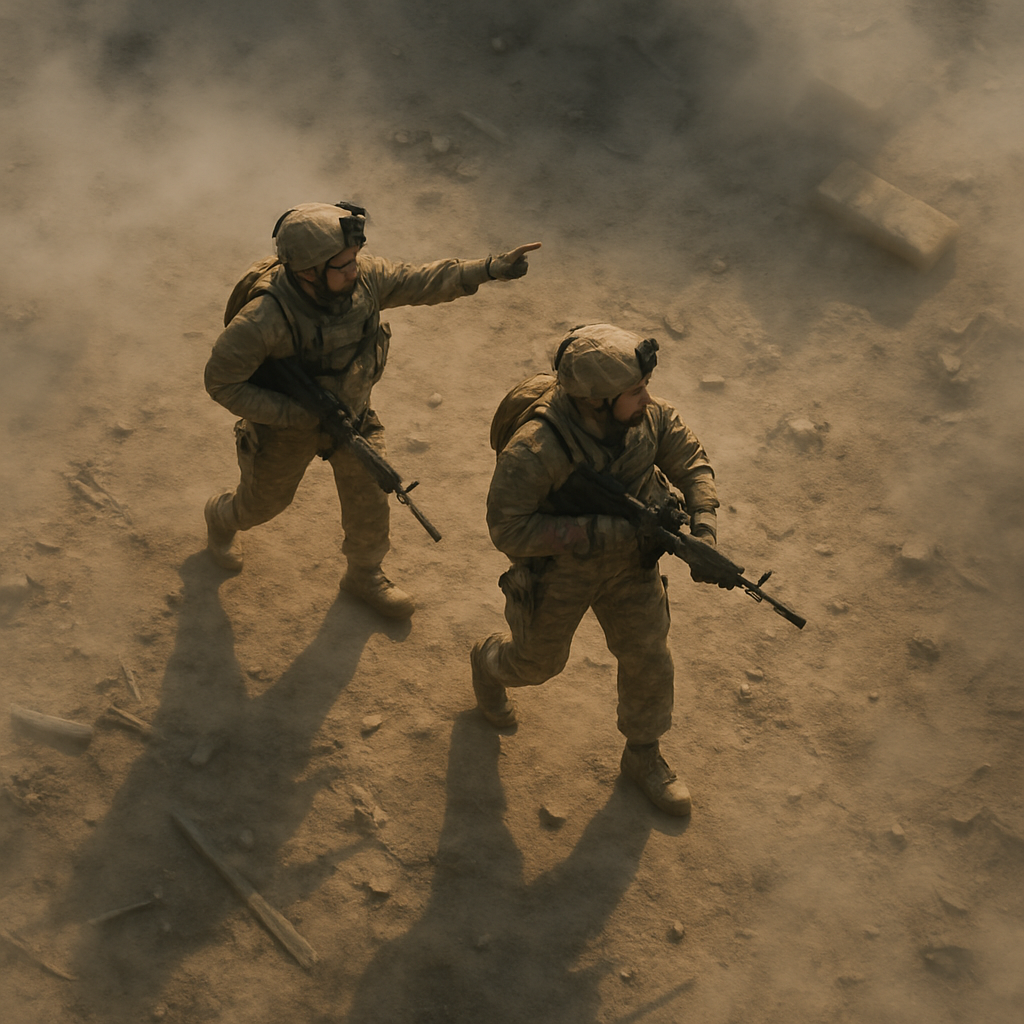}
    \end{subfigure} \hfill
    \begin{subfigure}[b]{0.11\textwidth}
        \includegraphics[width=\textwidth]{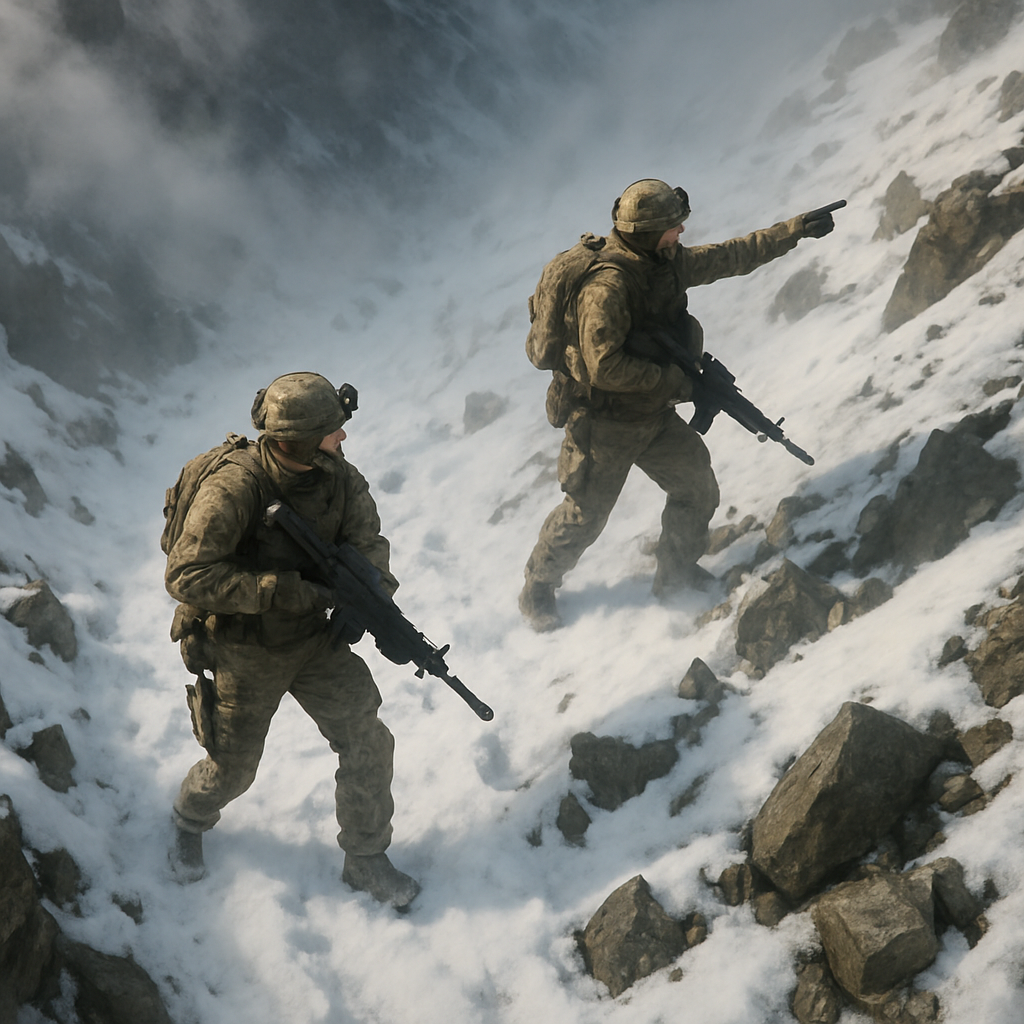}
    \end{subfigure} \hfill
    \begin{subfigure}[b]{0.11\textwidth}
        \includegraphics[width=\textwidth]{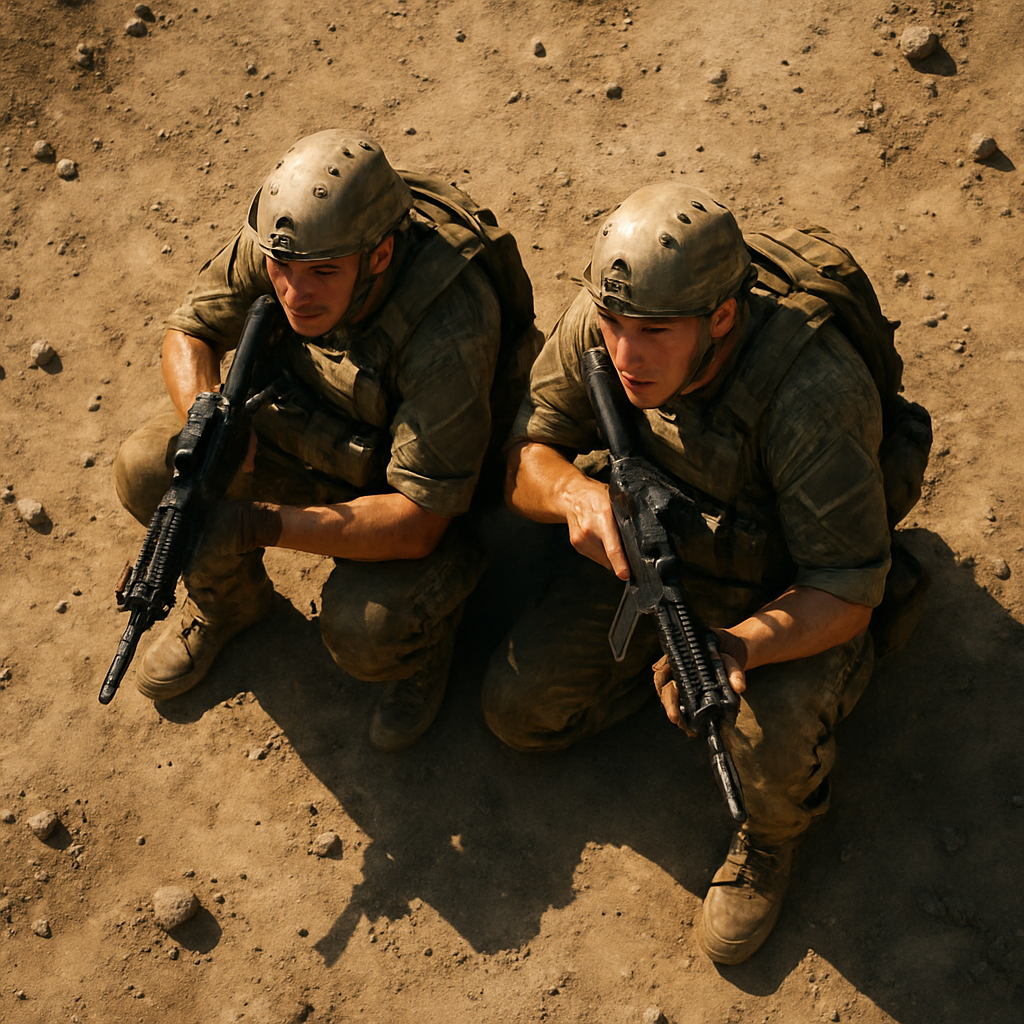}
    \end{subfigure} \hfill
    \begin{subfigure}[b]{0.11\textwidth}
        \includegraphics[width=\textwidth]{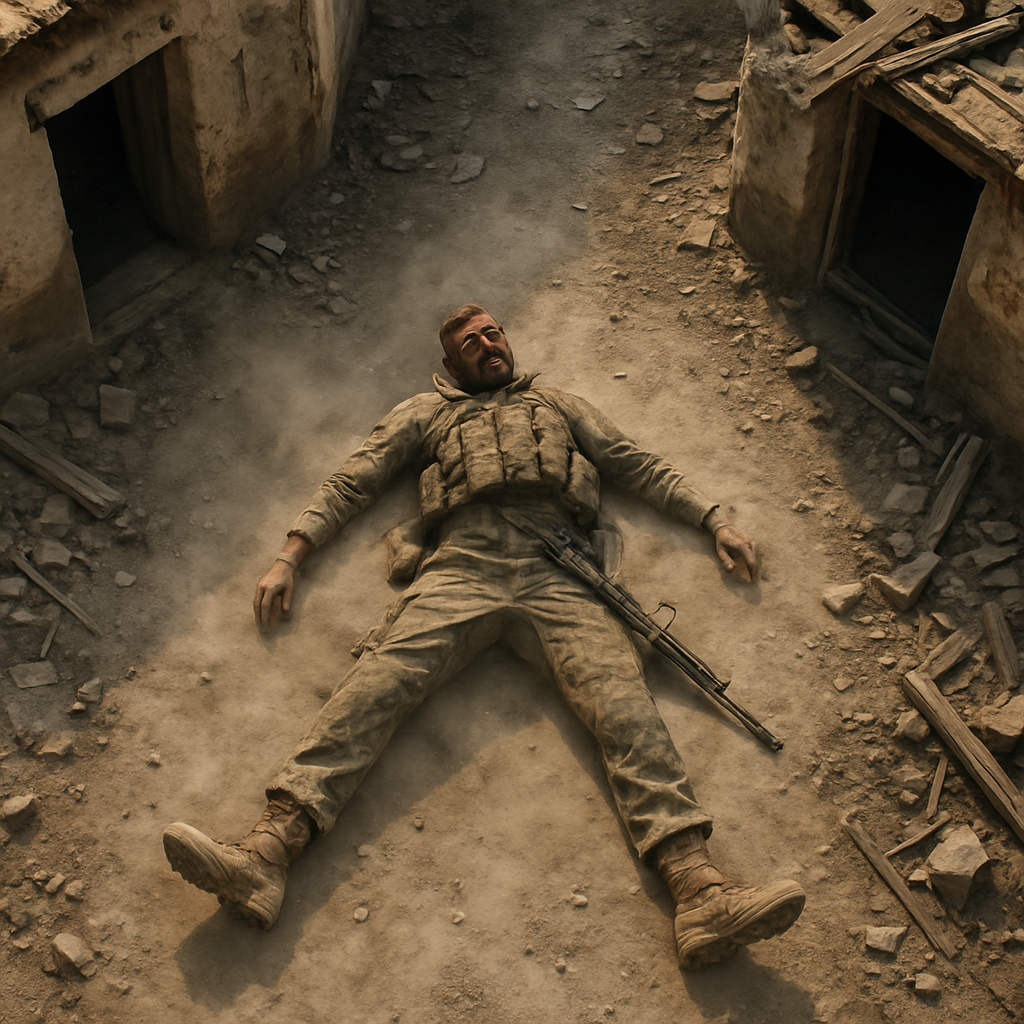}
    \end{subfigure} \hfill
    \begin{subfigure}[b]{0.11\textwidth}
        \includegraphics[width=\textwidth]{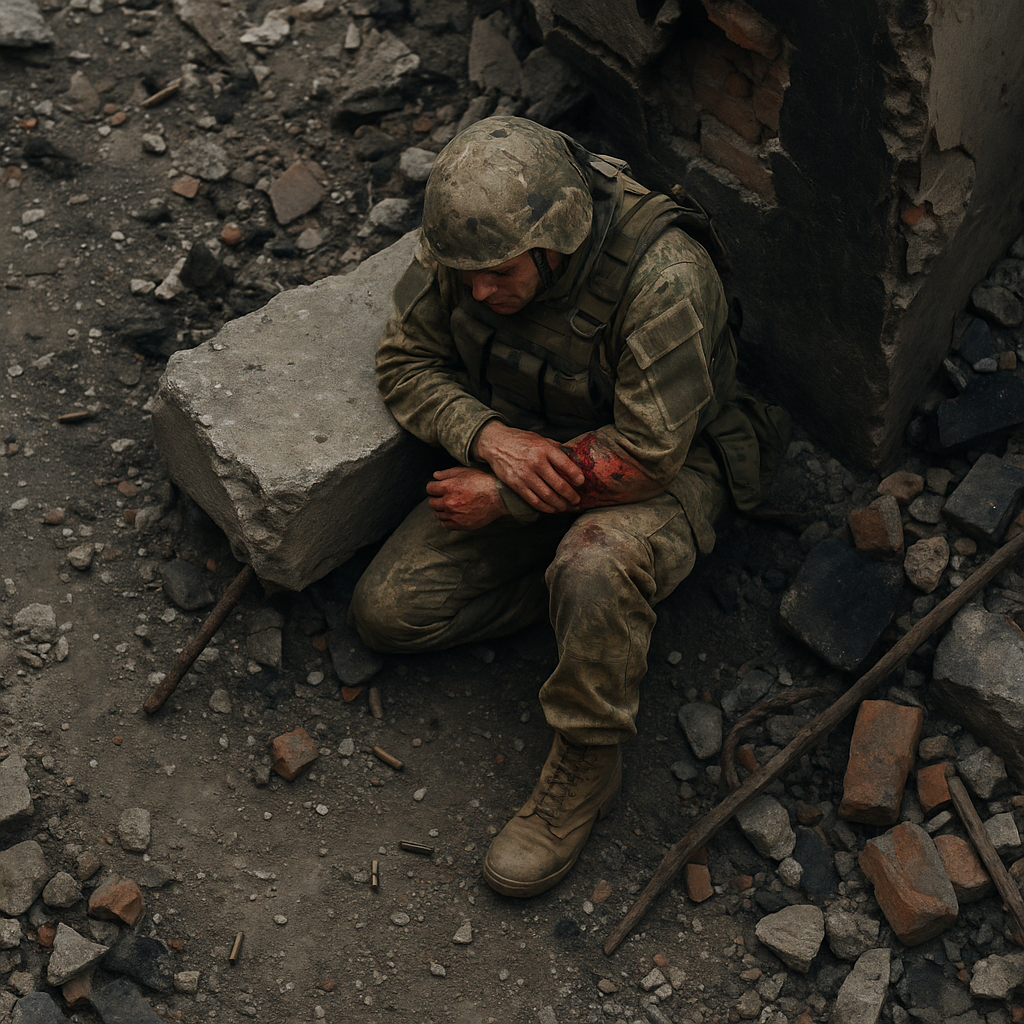}
    \end{subfigure} \hfill
    \begin{subfigure}[b]{0.11\textwidth}
        \includegraphics[width=\textwidth]{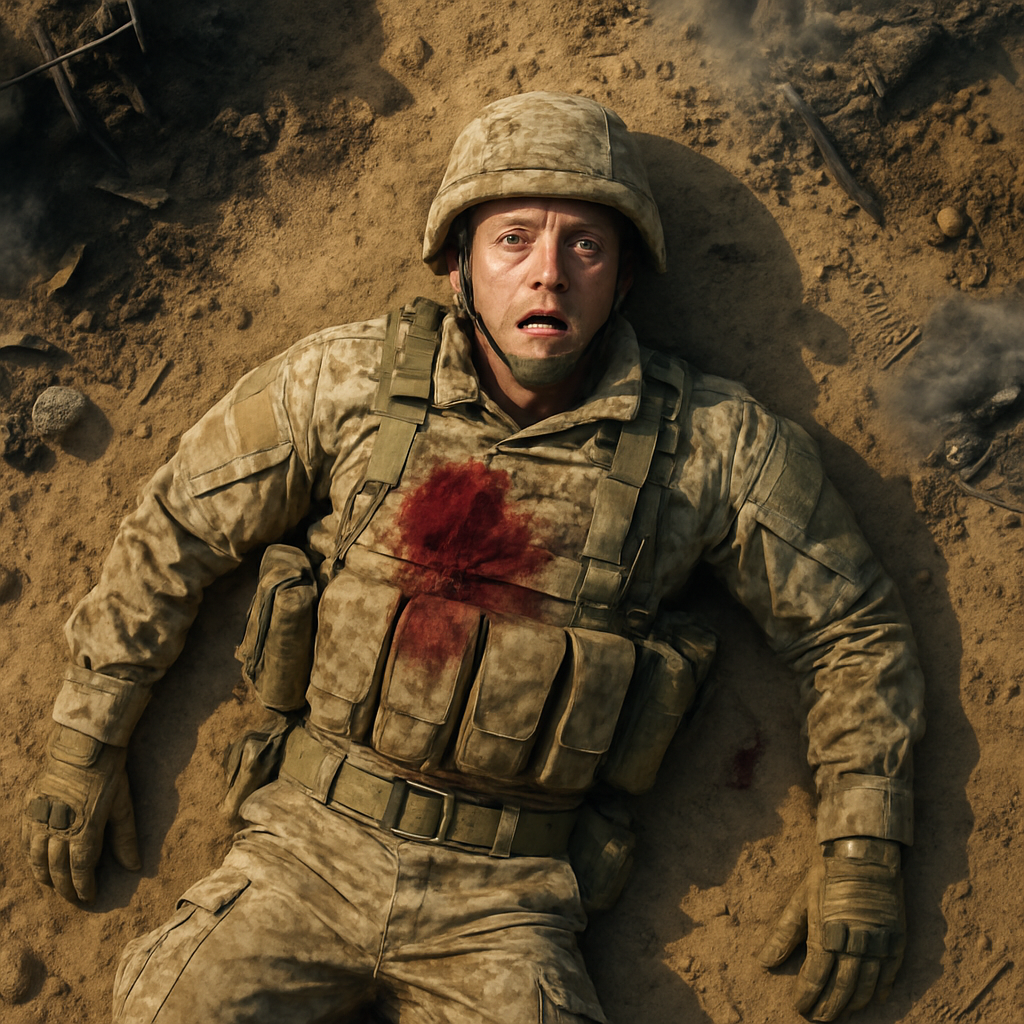}
    \end{subfigure} \hfill
    \begin{subfigure}[b]{0.11\textwidth}
        \includegraphics[width=\textwidth]{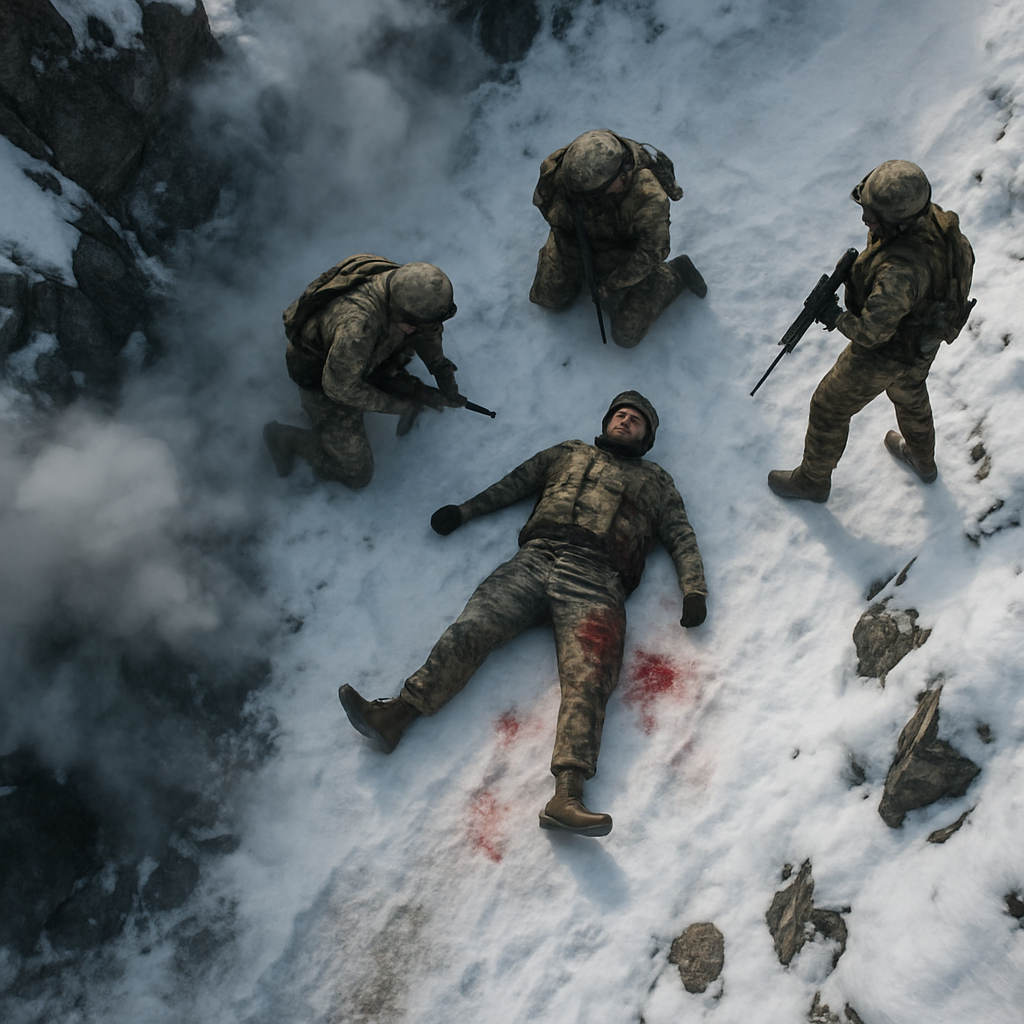}
    \end{subfigure} \hfill
    \caption{t-SNE visualizations of logit (see Eqn~(5)) features from selected samples of the \emph{Injured vs.~Uninjured Soldier} dataset using (a) Tip-Adapter, (b) Tip-Adapter-F, (c) ours with \texttt{Attention 1}, and (d) ours with combined attention (\texttt{Attention 1}$\cdot$\texttt{Attention 2}). Our method produces more compact, well-separated clusters especially with combined attention, improving injured vs.~uninjured discrimination under cluttered battlefield conditions. The bottom row shows representative samples: the first four are uninjured and the last four are injured.
    }
    \label{fig:t-sne-ours}
\end{figure*}

\paragraph{Additional Qualitative Analysis of Attention Mechanism:} 
Figs.~\ref{fig:t-sne-flowers}--\ref{fig:t-sne-ours} provide additional t-SNE visualizations of logit-space features (see Eqn.~(5) in the main paper) on four diverse datasets: Flowers102, EuroSAT, DTD, and our \emph{Injured vs.~Uninjured Soldier} dataset to qualitatively assess how the proposed patch-level relational attention reshapes the embedding space. We compare (a) Tip-Adapter, (b) Tip-Adapter-F, (c) our model with \texttt{Attention 1}, and (d) our full model with combined attention (\texttt{Attention 1}$\cdot$\texttt{Attention 2}). Across datasets, Tip-Adapter and Tip-Adapter-F form looser clusters with noticeable overlaps, suggesting that global cache adaptation alone can leave ambiguities unresolved when key evidence is localized. Adding \texttt{Attention 1} produces more structured embeddings by weighting patch interactions based on contextual compatibility, while the combined attention (\texttt{Attention 1}$\cdot$\texttt{Attention 2}) further improves separation by jointly enforcing contextual consistency and appearance similarity during message passing, yielding the most compact and well-separated clusters.

The effect is especially evident on fine-grained Flowers102 (Fig.~\ref{fig:t-sne-flowers}), where categories differ by subtle part-level variations (e.g., petal arrangement and color gradients); combined attention reduces overlap among visually similar flower species by emphasizing discriminative patch relations. On EuroSAT (Fig.~\ref{fig:t-sne-eurosat}), our method yields clearer partitioning of land-cover classes by focusing on consistent region-level patterns such as texture and spatial layout, which are naturally captured through patch-level relational aggregation. For DTD (Fig.~\ref{fig:t-sne-dtd}), the combined attention produces tighter clusters by highlighting discriminative texture primitives and suppressing noisy or redundant patch interactions, improving separability among visually similar material patterns. Finally, on the Injured vs.~Uninjured Soldier dataset (Fig.~\ref{fig:t-sne-ours}), combined attention yields a cleaner separation between the two labels under cluttered battlefield conditions, suggesting that relational attention helps aggregate localized injury evidence (e.g., blood patterns and posture cues) into more coherent cache representations. Overall, these qualitative results align with our quantitative findings, reinforcing that the proposed dual-attention relational modeling improves semantic organization and class separability across both fine-grained and domain-shifted recognition settings.

\bibliographystyle{IEEEtran}
\FloatBarrier
\bibliography{ref.bib}

@article{corso2020principal,
  title={Principal neighbourhood aggregation for graph nets},
  author={Corso, Gabriele and Cavalleri, Luca and Beaini, Dominique and Li{\`o}, Pietro and Veli{\v{c}}kovi{\'c}, Petar},
  journal={Advances in neural information processing systems},
  volume={33},
  pages={13260--13271},
  year={2020}
}

@inproceedings{jiang2025causal,
  title={Causal Disentanglement and Cross-Modal Alignment for Enhanced Few-Shot Learning},
  author={Jiang, Tianjiao and Zhang, Zhen and Liu, Yuhang and Shi, Javen Qinfeng},
  booktitle={Proceedings of the IEEE/CVF International Conference on Computer Vision},
  pages={890--900},
  year={2025}
}

@inproceedings{jia2021scaling,
  title={Scaling up visual and vision-language representation learning with noisy text supervision},
  author={Jia, Chao and Yang, Yinfei and Xia, Ye and Chen, Yi-Ting and Parekh, Zarana and Pham, Hieu and Le, Quoc and Sung, Yun-Hsuan and Li, Zhen and Duerig, Tom},
  booktitle={International conference on machine learning},
  pages={4904--4916},
  year={2021},
  organization={PMLR}
}

@inproceedings{li2021supervision,
      title={Supervision Exists Everywhere: A Data Efficient Contrastive Language-Image  Pre-training Paradigm},
      author={Yangguang Li and Feng Liang and Lichen Zhao and Yufeng Cui and Wanli Ouyang and Jing Shao and Fengwei Yu and Junjie Yan},
      booktitle={International Conference on Learning Representations},
      year={2022}
}

@inproceedings{wang2022medclip,
  title={Medclip: Contrastive learning from unpaired medical images and text},
  author={Wang, Zifeng and Wu, Zhenbang and Agarwal, Dinesh and Sun, Jimeng},
  booktitle={Proceedings of the Conference on Empirical Methods in Natural Language Processing. Conference on Empirical Methods in Natural Language Processing},
  volume={2022},
  pages={3876},
  year={2022}
}

@inproceedings{zhang2022tip,
  title={Tip-adapter: Training-free adaption of clip for few-shot classification},
  author={Zhang, Renrui and Zhang, Wei and Fang, Rongyao and Gao, Peng and Li, Kunchang and Dai, Jifeng and Qiao, Yu and Li, Hongsheng},
  booktitle={European conference on computer vision},
  pages={493--510},
  year={2022},
}

@inproceedings{li2022grounded,
  title={Grounded language-image pre-training},
  author={Li, Liunian Harold and Zhang, Pengchuan and Zhang, Haotian and Yang, Jianwei and Li, Chunyuan and Zhong, Yiwu and Wang, Lijuan and Yuan, Lu and Zhang, Lei and Hwang, Jenq-Neng and others},
  booktitle={Proceedings of the IEEE/CVF conference on computer vision and pattern recognition},
  pages={10965--10975},
  year={2022}
}

@inproceedings{yao2022filip,
  title={FILIP: Fine-grained Interactive Language-Image Pre-Training},
  author={Yao, Lewei and Huang, Runhui and Hou, Lu and Lu, Guansong and Niu, Minzhe and Xu, Hang and Liang, Xiaodan and Li, Zhenguo and Jiang, Xin and Xu, Chunjing},
  booktitle={International Conference on Learning Representations},
    year={2022}
}

@article{zhang2024vision,
  title={Vision-language models for vision tasks: A survey},
  author={Zhang, Jingyi and Huang, Jiaxing and Jin, Sheng and Lu, Shijian},
  journal={IEEE Transactions on Pattern Analysis and Machine Intelligence},
  year={2024},
  publisher={IEEE}
}

@article{li2024graphadapter,
  title={Graphadapter: Tuning vision-language models with dual knowledge graph},
  author={Li, Xin and Lian, Dongze and Lu, Zhihe and Bai, Jiawang and Chen, Zhibo and Wang, Xinchao},
  journal={Advances in Neural Information Processing Systems},
  volume={36},
  year={2024}
}

@article{boudiaf2020information,
  title={Information maximization for few-shot learning},
  author={Boudiaf, Malik and Ziko, Imtiaz and Rony, J{\'e}r{\^o}me and Dolz, Jos{\'e} and Piantanida, Pablo and Ben Ayed, Ismail},
  journal={Advances in Neural Information Processing Systems},
  volume={33},
  pages={2445--2457},
  year={2020}
}

@inproceedings{guneet2020baseline,
  title={A baseline for few-shot image classification},
  author={Guneet, S Dhillon and Pratik, Chaudhari and Avinash, Ravichandran and Stefano, S},
  booktitle={International Conference on Learning Representations (ICLR)},
  volume={10},
  year={2020}
}

@inproceedings{lazarou2021iterative,
  title={Iterative label cleaning for transductive and semi-supervised few-shot learning},
  author={Lazarou, Michalis and Stathaki, Tania and Avrithis, Yannis},
  booktitle={Proceedings of the ieee/cvf international conference on computer vision},
  pages={8751--8760},
  year={2021}
}

@inproceedings{cimpoi2014describing,
  title={Describing textures in the wild},
  author={Cimpoi, Mircea and Maji, Subhransu and Kokkinos, Iasonas and Mohamed, Sammy and Vedaldi, Andrea},
  booktitle={Proceedings of the IEEE conference on computer vision and pattern recognition},
  pages={3606--3613},
  year={2014}
}

@article{maji2013fine,
  title={Fine-grained visual classification of aircraft},
  author={Maji, Subhransu and Rahtu, Esa and Kannala, Juho and Blaschko, Matthew and Vedaldi, Andrea},
  journal={arXiv preprint arXiv:1306.5151},
  year={2013}
}

@inproceedings{radford2021learning,
  title={Learning transferable visual models from natural language supervision},
  author={Radford, Alec and Kim, Jong Wook and Hallacy, Chris and Ramesh, Aditya and Goh, Gabriel and Agarwal, Sandhini and Sastry, Girish and Askell, Amanda and Mishkin, Pamela and Clark, Jack and others},
  booktitle={International conference on machine learning},
  pages={8748--8763},
  year={2021},
  organization={PmLR}
}

@article{ramesh2022hierarchical,
  title={Hierarchical text-conditional image generation with clip latents},
  author={Ramesh, Aditya and Dhariwal, Prafulla and Nichol, Alex and Chu, Casey and Chen, Mark},
  journal={arXiv preprint arXiv:2204.06125},
  volume={1},
  number={2},
  pages={3},
  year={2022}
}

@inproceedings{lu2022prompt,
  title={Prompt distribution learning},
  author={Lu, Yuning and Liu, Jianzhuang and Zhang, Yonggang and Liu, Yajing and Tian, Xinmei},
  booktitle={Proceedings of the IEEE/CVF conference on computer vision and pattern recognition},
  pages={5206--5215},
  year={2022}
}

@inproceedings{patashnik2021styleclip,
  title={Styleclip: Text-driven manipulation of stylegan imagery},
  author={Patashnik, Or and Wu, Zongze and Shechtman, Eli and Cohen-Or, Daniel and Lischinski, Dani},
  booktitle={Proceedings of the IEEE/CVF international conference on computer vision},
  pages={2085--2094},
  year={2021}
}

@inproceedings{nichol2022glide,
  title={GLIDE: Towards Photorealistic Image Generation and Editing with Text-Guided Diffusion Models},
  author={Nichol, Alexander Quinn and Dhariwal, Prafulla and Ramesh, Aditya and Shyam, Pranav and Mishkin, Pamela and Mcgrew, Bob and Sutskever, Ilya and Chen, Mark},
  booktitle={International Conference on Machine Learning},
  pages={16784--16804},
  year={2022},
  organization={PMLR}
}

@article{zhou2022learning,
  title={Learning to prompt for vision-language models},
  author={Zhou, Kaiyang and Yang, Jingkang and Loy, Chen Change and Liu, Ziwei},
  journal={International Journal of Computer Vision},
  volume={130},
  number={9},
  pages={2337--2348},
  year={2022},
  publisher={Springer}
}

@article{gao2024clip,
  title={Clip-adapter: Better vision-language models with feature adapters},
  author={Gao, Peng and Geng, Shijie and Zhang, Renrui and Ma, Teli and Fang, Rongyao and Zhang, Yongfeng and Li, Hongsheng and Qiao, Yu},
  journal={International Journal of Computer Vision},
  volume={132},
  number={2},
  pages={581--595},
  year={2024},
  publisher={Springer}
}

@inproceedings{zhang2023prompt,
  title={Prompt, generate, then cache: Cascade of foundation models makes strong few-shot learners},
  author={Zhang, Renrui and Hu, Xiangfei and Li, Bohao and Huang, Siyuan and Deng, Hanqiu and Qiao, Yu and Gao, Peng and Li, Hongsheng},
  booktitle={Proceedings of the IEEE/CVF conference on computer vision and pattern recognition},
  pages={15211--15222},
  year={2023}
}

@article{zhang2022prompting,
  title={Prompting through prototype: A prototype-based prompt learning on pretrained vision-language models},
  author={Zhang, Yue and Fei, Hongliang and Li, Dingcheng and Yu, Tan and Li, Ping},
  journal={arXiv preprint arXiv:2210.10841},
  year={2022}
}

@inproceedings{zhou2022conditional,
  title={Conditional prompt learning for vision-language models},
  author={Zhou, Kaiyang and Yang, Jingkang and Loy, Chen Change and Liu, Ziwei},
  booktitle={Proceedings of the IEEE/CVF conference on computer vision and pattern recognition},
  pages={16816--16825},
  year={2022}
}

@inproceedings{chen2023plot,
  title={Prompt Learning with Optimal Transport for Vision-Language Models},
  author={Chen, Guangyi and Yao, Weiran and Song, Xiangchen and Li, Xinyue and Rao, Yongming and Zhang, Kun},
  booktitle={ICLR},
  year={2023}
}

@inproceedings{yao2023visual,
  title={Visual-language prompt tuning with knowledge-guided context optimization},
  author={Yao, Hantao and Zhang, Rui and Xu, Changsheng},
  booktitle={Proceedings of the IEEE/CVF conference on computer vision and pattern recognition},
  pages={6757--6767},
  year={2023}
}

@inproceedings{yu2023task,
  title={Task residual for tuning vision-language models},
  author={Yu, Tao and Lu, Zhihe and Jin, Xin and Chen, Zhibo and Wang, Xinchao},
  booktitle={Proceedings of the IEEE/CVF Conference on Computer Vision and Pattern Recognition},
  pages={10899--10909},
  year={2023}
}

@inproceedings{nilsback2008automated,
  title={Automated flower classification over a large number of classes},
  author={Nilsback, Maria-Elena and Zisserman, Andrew},
  booktitle={2008 Sixth Indian conference on computer vision, graphics \& image processing},
  pages={722--729},
  year={2008},
  organization={IEEE}
}

@inproceedings{fei2004learning,
  title={Learning generative visual models from few training examples: An incremental bayesian approach tested on 101 object categories},
  author={Fei-Fei, Li and Fergus, Rob and Perona, Pietro},
  booktitle={2004 conference on computer vision and pattern recognition workshop},
  pages={178--178},
  year={2004},
  organization={IEEE}
}

@inproceedings{xiao2010sun,
  title={Sun database: Large-scale scene recognition from abbey to zoo},
  author={Xiao, Jianxiong and Hays, James and Ehinger, Krista A and Oliva, Aude and Torralba, Antonio},
  booktitle={2010 IEEE computer society conference on computer vision and pattern recognition},
  pages={3485--3492},
  year={2010},
  organization={IEEE}
}

@article{soomro2012ucf101,
  title={UCF101: A dataset of 101 human actions classes from videos in the wild},
  author={Soomro, Khurram and Zamir, Amir Roshan and Shah, Mubarak},
  journal={arXiv preprint arXiv:1212.0402},
  year={2012}
}

@inproceedings{krause20133d,
  title={3d object representations for fine-grained categorization},
  author={Krause, Jonathan and Stark, Michael and Deng, Jia and Fei-Fei, Li},
  booktitle={Proceedings of the IEEE international conference on computer vision workshops},
  pages={554--561},
  year={2013}
}

@inproceedings{khattak2023maple,
  title={Maple: Multi-modal prompt learning},
  author={Khattak, Muhammad Uzair and Rasheed, Hanoona and Maaz, Muhammad and Khan, Salman and Khan, Fahad Shahbaz},
  booktitle={Proceedings of the IEEE/CVF conference on computer vision and pattern recognition},
  pages={19113--19122},
  year={2023}
}

@inproceedings{zhu2023prompt,
  title={Prompt-aligned gradient for prompt tuning},
  author={Zhu, Beier and Niu, Yulei and Han, Yucheng and Wu, Yue and Zhang, Hanwang},
  booktitle={Proceedings of the IEEE/CVF international conference on computer vision},
  pages={15659--15669},
  year={2023}
}

@inproceedings{bossard2014food,
  title={Food-101--mining discriminative components with random forests},
  author={Bossard, Lukas and Guillaumin, Matthieu and Van Gool, Luc},
  booktitle={Computer vision--ECCV 2014: 13th European conference, zurich, Switzerland, September 6-12, 2014, proceedings, part VI 13},
  pages={446--461},
  year={2014},
  organization={Springer}
}

@article{krizhevsky2012imagenet,
  title={Imagenet classification with deep convolutional neural networks},
  author={Krizhevsky, Alex and Sutskever, Ilya and Hinton, Geoffrey E},
  journal={Advances in neural information processing systems},
  volume={25},
  year={2012}
}

@inproceedings{orhan2018simple,
  title       = {{A Simple Cache Model for Image Recognition}},
  author      = {Orhan, Emin},
  booktitle   = {Advances in Neural Information Processing Systems},
  volume      = {31},
  year        = {2018},
  note        = {NeurIPS 2018}
}

@inproceedings{khandelwal2020knnlm,
  title       = {{Generalization through Memorization: Nearest Neighbor Language Models}},
  author      = {Khandelwal, Urvashi and Levy, Omer and Jurafsky, Dan and Zettlemoyer, Luke and Lewis, Mike},
  booktitle   = {International Conference on Learning Representations (ICLR)},
  year        = {2020}
}

@inproceedings{vaswani2017attention,
  title        = {Attention Is All You Need},
  author       = {Vaswani, Ashish and Shazeer, Noam and Parmar, Niki and Uszkoreit, Jakob and Jones, Llion and Gomez, Aidan N. and Kaiser, {\L}ukasz and Polosukhin, Illia},
  booktitle    = {Advances in Neural Information Processing Systems},
  volume       = {30},
  pages        = {5998--6008},
  year         = {2017}
}

@inproceedings{vinyals2016matching,
  title        = {Matching Networks for One-Shot Learning},
  author       = {Vinyals, Oriol and Blundell, Charles and Lillicrap, Timothy and Wierstra, Daan},
  booktitle    = {Advances in Neural Information Processing Systems},
  volume       = {29},
  pages        = {3630--3638},
  year         = {2016}
}

@article{helber2019eurosat,
  title={Eurosat: A novel dataset and deep learning benchmark for land use and land cover classification},
  author={Helber, Patrick and Bischke, Benjamin and Dengel, Andreas and Borth, Damian},
  journal={IEEE Journal of Selected Topics in Applied Earth Observations and Remote Sensing},
  volume={12},
  number={7},
  pages={2217--2226},
  year={2019},
  publisher={IEEE}
}

@inproceedings{snell2017protonet,
  title     = {Prototypical Networks for Few-Shot Learning},
  author    = {Snell, Jake and Swersky, Kevin and Zemel, Richard},
  booktitle = {Advances in Neural Information Processing Systems},
  volume    = {30},
  pages     = {4077--4087},
  year      = {2017}
}

@inproceedings{sung2018relationnet,
  title     = {Learning to Compare: Relation Network for Few-Shot Learning},
  author    = {Sung, Flood and Yang, Yongxin and Zhang, Li and Xiang, Tao and Torr, Philip and Hospedales, Timothy},
  booktitle = {Proceedings of the IEEE Conference on Computer Vision and Pattern Recognition},
  pages     = {1199--1208},
  year      = {2018}
}

@inproceedings{kipf2017gcn,
  title     = {Semi-Supervised Classification with Graph Convolutional Networks},
  author    = {Kipf, Thomas and Welling, Max},
  booktitle = {International Conference on Learning Representations},
  year      = {2017}
}

@inproceedings{garcia2018fewshot,
  title     = {Few-Shot Learning with Graph Neural Networks},
  author    = {Garcia, Victor and Bruna, Joan},
  booktitle = {International Conference on Learning Representations},
  year      = {2018}
}

@inproceedings{ziko2020laplacianshot,
  title     = {LaplacianShot: Laplacian-Based Representation for Few-Shot Classification},
  author    = {Ziko, Imtiaz and Gidel, Gauthier and Louedec, Karim and Belilovsky, Eugene and Mitliagkas, Ioannis},
  booktitle = {Proceedings of the IEEE/CVF Conference on Computer Vision and Pattern Recognition},
  pages     = {6548--6557},
  year      = {2020}
}

@inproceedings{silva2024closer,
  title={A closer look at the few-shot adaptation of large vision-language models},
  author={Silva-Rodriguez, Julio and Hajimiri, Sina and Ben Ayed, Ismail and Dolz, Jose},
  booktitle={Proceedings of the IEEE/CVF Conference on Computer Vision and Pattern Recognition},
  pages={23681--23690},
  year={2024}
}

@article{zhang2024ta,
  title={Ta-Adapter: Enhancing few-shot CLIP with task-aware encoders},
  author={Zhang, Wenbo and Zhang, Yifan and Deng, Yuyang and Zhang, Wenlong and Lin, Jianfeng and Huang, Binqiang and Zhang, Jinlu and Yu, Wenhao},
  journal={Pattern Recognition},
  volume={153},
  pages={110559},
  year={2024},
  publisher={Elsevier}
}

@inproceedings{kim2019metagnn,
  title     = {Meta-Graph Neural Network for Few-Shot Learning},
  author    = {Kim, Donghyun and Seo, In and Jeon, Woong and Kang, U.},
  booktitle = {Advances in Neural Information Processing Systems},
  volume    = {32},
  pages     = {2204--2214},
  year      = {2019}
}

@inproceedings{velivckovic2018graph,
  title={Graph Attention Networks},
  author={Veli{\v{c}}kovi{\'c}, Petar and Cucurull, Guillem and Casanova, Arantxa and Romero, Adriana and Li{\`o}, Pietro and Bengio, Yoshua},
  booktitle={International Conference on Learning Representations},
  year={2018}
}

@inproceedings{kim2022find,
    title={How to Find Your Friendly Neighborhood: Graph Attention Design with Self-Supervision},
    author={Dongkwan Kim and Alice Oh},
    booktitle={International Conference on Learning Representations},
    year={2021}
}

\end{document}